\documentclass[11pt]{article}

\usepackage[preprint]{acl}

\usepackage{times}
\usepackage{latexsym}

\usepackage[T1]{fontenc}

\usepackage[utf8]{inputenc}

\usepackage{microtype}

\usepackage{inconsolata}

\usepackage{graphicx}

\usepackage{amsmath}
\usepackage{amssymb}
\usepackage{mathtools}
\usepackage{amsthm}
\usepackage{booktabs}   
\usepackage{multirow}
\usepackage{siunitx}    
\usepackage{subcaption}
\usepackage{todonotes}
\usepackage{placeins}

\title{Beyond Retraining-Free MoE Compression: \\ A Cost-Normalized Study of Post-Compression Adjustment}

\author{
 \textbf{Sieun Hyeon\textsuperscript{1}}, 
 \textbf{Jaeyoung Do\textsuperscript{1,2}}\thanks{Corresponding author} 
\\
 \textsuperscript{1}Department of Electrical and Computer Engineering, Seoul National University, \\
 \textsuperscript{2}Interdisciplinary Program in Artificial Intelligence, Seoul National University
\\
\textit{\{zxc2692, jaeyoung.do\}@snu.ac.kr} \\
{\footnotesize\url{https://aidaslab.github.io/Post-Compression-Adjustment/}}
}

\begin{document}
\maketitle

\begin{abstract}
Retraining-free MoE compression reduces deployment memory by pruning or merging experts, but often treats the compressed checkpoint as the final artifact. 
We argue that this view is incomplete: compressed MoE checkpoints are better understood as compressed initializations that benefit from a tiny post-compression adjustment stage. 
Across two MoE LLM backbones, four pruning/merging methods, three expert-retention ratios, and 28 benchmarks, we compare LM fine-tuning and teacher-based KD under matched small-data budgets and measured GPU costs. 
Using only 3,000 C4 examples and a single epoch of adjustment, \textsc{Full FT} recovers 37.3\% of the original-to-compressed performance gap on average. 
Moreover, LM fine-tuning is more cost-effective than standard token-level KD, and full-parameter adjustment gives the strongest cost--recovery trade-off among the tested scopes. These results suggest that retraining-free compression should be paired with small post-compression adjustment to recover a substantial portion of the performance lost during compression.
\end{abstract}

\section{Introduction}
\label{sec:intro}

Large language models exhibit strong performance, but as the number of parameters increases, their memory footprint and deployment cost grow rapidly \cite{scalinglaws}. Mixture-of-Experts (MoE) \cite{MoE1, MoE3} reduces computation cost by activating only a subset of experts for each token, but it still requires all expert parameters to be loaded into memory during inference, making deployment difficult in resource-constrained environments.

Retraining-free MoE compression is an attractive approach for alleviating this memory bottleneck. For ordinary researchers or developers, preparing large-scale training data and sufficient GPU resources to deploy or compress a large-scale MoE LLM is burdensome. For this reason, many recent studies have emphasized retraining-free or calibration-free compression that reduces the memory footprint of MoE LLMs without full-scale retraining. In particular, because most MoE parameters are concentrated in experts, prior work has focused on expert-side compression, such as Expert Pruning~\cite{aimer} and Expert Merging~\cite{mcsmoe}.

However, retraining-free does not necessarily mean cost-free. Many methods still use calibration samples, model inference, and GPU computation to calculate expert importance~\cite{naee}, expert similarity~\cite{hcsmoe}, routing statistics~\cite{mcsmoe},  and so on. This point leads us to ask again whether retraining-free compression alone is sufficient. \textit{If a small amount of data and GPU computation is already allowed, is it truly optimal to use them only for the compression process itself?}

In this paper, we revisit the existing perspective on retraining-free MoE compression. We argue that a compressed checkpoint should be viewed not as the final artifact, but as a compressed initialization. Expert Pruning and Expert Merging change the expert landscape of the original MoE, leaving residual errors such as routing shift, expert function distortion, and router–expert interaction error. Therefore, if the goal is near-original recovery, a small post-compression adjustment after retraining-free compression is not merely an additional step, but a practically important recovery stage.

Here, post-compression adjustment refers to a procedure that applies a small amount of general-domain text and short gradient updates to a compressed checkpoint in order to reduce the residual performance gap. This is clearly distinguished from full-scale retraining, which requires a large-scale corpus or a long training schedule. Our goal is not to introduce another compressor, but to define and evaluate post-compression adjustment as a missing design axis in MoE compression. Rather than asking only which retraining-free compressor produces the best immediate checkpoint, we ask which compression-and-adjustment pipeline yields the best recovery under the same small data budget and measured GPU cost.

To verify the necessity of post-compression adjustment, we conduct large-scale comparative experiments across Expert Pruning and Expert Merging. Specifically, we construct a variety of compressed checkpoints using two MoE LLM backbones, four primary compression methods from Expert Pruning and Expert Merging, and three expert-parameter retention ratios, and apply the same small adjustment data budget and hyperparameters to all checkpoints. We then evaluate both the degree of performance recovery and GPU cost on 28 downstream benchmarks, thereby analyzing a general post-compression recovery stage that is not dependent on a specific compressor.

In addition, to closely analyze the design choices of post-compression adjustment, we compare two objectives, causal language modeling (LM) loss \cite{gpt1} and teacher-based knowledge distillation (KD) \cite{kd}, and evaluate several trainable parameter scopes: router-only, router+top-k experts, router+all-experts, and full-parameter adjustment. In particular, for top-k experts, we expand k to 8, 16, and 50, and analyze the trade-off among the number of selected experts, recovery gain, and GPU cost. Through this, we test whether carefully selecting and adjusting only a small subset of parameters is truly cost-efficient.

Our experiments present three practical lessons.

\noindent (1) With a single post-compression adjustment epoch over 3,000 C4
examples, compressed MoE LLMs recover 37.3\% of the original-to-compressed performance gap on average. This implies that retraining-free compression should be viewed not as producing a final artifact, but as producing a post-compression-adjustment-friendly initialization under a tiny adjustment budget.

\noindent (2) Under a small general-domain adjustment budget, causal LM fine-tuning is more cost-effective than standard token-level teacher KD in our setting. After compression, when the student’s expert landscape and feasible function class have changed, directly imitating the teacher distribution may not always be the most efficient local repair direction. In contrast, LM fine-tuning directly improves the observed next-token likelihood without teacher forward passes, providing a simpler and more cost-effective repair signal.

\noindent (3) Full-parameter causal LM adjustment is simple but highly competitive. Although it updates more parameters and may require more optimizer memory, its end-to-end GPU time can be competitive because it avoids expert-selection overhead and yields substantially larger recovery per run. Under our measured GPU-cost protocol, it emerges as the strongest default strategy among the tested scopes. Router-only adjustment can alleviate some routing mismatch, but it is not sufficient to recover expert function distortion and router–expert interaction error. Router+top-k experts adjustment limits the number of trainable parameters, but additionally incurs selection cost for identifying frequently activated experts.

As a result, this paper does not reject the recent trend of retraining-free MoE compression, but instead reinterprets it as a more practical two-stage pipeline. Retraining-free is a strong start, but not necessarily the end. The central lesson of this paper is that, for near-original MoE recovery, not only the compressor itself but also the post-compression adjustment stage should be designed together.

\section{Related Works}
\label{sec:related_works}

\textbf{Expert Pruning} removes a subset of redundant experts. Strategies include minimizing reconstruction loss~\cite{naee}, differentiable selection~\cite{bai2025diep}, and leveraging router magnitudes~\cite{reap}. Other approaches utilize output discrepancy bounds~\cite{anonymous2025compressing}, token variation~\cite{easy_ep}, trajectory-based importance~\cite{moepathfinder}, or coarser layer-level pruning~\cite{RS_TMLR}. \textbf{Expert Merging} is grounded in model merging hypotheses~\cite{modelsoup, mergingmodelsfisher}. This approach synthesizes experts via output similarity clustering~\cite{hcsmoe}, selective dual-masks~\cite{puzzlemoe}, or compression matrices~\cite{mergemoe}.


Some MoE compression studies include a recovery stage, such as expert-wise KD after pruning~\cite{moepruner}, condensed-layer fine-tuning~\cite{cdmoe}, LoRA-based recovery after hybrid compression~\cite{moei2}, or fine-tuning after subspace expert merging~\cite{submoe}; larger systems further rely on distillation or continual training at much larger data scales~\cite{slimmoe,slimqwen}.
However, these works do not systematically ask whether retraining-free compression itself is sufficient across already-deployable expert-compressed MoE checkpoints, nor which post-compression objective, trainable scope, and GPU-cost trade-off is optimal under a tiny calibration budget.
Our work fills this gap by treating post-compression adjustment as a first-class component of MoE compression rather than a method-specific add-on. See Appendix~\ref{appendix:related_works} for extended related works.

\section{Causes of Performance Degradation}
\label{sec:degradation}

Retraining-free MoE compression changes the computation path of the original
model. This change can come from removing experts or merging multiple experts
into a smaller set. As a result, the compressed model can deviate from the
original model through two coupled sources: the router may assign different
mixture weights, and the selected expert functions or expert paths may no
longer match the original ones. In this section, we use Expert Pruning as a
representative case and keep only the core equations in the main text. Full
derivations for Expert Pruning and Expert Merging are provided in
Appendices~\ref{appendix:equations_pruning}
and~\ref{appendix:equations_merging}, respectively.

We adopt the MoE notation and pruning scenario analysis of
\citet{hyeon2026retraining}. To isolate the local compression source, the
derivation below evaluates the original and compressed MoE layer maps at the
same reference hidden-state input $x$. Differences caused by the realized
end-to-end state shift are handled separately in the propagation analysis
below.

Let $\mathcal{S}$ be the top-$k$ expert set selected by the original MoE
layer, let $\mathcal{P}\subseteq\{0,\ldots,N-1\}$ be the set of experts
retained after pruning, and let $\mathcal{S}'\subseteq\mathcal{P}$ be the
top-$k$ set selected by the pruned layer at the same reference input $x$.
For any active set $\mathcal{A}$, let
$\widetilde g_i^{o,\mathcal{A}}(x)$ and
$\widetilde g_i^{p,\mathcal{A}}(x)$ denote the original and pruned router
weights renormalized over $\mathcal{A}$, respectively. For readability, we
omit $x$ below and also write $E_i=E_i(x)$.

Building on the prior three-part pruning analysis, we regroup the residual
into shared-path router reweighting and active-path replacement. The overlap
is defined between the expert sets that are actually selected:
$$
\begin{aligned}
\mathcal{T} &= \mathcal{S}\cap\mathcal{S}',
&\mathcal{D} &= \mathcal{S}\setminus\mathcal{S}',
&\mathcal{R} &= \mathcal{S}'\setminus\mathcal{S}.
\end{aligned}
\label{eq:tdr_sets}
$$
Thus,
$\mathcal{S}=\mathcal{T}\mathbin{\dot\cup}\mathcal{D}$ and
$\mathcal{S}'=\mathcal{T}\mathbin{\dot\cup}\mathcal{R}$.

\paragraph{Best scenario.}
The most favorable case is $\mathcal{S}'=\mathcal{S}$, which already implies
$\mathcal{S}\subseteq\mathcal{P}$. Then
$$
\left\lVert
 y_{\mathrm{orig}}-y_{\mathrm{pruned}}^{\mathrm{best}}
\right\rVert
=
\left\lVert
\sum_{i\in\mathcal{S}}
\left(
\widetilde g_i^{o,\mathcal{S}}
-
\widetilde g_i^{p,\mathcal{S}}
\right)E_i
\right\rVert.
\label{eq:best_pruning_short}
$$
At the common reference input, the active expert functions are
identical, so the local residual can contain only shared-path router
reweighting. If pure pruning leaves the router and gate implementation
unchanged, then $\tilde g_i^{p,S}(x)=\tilde g_i^{o,S}(x)$ and this
same-input best-case residual is exactly zero. Realized end-to-end
deviations may nevertheless persist through upstream hidden-state
shifts. If the gate implementation has changed or the router is
subsequently adjusted, router alignment is the only local correction
needed while $S'=S$.

\paragraph{Most common scenario.}
In the partial-overlap case, $0<|\mathcal{T}|<k$, and
\begin{equation}
\left\lVert
 y_{\mathrm{orig}}-y_{\mathrm{pruned}}^{\mathrm{common}}
\right\rVert
=
\left\lVert
 e_{\mathrm{route}}^{p}+e_{\mathrm{path}}^{p}
\right\rVert,
\label{eq:common_compact}
\end{equation}
where
\begin{align}
 e_{\mathrm{route}}^{p}
 &=
 \sum_{i\in\mathcal{T}}
 \left(
 \widetilde g_i^{o,\mathcal{S}}
 -
 \widetilde g_i^{p,\mathcal{S}'}
 \right)E_i,
 \label{eq:route_error}
 \\
 e_{\mathrm{path}}^{p}
 &=
 \sum_{i\in\mathcal{D}}
 \widetilde g_i^{o,\mathcal{S}}E_i
 -
 \sum_{i\in\mathcal{R}}
 \widetilde g_i^{p,\mathcal{S}'}E_i.
 \label{eq:path_error}
\end{align}
Within a fixed active-set region, $e_{\mathrm{route}}^{p}$ is the component
that router-only adjustment directly targets by changing mixture weights on
shared experts. Router updates may also move the top-$k$ boundary and thereby
change $\mathcal{T}$, $\mathcal{D}$, and $\mathcal{R}$. However, with expert
parameters frozen, router-only adjustment cannot change the expert functions
themselves and therefore cannot in general reconstruct a missing expert
contribution unless the available replacements already provide an adequate
functional approximation. Appendix~\ref{appendix:equations_pruning} gives the
unified derivation. The cluster-space counterpart for Expert Merging is given
in Appendix~\ref{appendix:equations_merging}.

\paragraph{Scenario frequency.}
The decomposition suggests that router-only adjustment is most favorable
when the active path is preserved and only mixture weights require
correction.
Using 100 questions sampled from ELI5 \citep{eli5}, we categorize token--layer
observations into Best, Most Common, and Worst scenarios using the active
sets from the original and compressed end-to-end forward passes.
These realized frequencies include upstream hidden-state perturbations
and therefore diagnose path preservation; they do not by themselves
isolate the magnitude of each local residual term.
For pruning, overlap is measured between the original and pruned active
sets.
For merging, both paths are compared in the physical cluster space
defined by the mapping $\phi$ in
Appendix~\ref{appendix:equations_merging}.

\begin{table}[t]
\centering
\small
\begin{tabular}{lrrrr}
\toprule
Model & Ratio & Best(\%) & Most(\%) & Worst(\%) \\
\midrule
Gemma4 & 50\% & 8.63 & 90.76 & 0.61 \\
Gemma4 & 62.5\% & 11.38 & 88.25 & 0.37 \\
Gemma4 & 75\% & 15.52 & 84.20 & 0.28 \\
Qwen3 & 50\% & 8.05 & 91.73 & 0.22 \\
Qwen3 & 62.5\% & 15.81 & 84.14 & 0.05 \\
Qwen3 & 75\% & 32.15 & 67.84 & 0.01 \\
\bottomrule
\end{tabular}
\caption{
Scenario-frequency summary by backbone and expert retention ratio. Results are based on 100
samples from the ELI5 dataset. Best denotes cases where the original selected expert set is preserved, Most denotes partial overlap between the original and compressed selected expert sets, and Worst denotes disjoint selected expert sets. For each model--ratio pair, results are averaged across the corresponding compression methods.
}
\label{tab:scenario_frequency}
\end{table}

Table~\ref{tab:scenario_frequency} shows that the best scenario is
relatively rare, while partial overlap dominates across both backbones
and all expert retention ratios.
This frequency analysis does not determine whether routing error or
expert/path error is larger in norm.
Rather, it shows that the case in which router-only correction can
account for the entire local residual is not the typical realized path.
In the dominant partial-overlap regime, both
$e_{\mathrm{route}}^{p}$ and $e_{\mathrm{path}}^{p}$ can contribute,
motivating our comparison of router-only, expert-inclusive, and
full-parameter adjustment.

\paragraph{Noise propagation caused by compression.}
Let $F_\ell$ and $\widehat F_\ell$ denote the original and compressed layer
maps, respectively. Let $h_\ell$ and $\widehat h_\ell$ be their hidden states
before layer $\ell$, and define
$\delta_\ell=\widehat h_\ell-h_\ell$. The same-input local compression source
is
$$
\epsilon_\ell(h_\ell)
:=
\widehat F_\ell(h_\ell)-F_\ell(h_\ell).
\label{eq:local_compression_source}
$$
For the pruning contribution in Eqs.~\eqref{eq:route_error}--
\eqref{eq:path_error}, which uses the original-minus-compressed convention,
$$
\epsilon_\ell^{p}
=
-\left(
 e_{\mathrm{route},\ell}^{p}
 +
 e_{\mathrm{path},\ell}^{p}
\right).
\label{eq:local_compression_error}
$$
Within a fixed-routing neighborhood, or away from top-$k$ decision
boundaries, first-order linearization gives
$$
\delta_{\ell+1}
\approx
\widehat J_\ell\delta_\ell+
\epsilon_\ell(h_\ell),
\qquad
\widehat J_\ell
:=
\left.
\frac{\partial\widehat F_\ell(h)}{\partial h}
\right|_{h=h_\ell}.
\label{eq:noise_recurrence}
$$
Assuming both models receive the same input, $\delta_0=0$, and unrolling the
recurrence yields
$$
\delta_L
\approx
\sum_{\ell=0}^{L-1}
P_{\ell\rightarrow L}\epsilon_\ell \quad, P_{\ell\rightarrow L}:=
\widehat J_{L-1}\widehat J_{L-2}
\cdots\widehat J_{\ell+1},
\label{eq:noise_unroll}
$$
\noindent with $P_{L-1\rightarrow L}=I$ for the empty product. Thus, even when only MoE
experts are compressed, the induced local source enters the residual stream
and is transformed by subsequent attention, normalization, dense, and MoE
modules.

This propagation view motivates a local sensitivity comparison of trainable
parameter scopes. Let $\theta_R$, $\theta_E$, and $\theta_D$ denote router,
expert, and remaining dense/shared parameters of the compressed model, and let
$\theta^0$ be the unadjusted compressed checkpoint. Define the final-state
discrepancy $d_L(\theta)=\widehat h_L(\theta)-h_L$ and, within a fixed-routing
local region,
$$
G_b
:=
\left.
\frac{\partial d_L}{\partial\theta_b}
\right|_{\theta=\theta^0},
\qquad
b\in\{R,E,D\}.
\label{eq:source_adjust}
$$
A first-order parameter perturbation then satisfies
$$
\Delta d_L
\approx
G_R\Delta\theta_R
+
G_E\Delta\theta_E
+
G_D\Delta\theta_D.
\label{eq:full_noise_adjust}
$$

\noindent Let $\mathcal{K}$ denote the selected trainable experts and let $G_{E,\mathcal{K}}$ be the sub-Jacobian formed by the corresponding columns of $G_E$. Then the locally reachable spaces satisfy

\begin{align*}
&\operatorname{col}(G_R)
\subseteq \operatorname{col}([G_R\;G_{E,\mathcal{K}}])
\subseteq \operatorname{col}([G_R\;G_E])
\\ 
&\subseteq \operatorname{col}([G_R\;G_E\;G_D]).
\end{align*}

\noindent Thus, a router+top-$k$-expert scope retains all router directions and
adds only the expert directions associated with the selected trainable
experts. These inclusions describe first-order reachable directions,
not optimization success or the relative causal importance of the
parameter groups.
\section{Experiment Setups}
\label{sec:experiment_setups}

Our experiments are designed to answer three questions: \textbf{(1) Can a small amount of post-compression adjustment substantially recover the performance of compressed MoE LLMs? 
(2) When performance recovery and GPU cost are considered, which adjustment strategy is most efficient? 
(3) Which training objective is more suitable for post-compression adjustment?} 
To answer these questions, we evaluate multiple MoE compression paradigms, compression ratios, adjustment objectives, and trainable parameter scopes under a matched adjustment budget.

\subsection{Backbones and Compression Baselines}
\label{sec:baselines}

We use two MoE LLM backbones: \textit{Qwen3-30B-A3B-Instruct-2507} \cite{qwen3technicalreport} and \textit{gemma-4-26B-A4B-it} \cite{gemma4}. 
For each backbone, we construct compressed checkpoints using representative methods from two primary parameter-level MoE compression paradigms: Expert Pruning and Expert Merging.

For \textit{Qwen3-30B-A3B-Instruct-2507}, we use the following methods. 
For Expert Pruning, we adopt REAP~\cite{reap}, which estimates expert importance using both router gate values and expert output magnitudes. 
For Expert Merging, we use HC-SMoE~\cite{hcsmoe}, which performs hierarchical clustering based on expert-output similarity computed on calibration data.

For \textit{gemma-4-26B-A4B-it}, we use a different set of representative methods. 
For Expert Pruning, we use AIMER~\cite{aimer}, a calibration-free pruning method based on RMS-normalized weight magnitudes. 
For Expert Merging, we use M-SMoE~\cite{mcsmoe}, which groups experts using routing-policy statistics, aligns their neurons by permutation, and then merges each group through activation-frequency-weighted averaging.

For each compression method, we evaluate three expert-parameter retention ratios: 50\%, 62.5\%, and 75\%. 
In the pruning setting, for example, a 50\% retention ratio means that half of the expert parameters are removed; for a layer with 128 experts, this corresponds to retaining 64 experts. 
All ratios are applied only to expert parameters, while non-expert parameters are kept unchanged.

\subsection{Benchmarks and Evaluation Protocol}
\label{sec:benchmarks}

We evaluate each model on 28 downstream benchmarks spanning five categories: general reasoning/QA, mathematical reasoning, coding, chain-of-thought (CoT) reasoning, and multiple-choice question answering (MCQA). All benchmarks are detailed in Appendix~\ref{app:benchmarks}.

All benchmark evaluations are conducted using \texttt{lm-evaluation-harness}~\cite{eval-harness} and \texttt{vLLM}~\cite{vLLM}. 
We use greedy decoding with temperature 0 and fix the random seed to 42 for all experiments. 
We report task-specific harness metrics, including accuracy, exact-match, and pass@1 provided by the evaluation harness.
We conduct every post-compression adjustment run on exactly two NVIDIA H200 GPUs (140 GB each), regardless of the backbone, compression method, retention ratio, objective, or trainable scope.
For post-compression adjustment, we use the same C4~\cite{c4} text subset, data budget, and hyperparameters across all models, compression methods, ratios, objectives, and trainable scopes. 
To specifically observe if performance recovery is achievable at a minimal scale, we restrict our adjustment to a single epoch over just 3,000 sampled examples with a batch size of 2.
More detailed hyperparameters and runtime settings are provided in Appendix~\ref{app:exp_settings}.

\subsection{Adjustment Strategies and Metrics}
\label{sec:strategies_metrics}

\subsubsection{Objectives}
We study post-compression adjustment along two axes: the training objective and the trainable parameter scope. 
Let $\theta_c$ denote the parameters of a compressed model and let $\mathcal{U}$ denote the subset of parameters updated during adjustment. 
All parameters outside $\mathcal{U}$ are frozen. 
We compare the following objectives.

\paragraph{Causal language modeling fine-tuning.}
The first objective is standard causal LM loss~\cite{gpt1} on a small general-domain text subset. 
For a sequence $x=(x_1,\ldots,x_L)$ and a loss mask $m_t$, the LM objective is
$$
\mathcal{L}_{\mathrm{LM}}
=
-\frac{1}{N_x}
\sum_{t=1}^{L-1}
m_{t+1}
\log p_{\theta_c}(x_{t+1}\mid x_{\leq t}),
\label{eq:lm_loss}
$$
where $N_x=\sum_{t=1}^{L-1}m_{t+1}$ is the number of valid target tokens. 
This objective directly improves the likelihood of observed text and does not require a teacher forward pass.

\paragraph{Teacher-based knowledge distillation.}
The second objective is token-level knowledge distillation (KD)~\cite{kd} from the original uncompressed model. 
Let $z_T^{(t)}$ and $z_S^{(t)}$ denote the teacher and student vocabulary logits at position $t$, and let $\tau$ be the distillation temperature. 
We define
$$
p_T^{(t)} = \mathrm{softmax}(z_T^{(t)}/\tau),
\quad
p_S^{(t)} = \mathrm{softmax}(z_S^{(t)}/\tau).
$$
The KD objective is
$$
\mathcal{L}_{\mathrm{KD}}
=
\frac{\tau^2}{N_x}
\sum_{t=1}^{L-1}
m_{t+1}
D_{\mathrm{KL}}
\left(
p_T^{(t)}
\Vert
p_S^{(t)}
\right).
\label{eq:kd_loss}
$$
In all experiments, the teacher is the original model before compression and the student is the compressed model after adjustment.

\paragraph{Direct router logit matching.}
We also include Direct Router Logit Matching (DRLM) as a diagnostic baseline when router coordinates can be aligned through the compression mapping. 
We use DRLM only as a diagnostic router-alignment baseline and provide its formulation in Appendix~\ref{appendix:drlm}.

\subsubsection{Trainable parameter scopes.}
For LM and KD, we evaluate multiple trainable parameter scopes:

\begin{itemize}
    \item \textbf{Router-only:} update only router/gating parameters while freezing experts and all shared parameters.
    \item \textbf{Router+top-$k$ experts:} update router parameters and the $k$ most frequently activated experts per MoE layer. 
    We select top-$k$ experts using the same adjustment data before training. 
    We evaluate $k\in\{8,16,50\}$.
    \item \textbf{Router+all-experts:} update router parameters and all expert parameters, while freezing non-expert dense/shared parameters.
    \item \textbf{Full-parameter:} update all parameters in the compressed checkpoint.
\end{itemize}

The router+top-$k$ setting is intended to test whether carefully selecting a small subset of frequently used experts is more cost-efficient than full adjustment. 
However, this strategy also introduces an expert-selection stage, which requires additional calibration inference before the actual adjustment training.

\subsubsection{Adjustment strategies.}
Combining LM/KD objectives with the trainable scopes yields the main set of post-compression adjustment strategies evaluated in this work. DRLM is reported separately as a diagnostic router-matching baseline.
For example, \textsc{Router-only FT} denotes causal LM fine-tuning with only router parameters trainable, while \textsc{Router+top-8 Expert KD} denotes KD where the router and the top-8 selected experts per layer are trainable. 
\textsc{Full FT} and \textsc{Full KD} update all parameters of the compressed checkpoint using LM loss and KD loss, respectively. 
We also compare against the unadjusted compressed baseline to measure recovery gain.

\subsubsection{Performance and GPU-cost metrics.}
For performance, we report the macro-average score over the 28 benchmarks and define recovery gain as the difference, in percentage points (pp), between the adjusted model's score and that of the unadjusted compressed baseline; the recovered fraction of the gap divides this gain by the Original-to-Compressed gap. For GPU cost, we report three complementary metrics: utilization-weighted GPU time, time-integrated GPU-memory occupancy, and GPU energy. Let $T_{\mathrm{hr}}$ be the total wall-clock duration, in hours, of the measured loop segments, $\bar{P}_g$ be the mean
power draw of GPU $g$ over its valid retained readings, $G$ be the
number of assigned GPUs, $\bar{u}$ be the mean GPU-utilization
percentage over valid retained readings, and $\bar{M}_{\mathrm{obs}}$
be the mean observed per-poll sum of \texttt{memory.used} across
assigned GPUs with valid readings. We compute
$$
H_{\mathrm{eff}}
=
G\,T_{\mathrm{hr}}\frac{\bar{u}}{100},
$$
measured in effective GPU-hours,
$$
H_{\mathrm{mem}}
=
T_{\mathrm{hr}}\bar{M}_{\mathrm{obs}},
$$
measured in GPU-memory GiB-hours, and 
$$
E_{\mathrm{GPU}}
=
\frac{T_{\mathrm{hr}}}{1000}
\sum_{g=1}^{G}\bar{P}_g,
$$
measured in kWh. All three metrics use the same measurement boundaries. For KD and DRLM, the monitored loop includes both teacher and student forward passes. When a strategy requires a separate expert-selection phase, all three metrics include both the expert-selection and training-loop segments.

\section{Experiment Results}
\label{sec:results}

\paragraph{Overall recovery.}

\begin{figure}[t]
    \centering 
    \includegraphics[width=\linewidth]{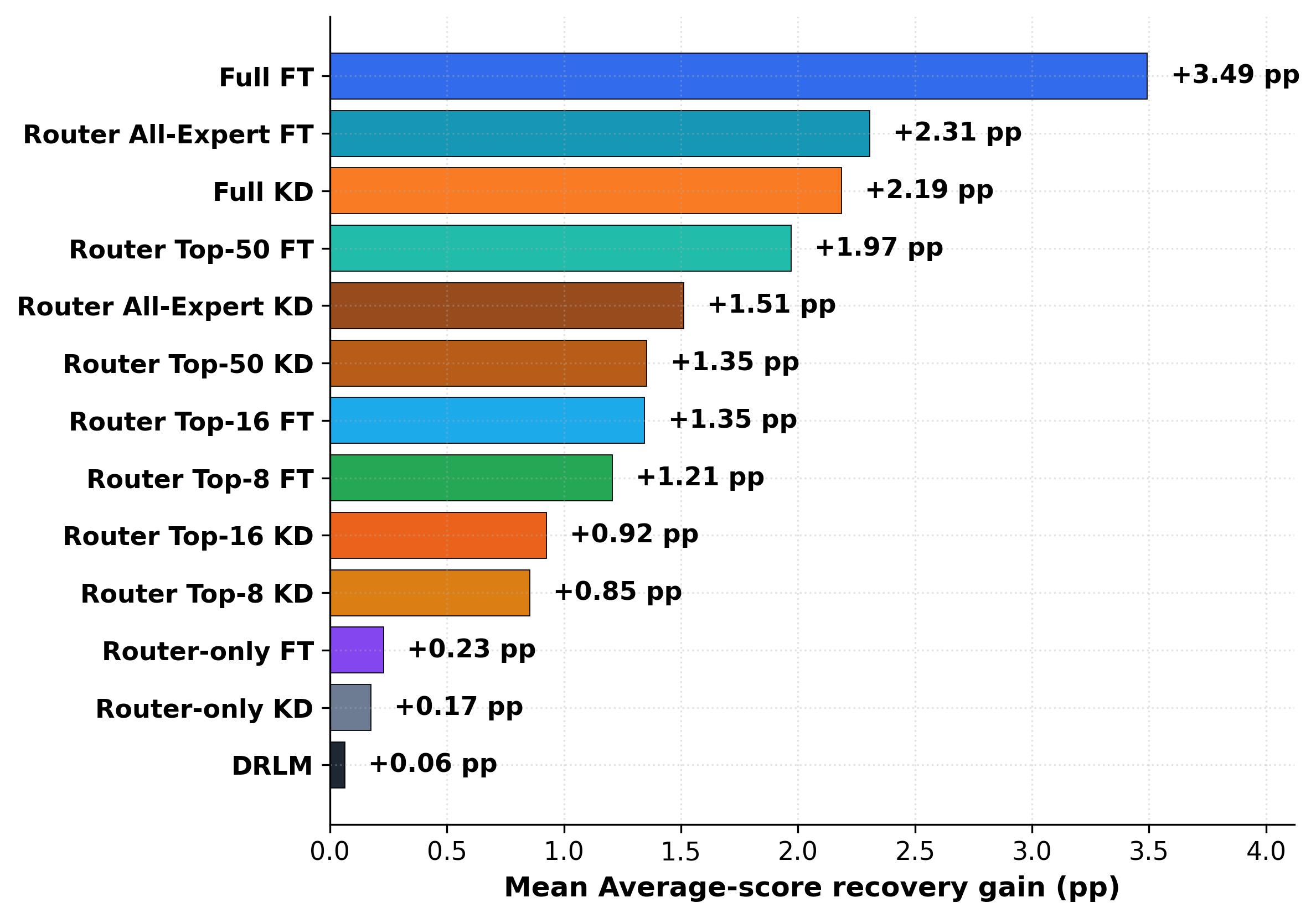}
    \caption{
Overall recovery gain of post-compression adjustment strategies.
Bars show the average-score recovery gain, measured in percentage points (pp), over the unadjusted compressed baseline.
Causal LM fine-tuning consistently outperforms KD under matched trainable scopes, and \textsc{Full FT} achieves the largest recovery.
}
    \label{fig:overall_recovery} 
    \vspace{-1.0em}
\end{figure}

Figure~\ref{fig:overall_recovery} summarizes the mean recovery gain of each post-compression adjustment strategy, averaged over all backbone, compressor, retention-ratio, and benchmark combinations. Appendix~\ref{app:exp_per_methods} further reports the same comparison broken down by compression method.
The strongest strategy is \textsc{Full FT}, which achieves a mean recovery gain of $+3.49$ pp over the unadjusted compressed baseline. In absolute mean score, the original model reaches 48.37\%, the unadjusted compressed baseline drops to 39.01\%, and \textsc{Full FT} improves it to 42.50\%, recovering 37.3\% of the original-to-compressed gap.
It is followed by \textsc{Router+All-Expert FT} ($+2.31$ pp) and \textsc{Full KD} ($+2.19$ pp).
A clear pattern is that, under the same trainable parameter scope, causal LM fine-tuning consistently outperforms teacher-based KD.
For example, \textsc{Router+Top-50 Expert FT} outperforms \textsc{Router+All-Expert KD}, and \textsc{Router+Top-8 Expert FT} outperforms \textsc{Router+Top-16 Expert KD}.
This suggests that, under a tiny adjustment budget, the choice of objective can be more important than simply increasing the number of trainable experts.

Importantly, this result should not be read as a claim that distillation is generally ineffective for post-compression recovery. Our comparison focuses on standard token-level KD under a tiny general-domain adjustment budget, and counts the additional teacher-forward cost required by KD. Under this protocol, causal LM fine-tuning provides a more cost-effective repair signal.

Router-only adjustment also improves performance, but only marginally.
\textsc{Router-only FT} and \textsc{Router-only KD} yield small gains of $+0.23$ pp and $+0.17$ pp, respectively, while direct router logit matching yields only $+0.06$ pp.
This supports the analysis in Section~\ref{sec:degradation}: router reweighting is a real source of error, but the best scenario where router-only correction can fully explain the residual is relatively rare.
In the more common partial-overlap scenario, expert-path and expert-function errors remain, and these cannot be sufficiently reduced by modifying the router alone.

\begin{figure}[!t]
    \centering
    \includegraphics[width=0.95\linewidth]{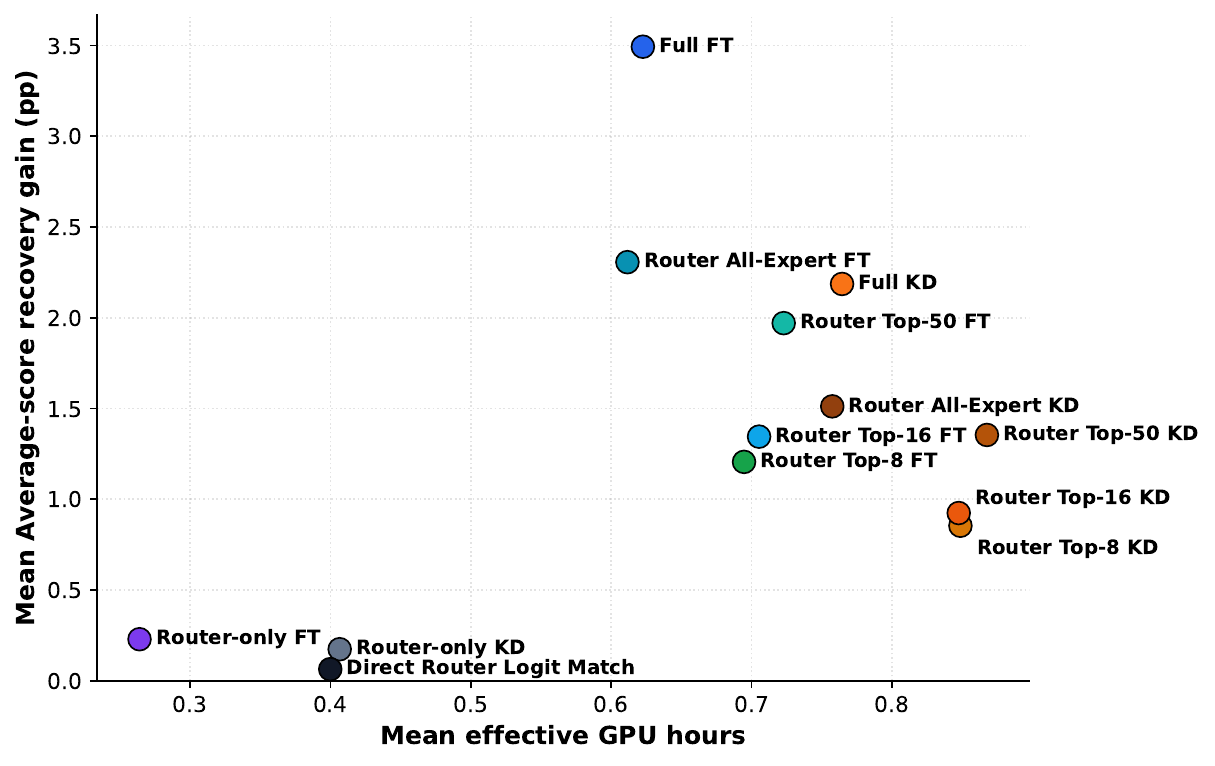}\\[1ex]
    \includegraphics[width=0.95\linewidth]{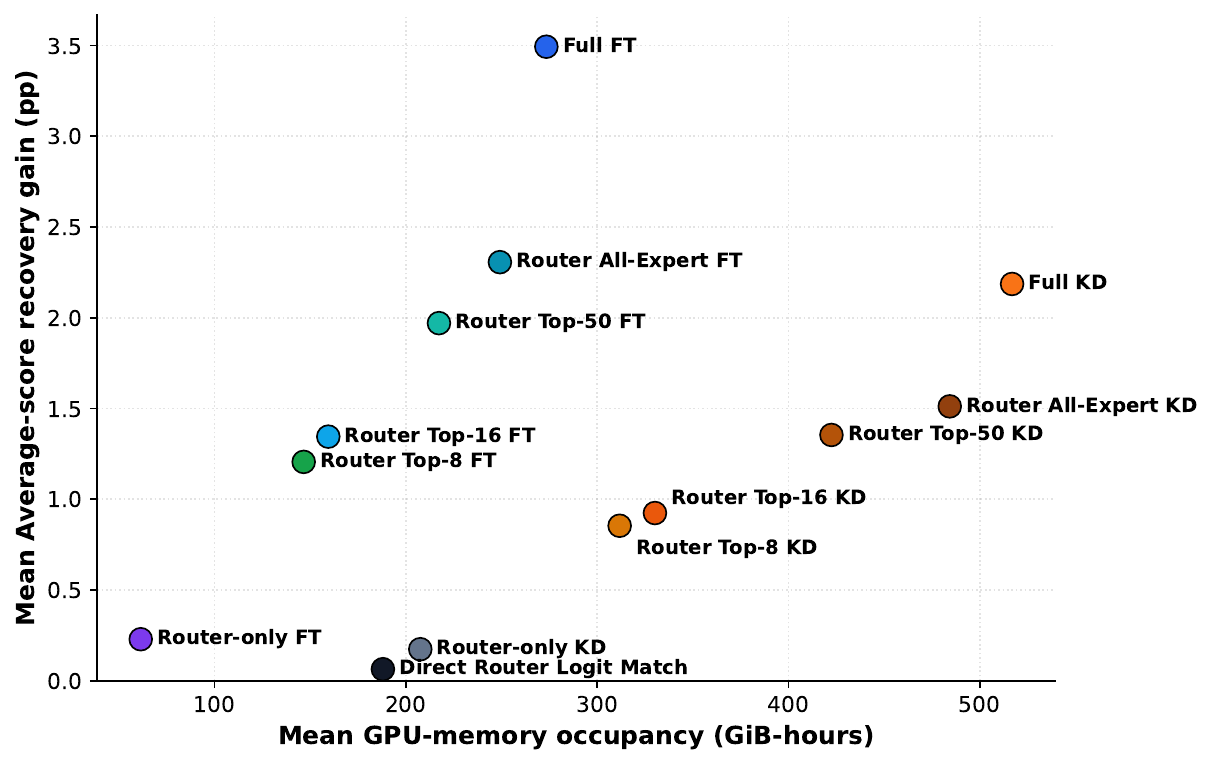}\\[1ex]
    \includegraphics[width=0.95\linewidth]{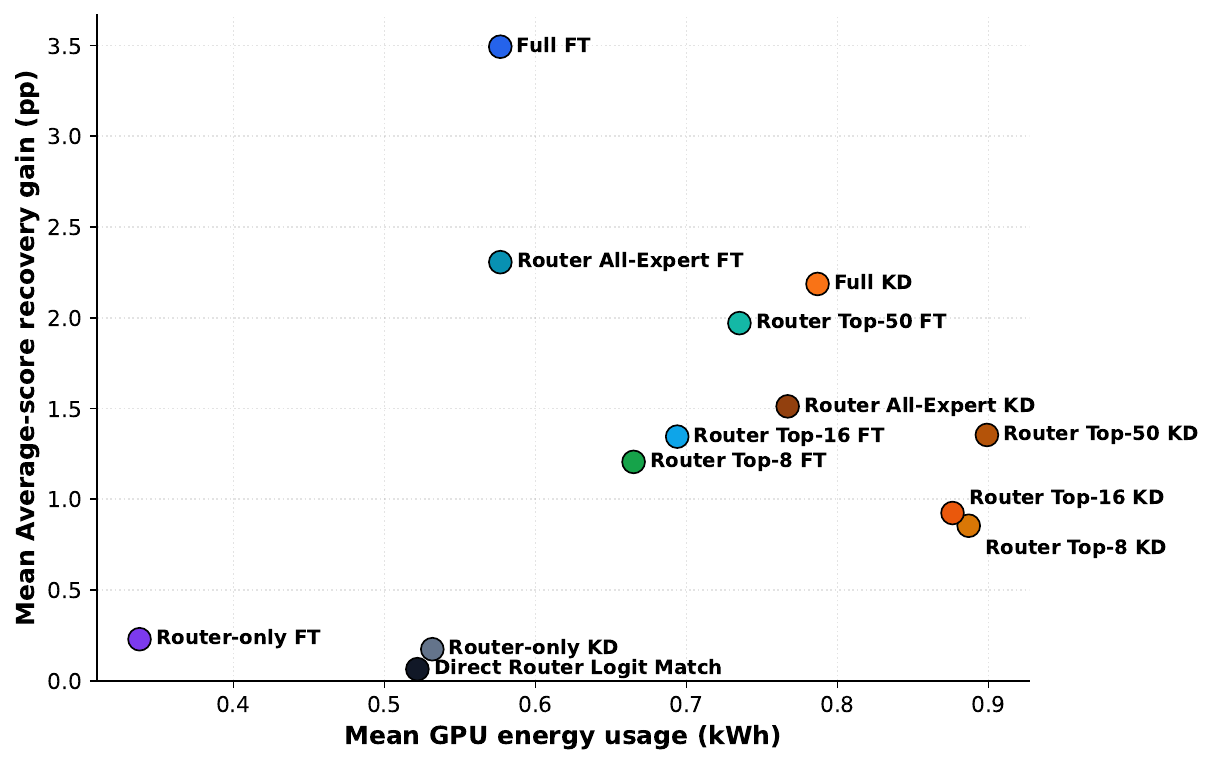}
    
    \caption{GPU cost--performance trade-off of post-compression
adjustment strategies. The y-axis shows mean average-score recovery
gain, while the x-axis shows effective GPU-hours (top), GPU-memory
occupancy in GiB-hours (middle), and GPU energy in kWh (bottom).
Each point is an unweighted arithmetic mean over 12 method--ratio
settings (four compression methods $\times$ three retention ratios).
Both expert-selection cost and training-loop cost are included when
applicable. Upper-left is better.}
    \label{fig:cost_tradeoff}
    \vspace{-1.2em}
\end{figure}

\paragraph{GPU cost--performance trade-off.}
Figure~\ref{fig:cost_tradeoff} compares recovery gain against three GPU-cost
metrics: effective GPU-hours, GPU-memory occupancy in GiB-hours, and GPU energy in kWh.
The desirable region is the upper-left corner, where a strategy achieves high recovery with low cost.
\textsc{Full FT} achieves the highest recovery among all strategies.
Although it is not the cheapest option in absolute GPU cost, it provides the best recovery per end-to-end adjustment run among the high-recovery strategies. This is because expert-subset strategies incur additional selection inference and still recover less than full adjustment.
In particular, the gap between \textsc{Router+All-Expert FT} and \textsc{Full FT} shows that updating only the compressed MoE modules is not necessarily the most efficient use of the adjustment budget.

The router+top-$k$ strategies are also less cost-efficient than their trainable-parameter count might suggest.
These strategies update only a subset of frequently activated experts, but they require a separate expert-selection stage before training.
This additional calibration inference increases GPU cost, while the recovery gain remains lower than that of \textsc{Router+All-Expert FT} or \textsc{Full FT}.
Thus, unless the selected experts can be identified almost perfectly at negligible cost, restricting adjustment to a small top-$k$ expert subset is not clearly more efficient than adjusting all experts or the full model.

Router-only strategies occupy the low-cost region, but their recovery gains are also small.
They may be useful when GPU resources are extremely limited, but they should not be expected to approach the recovery of full-parameter adjustment.
Overall, when performance recovery and GPU cost are considered, \textsc{Full FT} emerges as the strongest default strategy among the tested methods.

\begin{figure}[t]
    \centering 
    \includegraphics[width=\linewidth]{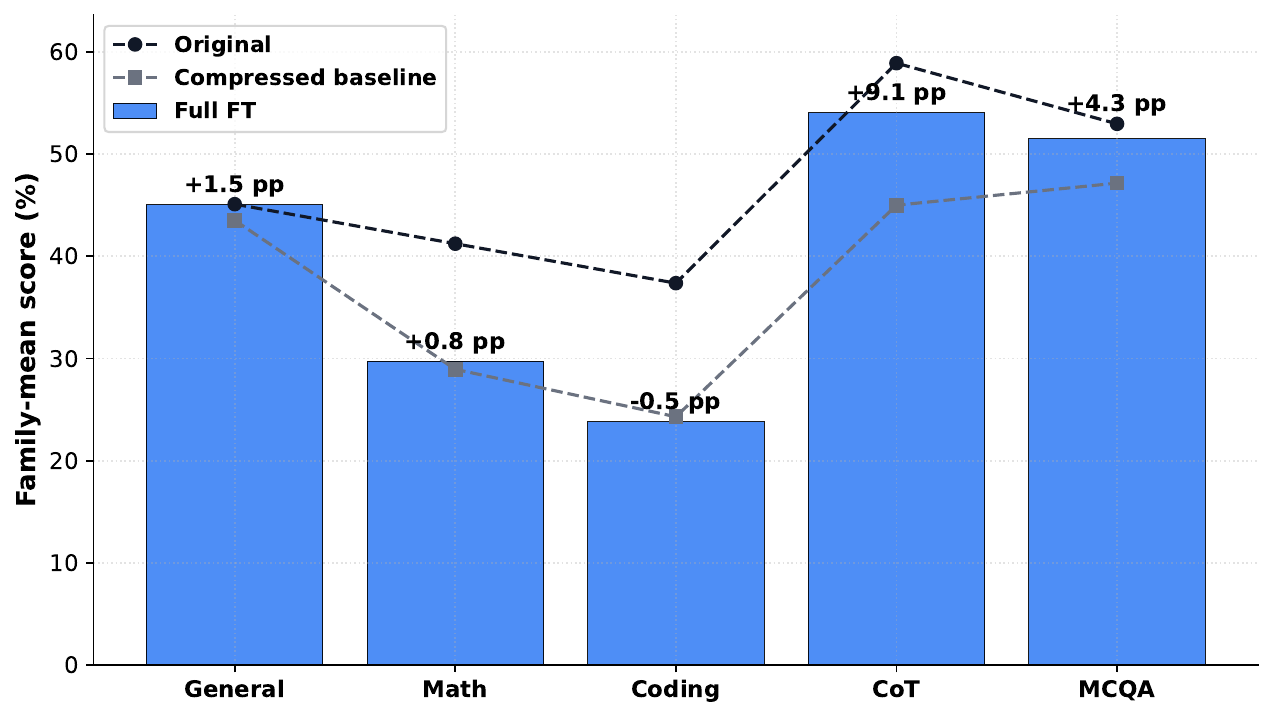}
    \caption{
Benchmark-family recovery of \textsc{Full FT}. Bars show the post-adjustment performance, and dashed lines denote the original model and the unadjusted compressed baseline. \textsc{Full FT} yields the largest recovery on chain-of-thought and MCQA benchmarks, while math and coding show only marginal changes.}
    \label{fig:family_recovery} 
    \vspace{-1.2em}
\end{figure}

\paragraph{Which benchmark families recover?}
Figure~\ref{fig:family_recovery} shows the benchmark-family recovery obtained by \textsc{Full FT}.
The largest gains appear in chain-of-thought reasoning and multiple-choice QA, followed by general reasoning and QA.
In contrast, math and coding benchmarks show only marginal changes.
This pattern suggests that a small general-domain LM adjustment is especially effective at restoring the model's general generation behavior and long-form reasoning stability, while highly specialized domains such as mathematics and coding may require more task-specific knowledge or longer adaptation.

We also qualitatively observe generation-level failures after compression, including token repetition, premature termination, and loss of coherent continuation. Appendix~\ref{app:qual_rep_example} provides representative examples where \textsc{Full FT} restores fluent continuation without task-specific supervision. This indicates that compression can damage not only factual knowledge but also the model's ability to express and compose its remaining knowledge fluently.
Causal LM fine-tuning directly optimizes observed next-token likelihood on natural text, and therefore provides a simple signal for repairing such generation-level degradation.
This helps explain why \textsc{Full FT} is particularly effective on general, CoT, and MCQA-style benchmarks, where fluent continuation and stable reasoning traces are important.

\begin{figure}[t]
    \centering 
    \includegraphics[width=\linewidth]{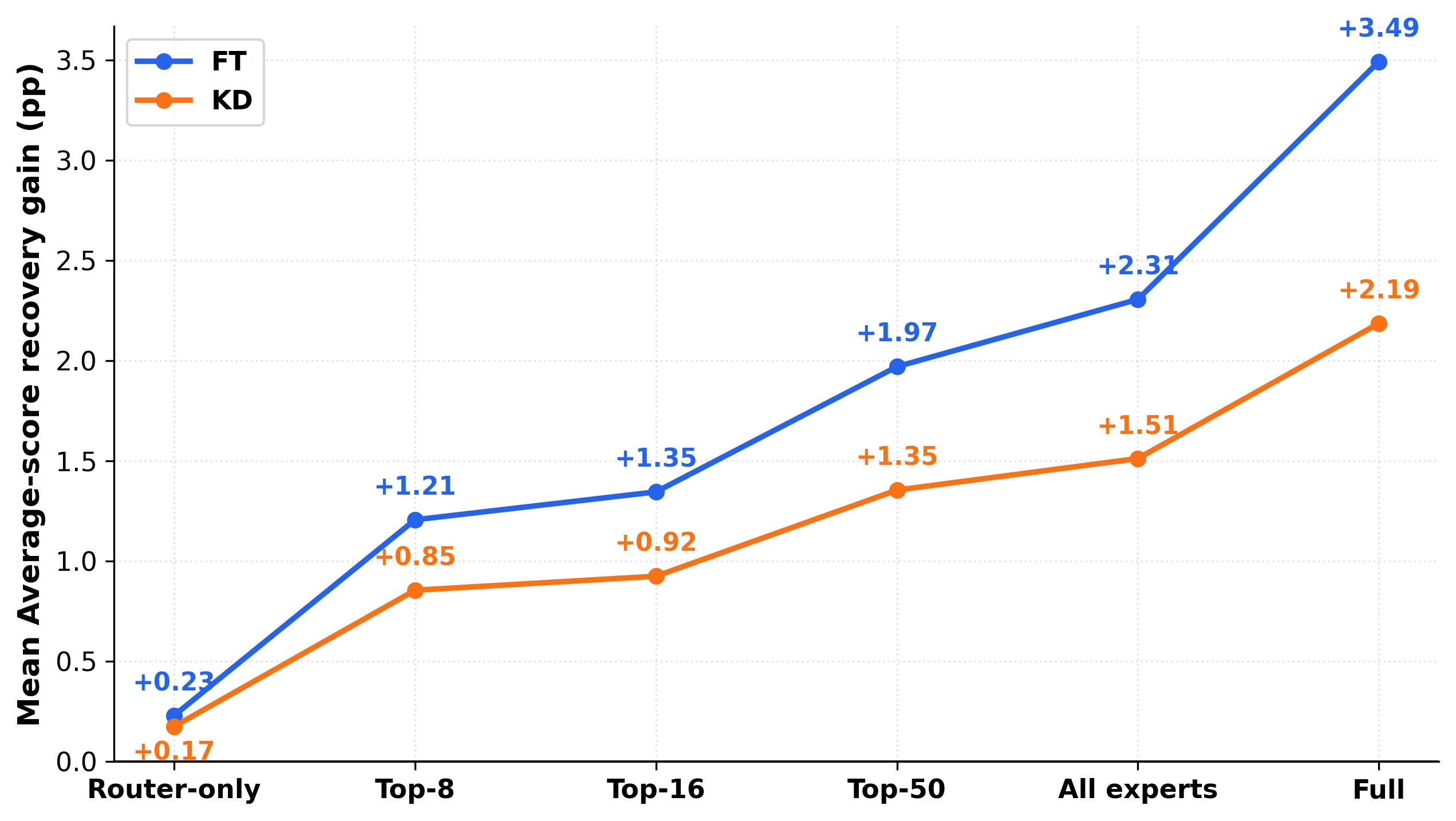}
    \caption{
Recovery gain as the trainable parameter scope expands.
For both LM fine-tuning and KD, recovery increases from router-only to router+top-$k$ experts, router+all-experts, and full-parameter adjustment.
The large gain from router+all-experts to full adjustment suggests that compression-induced noise propagates beyond MoE modules into dense/shared components.
}
    \label{fig:scope_progression} 
    \vspace{-1.2em}
\end{figure}

\paragraph{Effect of expanding the trainable scope.}
Figure~\ref{fig:scope_progression} analyzes how recovery changes as the trainable parameter scope expands.
For both FT and KD, recovery increases as we move from router-only adjustment to router+top-$k$ experts, router+all-experts, and full-parameter adjustment.
This monotonic trend indicates that compression-induced degradation is not confined to a single component.
Router-only adjustment can reduce part of the routing error, and expert-inclusive adjustment can further reduce expert-path or expert-function errors.

The most notable pattern is the large jump from \textsc{Router+All-Expert} adjustment to \textsc{Full} adjustment.
Even though the compression is applied to expert parameters, allowing non-expert dense/shared parameters to update provides a substantial additional gain.
This suggests that local perturbations introduced in MoE layers propagate through the residual stream and affect subsequent attention, normalization, and dense transformations.
Consequently, correcting only the compressed MoE modules may leave downstream propagated noise unresolved.
Full-parameter fine-tuning can compensate for both the local MoE error and its downstream propagation, which explains why it achieves the strongest recovery in Figure~\ref{fig:scope_progression}.

\section{Robustness Check}

\begin{figure}[t]
\centering
\includegraphics[width=\columnwidth]{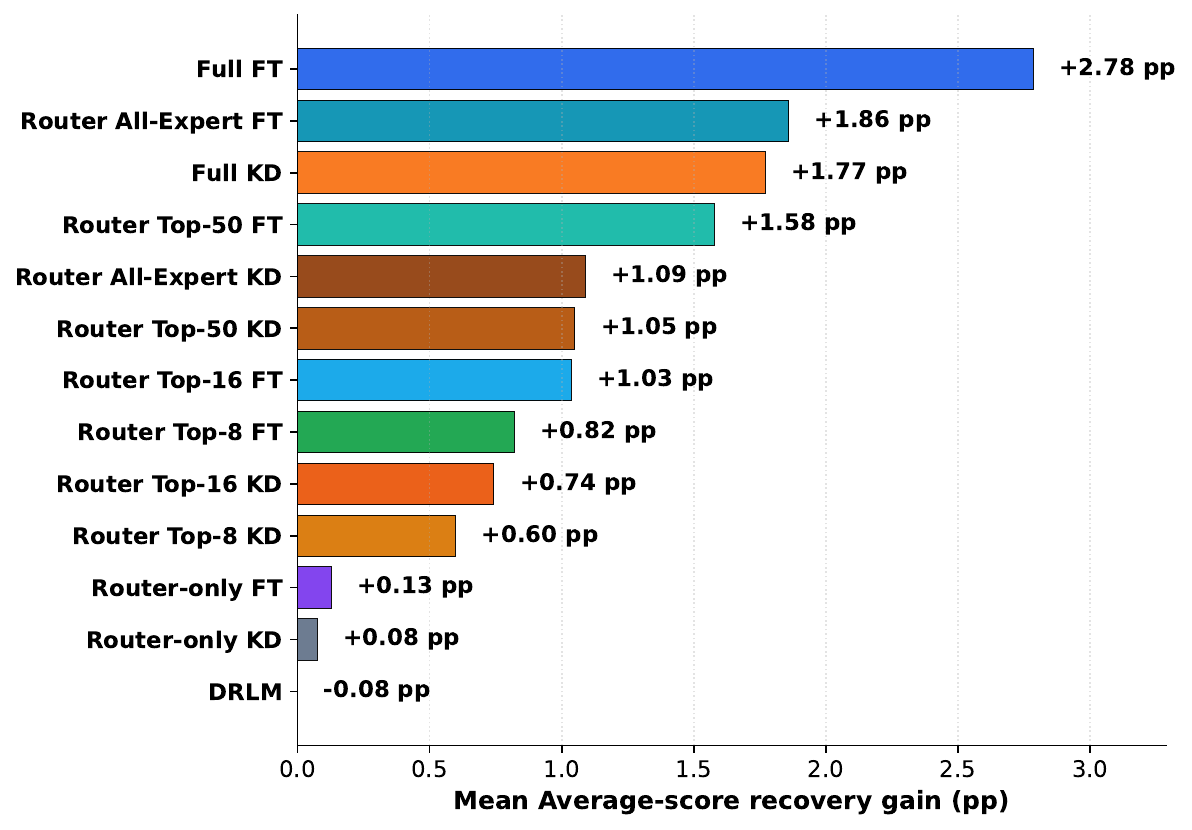}
\caption{Recovery with 1,024 C4 calibration examples. Bars show the mean score change from the unadjusted compressed model across 28 benchmarks and 12 backbone--compressor--retention settings. Full FT remains the strongest strategy, followed by Router All-Expert FT and Full KD, showing that reducing the main study's 3,000-example budget preserves the central strategy-level conclusion.}
\label{fig:small-budget-recovery}
\vspace{-1.2em}
\end{figure}

Reducing the C4 calibration budget from 3,000 to 1,024 examples preserves the main strategy-level pattern (Figure~\ref{fig:small-budget-recovery}). Across four backbone--compressor pairs and three retention ratios, Full FT gives the largest mean recovery ($+2.78$ pp), followed by Router All-Expert FT ($+1.86$ pp) and Full KD ($+1.77$ pp). Causal LM fine-tuning also remains ahead of KD at each matched trainable scope. Thus, the central conclusion is not contingent on the 3,000-example calibration budget used in the main study.

\begin{figure}[t]
\centering
\includegraphics[width=\columnwidth]{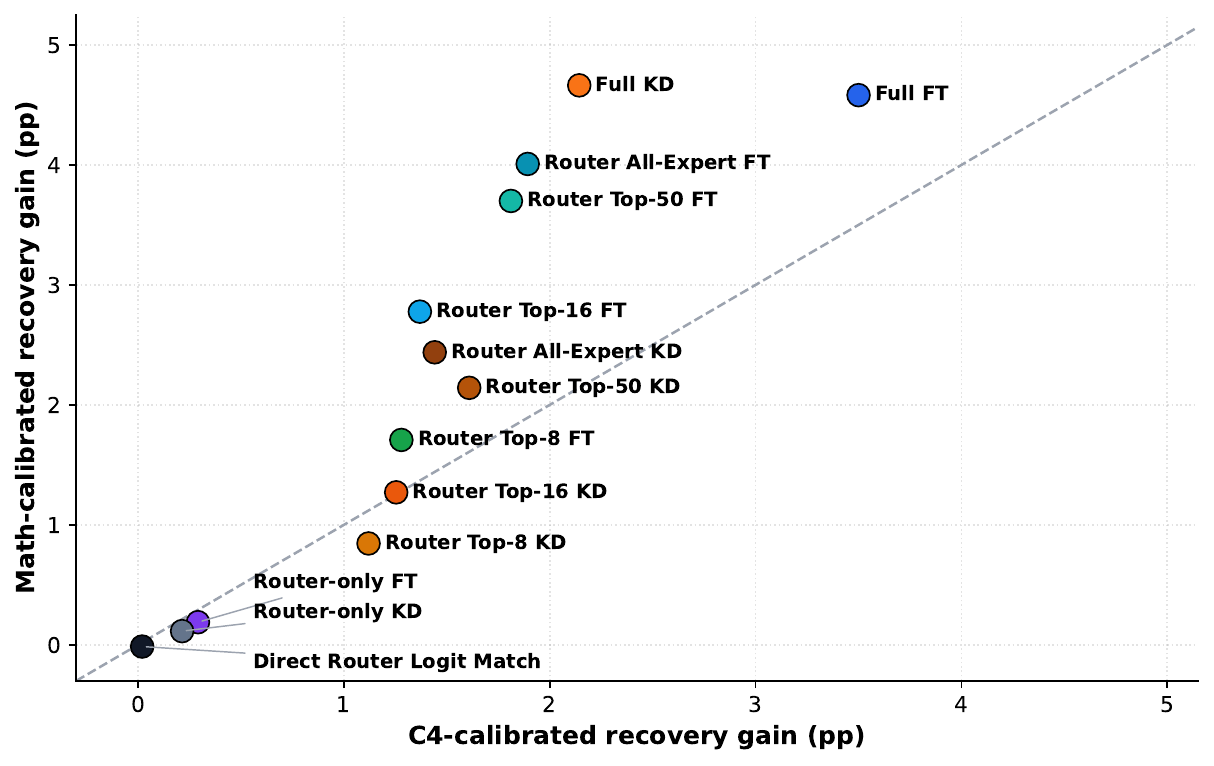}
\caption{Calibration-domain robustness with 1,024 examples each from C4 and OpenR1-Math-220k, averaged over six matched settings and 28 benchmarks. Each point is an adjustment strategy; upper-right indicates stronger recovery in both domains. Full FT has the largest two-domain mean, and aggregate rankings are strongly associated ($\rho=0.973$, $\tau_b=0.897$).}
\label{fig:calibration-domain-consistency}
\vspace{-1.2em}
\end{figure}

We further test calibration-domain sensitivity under matched 1,024-example budgets by replacing C4 with the math-domain OpenR1-Math-220k dataset~\cite{openr1} for six Qwen3/HC-SMoE and Gemma4/AIMER settings, while keeping the 28-task evaluation suite fixed (Figure~\ref{fig:calibration-domain-consistency}). Points farther toward the upper right indicate stronger recovery under both domains. Full FT has the largest recovery averaged across the two domains (4.04 pp, versus 3.40 pp for Full KD). Across the 13 adjustment strategies, C4- and OpenR1-Math-calibrated gains are strongly associated (Spearman $\rho=0.973$; Kendall $\tau_b=0.897$), and Full FT, Full KD, and Router All-Expert FT form the same top-three set. We interpret this as robustness of the aggregate strategy trend, not as invariant ranking in every individual setting; Appendix~\ref{sec:additional-robustness} reports the full aggregate results, costs, and setting-level variation.

\section{Conclusion}

We show that retraining-free MoE compression should be treated as a compressed initialization rather than a final artifact. 
With a tiny general-domain adjustment budget, compressed MoE LLMs can recover a substantial portion of the original-to-compressed performance gap. 
Our results indicate that causal LM fine-tuning, especially full-parameter adjustment, offers the strongest cost--recovery trade-off among the strategies. 
Retraining-free compression is a strong start, and recovery toward the original model can be substantially improved through a small post-compression adjustment stage.



\section*{Limitations}


First, our GPU-cost comparison is controlled but not fully system-optimized.
For fair comparison, the GPU costs reported for post-compression adjustment are measured under a unified \texttt{PyTorch/Transformers}-based implementation for expert selection and adjustment training, using the same hardware, data budget, and monitoring protocol.
Benchmark evaluation is performed separately with \texttt{lm-evaluation-harness} and \texttt{vLLM}, and is not included in the adjustment-training cost.
This design avoids mixing strategy-level effects with implementation-specific optimization differences.
However, each stage could be further optimized with specialized systems.
For example, expert-selection inference may be accelerated with optimized serving engines such as \texttt{vLLM}, while adjustment training may benefit from optimized distributed-training stacks such as DeepSpeed, FSDP/Accelerate, Megatron-style training, or custom fused kernels.
We do not evaluate how such system-level optimizations would change the cost--recovery trade-off. Second, our main experiments focus on Expert Pruning and Expert Merging. 
Expert Editing is an important MoE compression paradigm, but many editing methods rely on native factorized or tensor-decomposed expert representations that are not yet efficiently supported by the same inference and training stack used for standard MoE checkpoints. 
In practice, methods such as MoBE or TD-MoE often provide reconstructed standard-MoE realizations for compatibility with existing serving engines. 
Post-adjusting these reconstructed checkpoints is not parameterization-equivalent to updating the native factorized editing model. 
For this reason, we report Expert Editing results separately in Appendix~\ref{appendix:equations_editing}, and do not use them as primary evidence for deployable compressed-artifact efficiency. Third, our study does not evaluate models above the 100B-parameter scale. Larger MoE models may exhibit different routing behavior, different expert redundancy patterns, and different cost trade-offs. Extending post-compression adjustment to such models is an important direction for future work. Finally, we use a tiny general-domain C4 adjustment budget. Task-specific data, multilingual data, safety-alignment data, or longer adjustment schedules may change which objective and trainable scope is optimal. 

Our conclusions should therefore be interpreted as evidence for a small-budget, general-domain post-compression adjustment regime rather than as a universal optimum for all deployment settings.
\section*{Ethical Considerations}

This work studies post-compression adjustment for MoE LLMs, with the goal of reducing the memory and compute burden required to deploy large models. We do not collect human-subject data, private user data, or personally identifiable information. The main adjustment experiments use a sampled C4 subset, while the calibration-domain robustness check additionally uses 1,024 examples from OpenR1-Math-220k; all evaluations use public benchmark datasets. Appendix~\ref{sec:behavioral-proxies} reports metric-specific behavioral proxies, which we treat as diagnostics rather than as a safety certification. A positive implication of this work is that small post-compression adjustment may make compressed MoE models more accessible to researchers and practitioners with limited GPU resources. At the same time, improving the efficiency of LLM deployment can also lower the cost of deploying models for harmful or unintended applications. Our work does not introduce new safety-alignment methods, and we do not claim that compression or post-compression adjustment improves the safety, fairness, or reliability of the underlying models. Compressed and adjusted models may inherit biases, hallucination behavior, unsafe generations, or misuse risks from their original base models. We therefore encourage practitioners to treat post-compression adjustment as an efficiency technique rather than a safety intervention. Any deployment of compressed MoE LLMs should follow the license and usage restrictions of the original models and datasets, and should be accompanied by appropriate safety evaluation for the intended application domain. Finally, because our experiments involve large GPU workloads, we report effective GPU-hours, GPU-memory occupancy in GiB-hours, and GPU energy in kWh to make the computational cost of different adjustment strategies transparent.

In accordance with the ACL Policy on AI Assistance, we acknowledge the use of Gemini 3.1 Pro\footnote{\url{https://deepmind.google/technologies/gemini/}} to assist with code debugging and writing polishing. All experimental designs, data analyses, and scientific claims presented in this work were verified by the authors.
\section*{Acknowledgments}

This work was supported in part by the National Research Foundation of Korea (NRF) grant (RS-2025-00560762), the Institute of Information \& Communications Technology Planning \& Evaluation (IITP) grants (RS-2025-25442338, RS-2026-25518317, RS-2026-25551943, RS-2021-II211343), the AI Seoul Tech Research Support Program of the Seoul Future Foundation, Samsung Electronics Co., Ltd. (IO250817-13428-01), and the Ministry of Science and ICT under the Advanced GPU Utilization Support Program. The authors thank Samsung MAX Lab for providing research infrastructure. J. Do is with ASRI, Seoul National University.

\bibliography{custom}

\newpage

\appendix
\section{Experiment Settings}
\label{app:exp_settings}

This appendix reports the implementation settings used for compression, post-compression adjustment, and downstream evaluation.  Unless otherwise stated, random seeds were fixed to 42, models were loaded and saved in bfloat16, tokenizer and generation sidecar files were copied from the source checkpoint, and dense non-expert parameters were kept unchanged by the compression method itself.  The reported expert-retention ratios are 50\%, 62.5\%, and 75\%.  For pruning and merging methods, these correspond to 64, 80, and 96 retained or physical routed experts per MoE layer, respectively, for models with 128 routed experts per layer.

\subsection{Backbones and Compression Targets}
\label{app:compression-settings}

We use two instruction-tuned MoE backbones: Qwen3-30B-A3B-Instruct-2507 and gemma-4-26B-A4B-it.  Qwen3 contains 48 sparse MoE blocks with 128 routed experts per block, while the Gemma4 text stack contains 30 routed MoE layers with 128 routed experts per layer.  For all pruning and merging methods, the router, attention layers, embeddings, layer normalizations, and other dense components are preserved; only routed expert capacity is reduced.

\subsection{Post-Compression Adjustment}
\label{app:pca-settings}

All Post-Compression Adjustment runs use the same text-only calibration corpus, optimizer settings, and tokenization settings across backbones, methods, retention ratios, objectives, and trainable scopes.  The adjustment data are the first 3000 non-empty text examples read from the C4.  Examples are tokenized with padding and truncation to 512 tokens.  The dataloader is shuffled with a seed-42 \texttt{torch.Generator}, uses \texttt{drop\_last=false}, and uses zero worker processes.

\begin{table}[h]
\centering
\small
\begin{tabular}{p{0.40\columnwidth} p{0.50\columnwidth}}
\hline
\textbf{Setting} & \textbf{Value} \\
\hline
Calibration samples & 3000 \\
Epochs & 1 \\
Per-device batch size & 2 \\
Gradient accumulation & 4 steps \\
Effective batch size & 8 examples \\
Maximum sequence length & 512 tokens \\
Optimizer & AdamW, \texttt{foreach=false} \\
Learning rate & $5\times10^{-5}$ \\
Weight decay & 0.0 \\
LR scheduler & Linear decay with warmup \\
Warmup ratio & 0.03 \\
Gradient clipping & Global norm 1.0 \\
KD temperature & 1.0 \\
Maximum training steps & Not set; one full epoch is used \\
Logging interval & 10 optimizer steps \\
Model dtype & bfloat16 on GPU; float32 only for CPU fallback \\
Memory cap & 110GiB per GPU for automatic device maps \\
Gradient checkpointing & Disabled \\
Saved shard size & 10GB \\
Determinism & Python, NumPy, PyTorch, CUDA, and dataloader seeds fixed to 42; deterministic algorithms enabled with warnings only \\
\hline
\end{tabular}
\caption{Common Post Compression Adjustment hyperparameters.}
\label{tab:pca-common-hparams}
\end{table}

\paragraph{Objectives.}
For fine-tuning (FT), we optimize the standard causal language-modeling loss, masking padding tokens with label value $-100$.  For knowledge distillation (KD), the teacher is the corresponding uncompressed backbone, either Qwen3-30B-A3B-Instruct-2507 or gemma-4-26B-A4B-it.  The teacher is evaluated in no-gradient mode, and the student is optimized with token-level KL divergence between shifted teacher and student logits at temperature 1.0. Both teacher and student forward passes are included in the monitored
KD training-loop segment and therefore in effective
GPU-hours, GPU-memory GiB-hours, and GPU energy. Direct router logit matching (DRLM) is used as a diagnostic router-only objective when a teacher-to-student router-coordinate mapping is available; it uses the same data, optimizer, epoch count, batch size, gradient accumulation, temperature, and gradient clipping as the other Post-Compression Adjustment objectives.

\paragraph{Trainable scopes.}
The unadjusted baseline is evaluated immediately after compression.  Post-Compression Adjustment then uses the following trainable scopes:

\begin{itemize}
\item \textbf{Router-only:} only MoE router parameters are trainable.
\item \textbf{Router + top-$k$ experts:} routers and a fixed per-layer subset of physical experts are trainable.  The main experiments use $k=8$.  Additional scope ablations use $k=16$ and $k=50$.
\item \textbf{Router + all experts:} routers and all routed expert parameters are trainable while dense non-expert parameters remain frozen.  This is stored as \texttt{router\_top128\_expert\_finetune} and \texttt{router\_top128\_expert\_kd} because each original layer has 128 logical routed experts.
\item \textbf{Full-parameter:} all student parameters are trainable.  This corresponds to \texttt{full\_finetune} and \texttt{full\_kd}.
\end{itemize}

For top-$k$ expert scopes, the selected experts are chosen before training using the same adjustment data.  We count router top-8 assignments, map logical expert IDs to physical expert IDs when the compressed model has a logical-to-physical mapping, and keep the most frequently activated $k$ physical experts per layer. The expert-selection inference is monitored as a separate segment;
whenever a strategy requires selection, its samples and wall-clock
duration are aggregated with those of the subsequent training loop
for effective GPU-hours, GPU-memory GiB-hours, and GPU energy.

\paragraph{Hardware and cost logging.}
All adjustment runs used two NVIDIA H200 140GB GPUs. GPU power draw,
utilization, and \texttt{memory.used} were polled periodically with
\texttt{nvidia-smi} on all assigned GPUs, using a nominal one-second
sleep interval between successive poll attempts. Including query
overhead, the observed median effective polling interval was 1.685
seconds across runs. Let $T_{\mathrm{hr}}$ be the sum, in hours, of
the included measured-segment durations and let $G=2$. Let
$\bar{P}_g$ be the mean power draw of GPU $g$ over its valid retained
readings, and let $\bar{u}$ be the mean utilization percentage over
valid retained GPU readings. At poll $i$, let $\mathcal{G}_i$ denote
the subset of assigned GPUs with a valid retained reading and let
$M_{g,i}$ denote \texttt{memory.used} in GiB. Across the $N_M$ polls
with at least one valid memory reading, we define
$
\bar{M}_{\mathrm{obs}}
=
\frac{1}{N_M}
\sum_{i=1}^{N_M}
\sum_{g\in\mathcal{G}_i}M_{g,i}.$
We report
$$
H_{\mathrm{eff}}
=
G T_{\mathrm{hr}}\frac{\bar{u}}{100},
$$
$$
H_{\mathrm{mem}}
=
T_{\mathrm{hr}}\bar{M}_{\mathrm{obs}},
$$
$$
E_{\mathrm{GPU}}
=
\frac{T_{\mathrm{hr}}}{1000}\sum_{g=1}^{G}\bar{P}_g,
$$
measured in effective GPU-hours, GPU-memory GiB-hours, and kWh, respectively. All three metrics use the same measured-segment boundaries. Metric averages are computed from their valid retained readings; 147 of 704,576 poll attempts (0.021\%) contained only one
of the two assigned GPU rows. Model loading occurs before monitoring and is excluded, whereas resident-model memory during the measured segments is included in \texttt{memory.used}. The monitored KD and DRLM loops include both teacher and student forward passes. For router+top-$k$ strategies, separately monitored expert-selection and
training-loop segments are both included; gaps between measured segments are not counted.

\subsection{Evaluation with \texttt{lm-evaluation-harness}}
\label{app:evaluation-settings}

All downstream evaluations use \texttt{lm-evaluation-harness} with the vLLM backend.  The recorded harness version is \texttt{0.4.13.dev0}, and the recorded Transformers version is \texttt{5.8.1}.  Each evaluation job uses \texttt{model=vllm}, bfloat16 weights, \texttt{tensor\_parallel\_size=1}, \texttt{max\_model\_len=4096}, \texttt{trust\_remote\_code=true}, automatic harness batch sizing, chat-template application, multi-turn few-shot formatting, and sample logging.  The limit is unset, so all task examples are evaluated.  Random, NumPy, PyTorch, and few-shot seeds are all 42.  GPU memory utilization is 0.9 for reasoning, math, chain-of-thought, code, and AIME categories, and 0.8 for the multiple-choice category.

For Gemma4 evaluation, \texttt{enable\_thinking=false} is forced for all categories.  For Qwen3, \texttt{enable\_thinking=false} is explicitly set for the AIME category.  HC-SMoE Qwen checkpoints are evaluated with \texttt{enforce\_eager=true}.  Gemma4 M-SMoE checkpoints are evaluated with \texttt{model\_impl=transformers}, \texttt{enforce\_eager=true}, and \texttt{attention\_backend=TRITON\_ATTN}.  Code tasks are run with \texttt{HF\_ALLOW\_CODE\_EVAL=1} and harness unsafe-code confirmation enabled.  For AIME tasks, generation is non-sampling with \texttt{do\_sample=false}, temperature 0.0, and \texttt{max\_gen\_toks=2780}.  Other categories use the deterministic generation or scoring defaults defined by the corresponding harness task YAML files.

\begin{table*}[t]
\centering
\small
\begin{tabular}{p{0.15\textwidth} p{0.30\textwidth} p{0.45\textwidth}}
\hline
\textbf{Category} & \textbf{Tasks} & \textbf{Few-shot / metric settings} \\
\hline
Reasoning and math & \texttt{gsm8k}, \texttt{gsm8k\_platinum}, \texttt{minerva\_math}, \texttt{bbh\_fewshot}, \texttt{bbh\_zeroshot}, \texttt{coqa} & GSM8K and GSM8K-Platinum use 5-shot; Minerva Math uses 4-shot; BBH few-shot uses 3-shot; BBH zero-shot and CoQA use 0-shot. Primary metrics are exact-match style harness metrics. \\
\hline
Multiple choice & \texttt{medqa\_4options}, \texttt{medmcqa}, \texttt{openbookqa}, \texttt{arc\_easy}, \texttt{winogrande},  \texttt{arc\_challenge}, \texttt{piqa}, \texttt{mmlu}, \texttt{hellaswag} & All are evaluated 0-shot.  Metrics are the harness-provided accuracy and normalized-accuracy variants where available. \\
\hline
Chain-of-thought & \texttt{gsm8k\_cot}, \texttt{gsm8k\_platinum\_cot}, \texttt{gsm8k\_cot\_zeroshot}, \texttt{gsm8k\_platinum\_cot\_zeroshot}, \texttt{bbh\_cot\_fewshot}, \texttt{bbh\_cot\_zeroshot} & GSM8K-CoT and GSM8K-Platinum-CoT use 8-shot; their zero-shot variants use 0-shot.  BBH-CoT few-shot uses 3-shot and BBH-CoT zero-shot uses 0-shot.  Metrics are exact-match style harness metrics. \\
\hline
Code & \texttt{humaneval}, \texttt{mbpp}, \texttt{mbpp\_plus},  \texttt{humaneval\_plus} & HumanEval uses 0-shot.  MBPP and MBPP-Plus use 3-shot.  Metrics are pass@1 style harness metrics. \\
\hline
AIME & \texttt{aime}, \texttt{aime24}, \texttt{aime25} & All are 0-shot.  Generation uses \texttt{max\_gen\_toks=2780}, \texttt{do\_sample=false}, and temperature 0.0.  The metric is exact match. \\
\hline
\end{tabular}
\caption{Downstream evaluation categories and task-level settings.}
\label{tab:evaluation-settings}
\end{table*}

\subsection{Benchmarks}
\label{app:benchmarks}

We evaluate each model on 28 downstream benchmarks covering general reasoning, mathematical reasoning, coding, chain-of-thought reasoning, and multiple-choice question answering. 
For general reasoning and QA, we use BBH~\cite{bbh} in both few-shot and zero-shot settings, CoQA~\cite{coqa}, and HellaSwag~\cite{hellaswag}. 
For mathematical reasoning, we use GSM8K~\cite{gsm8k}, GSM8K Platinum~\cite{gsm8k_Platinum}, MATH~\cite{hendrycksmath2021,lewkowycz2022solving,kydlicek2025fixing}, AIME 1983--2024~\cite{aime_1983_2024}, and AIME 2025~\cite{aime_2025}. 
For coding, we evaluate MBPP~\cite{mbpp}, HumanEval~\cite{humaneval}, MBPP+ \& HumanEval+ \cite{evalplus}. 
We further evaluate chain-of-thought reasoning on BBH, GSM8K, and GSM8K Platinum under both few-shot and zero-shot settings. 
Finally, we evaluate multiple-choice question answering on ARC-Challenge, ARC-Easy,  MedMCQA, MedQA, OpenBookQA, PIQA, WinoGrande, and MMLU~\cite{arc,medmcqa,medqa,OpenBookQA,piqa,WinoGrande,mmlu}.

\subsection{Direct router logit matching}
\label{appendix:drlm}

We also include Direct Router Logit Matching (DRLM) as a diagnostic baseline when router coordinates can be aligned through the compression mapping. 
Let $r_{T,\ell,t}$ and $r_{S,\ell,t}$ be the teacher and student router logits at layer $\ell$ and token position $t$, and let $\Pi_\ell$ map teacher router coordinates to the student router space. 
The objective is
\begin{equation}
\mathcal{L}_{\mathrm{DRLM}}
=
\frac{1}{|\mathcal{I}|}
\sum_{(\ell,t)\in\mathcal{I}}
\left\|
\Pi_\ell(r_{T,\ell,t})
-
r_{S,\ell,t}
\right\|_2^2,
\label{eq:drlm_loss}
\end{equation}
where $\mathcal{I}$ is the set of valid layer-token observations. 
Unlike KD, DRLM directly aligns router behavior rather than output-token distributions. 
We use it only as a diagnostic comparison because it requires a well-defined router-coordinate mapping, which may not exist uniformly across all compression paradigms.

\begin{table*}[t]
\centering
\small
\begin{tabular}{p{0.06\textwidth} p{0.06\textwidth} p{0.085\textwidth} p{0.09\textwidth} p{0.555\textwidth}}
\hline
\textbf{Backbone} & \textbf{Paradigm} & \textbf{Method} & \textbf{Target} & \textbf{Compression settings} \\
\hline
Qwen3 & Pruning & REAP & 128 experts/layer $\rightarrow$ 64, 80, 96 retained experts/layer & Prune fractions 0.50, 0.375, and 0.25.  Observation data used the \texttt{allenai/c4} train split with shuffling, observer batch size 1, 1024 batches per category, maximum sequence length 2048, cosine distance, and seed 42.  Router weights were renormalized after pruning.  The saved configuration records \texttt{prune\_method=reap}, \texttt{record\_pruning\_metrics\_only=true}, \texttt{renormalize\_router\_weights=true}, \texttt{expert\_sim=ttm}, \texttt{frequency\_penalty=true}, agglomerative clustering metadata, and average linkage. \\
\hline
Qwen3 & Merging & HC-SMoE & 128 logical experts/layer $\rightarrow$ 64, 80, 96 physical experts/layer & Hierarchical expert merging was run on the RTE calibration task with \texttt{n\_sentences=8}, train batch size 2, \texttt{data\_limit=1000000}, \texttt{partition=8}, \texttt{start\_layer=0}, and \texttt{group\_limit=4}.  Experts were grouped by expert-output activation similarity using hierarchical clustering with average linkage and silhouette stopping; ZipIt was used as the merge operator with \texttt{ingredient=act} and \texttt{dominant=no}. \\
\hline
Qwen3 & Editing & MoBE & Ratio labels 50, 62.5, 75 & MoBE factor fitting was run over all 48 layers and 128 experts with expert matrix shape $768 \times 2048$, SiLU activation, 8000 fitting epochs, batch size 32, 4 batches, and learning rate 0.07.  The fitted matrices were \texttt{gate\_proj} and \texttt{up\_proj}.  The basis count \texttt{num\_B} was 8, 8, and 32 for the 50\%, 62.5\%, and 75\% settings; truncation was 438, 768, and 768, respectively.  Native MoBE checkpoints were reconstructed to Hugging Face checkpoints for vLLM evaluation. \\
\hline
Gemma4 & Pruning & AIMER & 128 experts/layer $\rightarrow$ 64, 80, 96 retained experts/layer & AIMER pruning was calibration-free.  The score was $\mathrm{mean}(|w|)/\mathrm{rms}(w)$ computed over each routed expert's gate, up, and down weights.  Runs used \texttt{metric=aimer}, \texttt{metric\_device=auto}, \texttt{prune\_highest\_score=true}, seed 42, bfloat16 loading, \texttt{device\_map=auto}, \texttt{trust\_remote\_code=true}, and \texttt{model\_class=causal-lm}.  The dense Gemma4 \texttt{layer.mlp} shared component was preserved and not treated as part of AIMER pruning. \\
\hline
Gemma4 & Merging & M-SMoE & 128 logical experts/layer $\rightarrow$ 64, 80, 96 physical experts/layer & M-SMoE used the shared C4 calibration shard \texttt{c4-train.00000-of-01024.json}.  Router statistics were collected with \texttt{similarity\_base=router-logits}, \texttt{subset\_ratio=0.01}, block size 512, and batch size 1.  The implementation confirms 128 logical Gemma4 experts and rewrites the 30 routed layers to the requested physical expert count. \\
\hline
Gemma4 & Editing & TD-MoE & Target total parameters 14.387B, 17.242B, 20.096B & TD-MoE was applied to all 30 Gemma4 routed layers using the same C4 calibration shard, 256 whitening samples, model sequence length 2048, seed 3, input whitening, SVD decomposition, bfloat16 inference, and \texttt{device\_map=auto}.  Native TD-MoE checkpoints were materialized back to Hugging Face/vLLM-compatible checkpoints with bfloat16 weights and 10GB maximum shard size. \\
\hline
\end{tabular}
\caption{Compression settings.}
\label{tab:compression-settings}
\end{table*}

\vspace*{\fill}
\null

\newpage

\vspace*{\fill}
\null

\newpage

\vspace*{\fill}
\null

\newpage

\newpage

\section{Original MoE (before Compression)}
\label{appendix:equations1}

We adopt the MoE notation of
\citet{hyeon2026retraining}. We restate only the definitions
needed for the residual decompositions below; the underlying MoE formulation
and the original scenario analysis are not claimed as new contributions of
the present paper.

Throughout Appendices~B--D, $x$ denotes a common local hidden-state input
supplied to the original and compressed layer maps. This same-input convention
isolates the local compression source. The end-to-end input shift between the
two networks is handled separately by the propagation analysis in
Section~\ref{sec:degradation}.

Consider one sparse MoE layer with routed-expert index set
$\mathcal{E}=\{0,\ldots,N-1\}$. Let
$\mathcal{S}\subseteq\mathcal{E}$ be the top-$k$ active set selected by the
original router, with $|\mathcal{S}|=k<N$. Let
$g_i^{\mathrm{orig}}(x)\ge 0$ denote the nonnegative gate mass used for
mixture weighting, such as a softmax probability. This avoids normalizing raw
router logits, which need not be nonnegative. We assume that the gate masses
are positive on the selected set.

For any nonempty active set $\mathcal{A}\subseteq\mathcal{E}$, define
\begin{equation}
\widetilde g_i^{o,\mathcal{A}}(x)
:=
\frac{g_i^{\mathrm{orig}}(x)}
     {\sum_{j\in\mathcal{A}}g_j^{\mathrm{orig}}(x)},
\qquad i\in\mathcal{A}.
\label{eq:appendix-original-renormalized-gate}
\end{equation}
The routed MoE contribution is
\begin{equation}
y_{\mathrm{orig}}(x)
=
\sum_{i\in\mathcal{S}}
\widetilde g_i^{o,\mathcal{S}}(x)E_i(x).
\label{eq:appendix-original-output}
\end{equation}
Any unchanged shared-expert, residual, or dense branch is omitted from this
local expression because it cancels in the same-input difference. In the
following appendices, superscripts $p$, $m$, and $e$ denote the pruned,
merged, and edited models, respectively.

\section{Expert Pruning ($N \rightarrow N-\alpha$)}
\label{appendix:equations_pruning}

The exhaustive best/partial/disjoint pruning analysis was introduced in prior work~\cite{hyeon2026retraining}. Here we replace the repeated scenario-wise algebra with one exact active-path decomposition that directly supports the routing-versus-expert-path interpretation in Section~\ref{sec:degradation}.

Let $\mathcal{P}\subseteq\mathcal{E}$ be the set of retained experts, with $|\mathcal{P}|\ge k$, and let $\mathcal{S}'\subseteq\mathcal{P}$ be the active set selected by the pruned layer at the common reference input $x$. We assume $|\mathcal{S}|=|\mathcal{S}'|=k$. For any nonempty active set $A\subseteq P$ satisfying $\sum_{j\in A}g_j^{\mathrm{pruned}}(x)>0$, define
\begin{equation}
\widetilde g_i^{p,\mathcal{A}}(x)
:=
\frac{g_i^{\mathrm{pruned}}(x)}
     {\sum_{j\in\mathcal{A}}g_j^{\mathrm{pruned}}(x)},
\qquad i\in\mathcal{A}.
\label{eq:appendix-pruned-renormalized-gate}
\end{equation}
The two routed outputs are
\begin{align}
y_{\mathrm{orig}}(x)
&=
\sum_{i\in\mathcal{S}}
\widetilde g_i^{o,\mathcal{S}}(x)E_i(x),
\label{eq:appendix-pruning-original-output}
\\
y_{\mathrm{pruned}}(x)
&=
\sum_{i\in\mathcal{S}'}
\widetilde g_i^{p,\mathcal{S}'}(x)E_i(x).
\label{eq:appendix-pruning-compressed-output}
\end{align}
For compact displays, write
\begin{equation}
\widetilde g_i^{o}
:=
\widetilde g_i^{o,\mathcal{S}}(x),
\quad
\widetilde g_i^{p}
:=
\widetilde g_i^{p,\mathcal{S}'}(x),
\quad
E_i:=E_i(x).
\label{eq:appendix-pruning-display-shorthand}
\end{equation}

\paragraph{Active-set decomposition.}
The availability set $\mathcal{P}$ and the active set $\mathcal{S}'$ have
different roles. Define
\begin{equation}
\begin{aligned}
\mathcal{T} &= \mathcal{S}\cap\mathcal{S}',
&\mathcal{D} &= \mathcal{S}\setminus\mathcal{S}',
&\mathcal{R} &= \mathcal{S}'\setminus\mathcal{S}.
\end{aligned}
\label{eq:appendix-pruning-set-decomposition}
\end{equation}
Then
\begin{equation}
\mathcal{S}
=
\mathcal{T}\mathbin{\dot\cup}\mathcal{D},
\qquad
\mathcal{S}'
=
\mathcal{T}\mathbin{\dot\cup}\mathcal{R}.
\label{eq:appendix-pruning-set-partitions}
\end{equation}
Moreover,
\begin{equation}
\begin{aligned}
\mathcal{D}
&=
(\mathcal{S}\setminus\mathcal{P})
\mathbin{\dot\cup}
\bigl((\mathcal{S}\cap\mathcal{P})
      \setminus\mathcal{S}'\bigr).
\end{aligned}
\label{eq:appendix-pruning-missing-path-sources}
\end{equation}
The first subset contains physically removed experts. The second contains
retained experts absent from the compressed active path. In a realized
end-to-end comparison, the latter can arise from an upstream hidden-state
shift even when the router parameters themselves are unchanged.

Using Eq.~\eqref{eq:appendix-pruning-set-partitions}, the exact vector
residual is
\begin{equation}
\begin{aligned}
y_{\mathrm{orig}}-y_{\mathrm{pruned}}
&=
\sum_{i\in\mathcal{T}}
(\widetilde g_i^{o}-\widetilde g_i^{p})E_i
\\
&\quad+
\sum_{i\in\mathcal{D}}
\widetilde g_i^{o}E_i
-
\sum_{i\in\mathcal{R}}
\widetilde g_i^{p}E_i.
\end{aligned}
\label{eq:appendix-pruning-exact-residual}
\end{equation}
Define
\begin{align}
e_{\mathrm{route}}^{p}(x)
&:=
\sum_{i\in\mathcal{T}}
(\widetilde g_i^{o}-\widetilde g_i^{p})E_i,
\label{eq:appendix-pruning-routing-error}
\\
e_{\mathrm{path}}^{p}(x)
&:=
\sum_{i\in\mathcal{D}}
\widetilde g_i^{o}E_i
-
\sum_{i\in\mathcal{R}}
\widetilde g_i^{p}E_i.
\label{eq:appendix-pruning-path-error}
\end{align}
For any vector norm,
\begin{equation}
\begin{aligned}
\lVert y_{\mathrm{orig}}-y_{\mathrm{pruned}}\rVert
&=
\lVert e_{\mathrm{route}}^{p}
      +e_{\mathrm{path}}^{p}\rVert
\\
&\le
\lVert e_{\mathrm{route}}^{p}\rVert
+
\lVert e_{\mathrm{path}}^{p}\rVert.
\end{aligned}
\label{eq:appendix-pruning-residual-norm}
\end{equation}

\paragraph{Best scenario: preserved active path.}
The favorable condition is $\mathcal{S}'=\mathcal{S}$, which implies
$\mathcal{S}\subseteq\mathcal{P}$. Then
$\mathcal{T}=\mathcal{S}$ and
$\mathcal{D}=\mathcal{R}=\emptyset$, so
\begin{equation}
\begin{aligned}
y_{\mathrm{orig}}-y_{\mathrm{pruned}}
&=
e_{\mathrm{route}}^{p}(x)
\\
&=
\sum_{i\in\mathcal{S}}
(\widetilde g_i^{o}-\widetilde g_i^{p})E_i.
\end{aligned}
\label{eq:appendix-pruning-best}
\end{equation}
The normalized router weights can differ if the gate implementation is changed or after the router itself is adjusted. For pure expert pruning that leaves the router and gate implementation unchanged, however, the common-input condition together with $S'=S$ implies $\tilde g_i^{p,S}(x)=\tilde g_i^{o,S}(x)$ for every $i\in S$; hence the initial same-input residual in Eq.~\ref{eq:appendix-pruning-best} is exactly zero. Realized end-to-end deviations can still occur because upstream compression changes the hidden state supplied to deeper routers. More generally, Eq.~\ref{eq:appendix-pruning-best} is the only active-set case whose local residual has no expert-path replacement term.

\paragraph{Most common scenario: partial active-set overlap.}
If $0<|\mathcal{T}|<k$, then
$\mathcal{D}\neq\emptyset$ and $\mathcal{R}\neq\emptyset$, with
$|\mathcal{D}|=|\mathcal{R}|=k-|\mathcal{T}|$. Both residual components may
contribute. Router-only adjustment can change shared-path weights and may move
the top-$k$ boundary, but it cannot directly change the expert functions.

\paragraph{Worst scenario: disjoint active paths.}
If $\mathcal{T}=\emptyset$, then
\begin{equation}
\begin{aligned}
y_{\mathrm{orig}}-y_{\mathrm{pruned}}
&=
e_{\mathrm{path}}^{p}(x)
\\
&=
\sum_{i\in\mathcal{S}}
\widetilde g_i^{o}E_i
-
\sum_{i\in\mathcal{S}'}
\widetilde g_i^{p}E_i.
\end{aligned}
\label{eq:appendix-pruning-disjoint}
\end{equation}
Here, ``worst'' denotes complete active-set replacement. Disjointness alone
does not prove that the residual norm is maximal for every input.

\paragraph{Sign convention for the propagation analysis.}
Equations~\eqref{eq:appendix-pruning-exact-residual}--
\eqref{eq:appendix-pruning-path-error} use the original-minus-compressed
convention. Because Section~\ref{sec:degradation} defines
$\delta_\ell=\widehat h_\ell-h_\ell$, the same-input source injected into
that recurrence is
\begin{equation}
\epsilon_\ell^{p}
:=
y_{\mathrm{pruned},\ell}-y_{\mathrm{orig},\ell}
=
-\left(
e_{\mathrm{route},\ell}^{p}
+
e_{\mathrm{path},\ell}^{p}
\right).
\label{eq:appendix-pruning-propagation-sign}
\end{equation}

\section{Expert Merging ($N \rightarrow M$, where $M<N$)}
\label{appendix:equations_merging}

We adopt the cluster mapping and projected-path notation introduced in
\citet[Appendix~D]{hyeon2026retraining}. Rather than repeating the prior
nine-case enumeration, we give one cluster-space identity tailored to the
present adjustment-scope analysis.

Let $\mathcal{C}=\{0,\ldots,M-1\}$ be the physical merged-expert index set,
and let
\begin{equation}
\phi:\mathcal{E}\rightarrow\mathcal{C}
\label{eq:appendix-merging-map}
\end{equation}
be a surjective map assigning each original routed expert to a merged
cluster. The function implemented by physical merged expert $c$ is denoted
by $M_c(x)$. We do not assume that $M_c$ is exactly equal to every original
expert in its preimage; it is the synthesized function produced by the
merging method.

For the original active set $\mathcal{S}$, define
\begin{equation}
\mathcal{C}_{\mathrm{proj}}
:=
\phi(\mathcal{S})
=
\{\phi(i):i\in\mathcal{S}\}.
\label{eq:appendix-merging-projected-set}
\end{equation}
Let
$\mathcal{S}_{m}:=\mathcal{S}^{\mathrm{merge}}\subseteq\mathcal{C}$ be the
set of physical merged experts receiving nonzero active gate mass. This
covers both routers with $M$ physical coordinates and routers that retain
$N$ logical coordinates and map several active logical indices to one
physical expert. In the latter case, $g_c^{\mathrm{merge}}(x)$ is the sum of
the active logical gate masses mapped to $c$.

For any nonempty active set $A\subseteq\mathcal C$ satisfying $\sum_{d\in A}g_d^{\mathrm{merge}}(x)>0$, define
\begin{equation}
\widetilde g_c^{m,\mathcal{A}}(x)
:=
\frac{g_c^{\mathrm{merge}}(x)}
     {\sum_{d\in\mathcal{A}}g_d^{\mathrm{merge}}(x)},
\qquad c\in\mathcal{A}.
\label{eq:appendix-merged-renormalized-gate}
\end{equation}
The merged output is
\begin{equation}
y_{\mathrm{merge}}(x)
=
\sum_{c\in\mathcal{S}_{m}}
\widetilde g_c^{m,\mathcal{S}_{m}}(x)M_c(x).
\label{eq:appendix-merged-output}
\end{equation}

\paragraph{Projecting the original output into cluster space.}
For each $c\in\mathcal{C}_{\mathrm{proj}}$, define the original gate mass
assigned to that cluster by
\begin{equation}
\alpha_c^{o,\mathcal{S}}(x)
:=
\sum_{\substack{i\in\mathcal{S}\\\phi(i)=c}}
\widetilde g_i^{o,\mathcal{S}}(x).
\label{eq:appendix-merging-original-cluster-mass}
\end{equation}
Because the selected gate masses are positive,
$\alpha_c^{o,\mathcal{S}}(x)>0$. Define the corresponding gate-weighted
original cluster output by
\begin{equation}
\overline E_c^{o,\mathcal{S}}(x)
:=
\frac{
\sum_{\substack{i\in\mathcal{S}\\\phi(i)=c}}
\widetilde g_i^{o,\mathcal{S}}(x)E_i(x)
}{
\alpha_c^{o,\mathcal{S}}(x)
}.
\label{eq:appendix-merging-original-cluster-output}
\end{equation}
Then
\begin{equation}
y_{\mathrm{orig}}(x)
=
\sum_{c\in\mathcal{C}_{\mathrm{proj}}}
\alpha_c^{o,\mathcal{S}}(x)
\overline E_c^{o,\mathcal{S}}(x).
\label{eq:appendix-merging-original-output-cluster-space}
\end{equation}
For compact displays below, suppress $x$ and write
\begin{equation}
\begin{aligned}
\alpha_c^{o}&:=\alpha_c^{o,\mathcal{S}}(x),
&\overline E_c^{o}&:=\overline E_c^{o,\mathcal{S}}(x),\\
\widetilde g_c^{m}&:=
\widetilde g_c^{m,\mathcal{S}_{m}}(x),
&M_c&:=M_c(x).
\end{aligned}
\label{eq:appendix-merging-display-shorthand}
\end{equation}

Define the shared, missing, and newly active physical clusters by
\begin{equation}
\begin{aligned}
\mathcal{T}
&=
\mathcal{C}_{\mathrm{proj}}\cap\mathcal{S}_{m},\\
\mathcal{D}
&=
\mathcal{C}_{\mathrm{proj}}\setminus\mathcal{S}_{m},\\
\mathcal{R}
&=
\mathcal{S}_{m}\setminus\mathcal{C}_{\mathrm{proj}}.
\end{aligned}
\label{eq:appendix-merging-cluster-sets}
\end{equation}
The exact residual is
\begin{equation}
\begin{aligned}
y_{\mathrm{orig}}-y_{\mathrm{merge}}
&=
\sum_{c\in\mathcal{T}}
\left(
\alpha_c^{o}\overline E_c^{o}
-
\widetilde g_c^{m}M_c
\right)
\\
&\quad+
\sum_{c\in\mathcal{D}}
\alpha_c^{o}\overline E_c^{o}
-
\sum_{c\in\mathcal{R}}
\widetilde g_c^{m}M_c.
\end{aligned}
\label{eq:appendix-merging-exact-residual}
\end{equation}
For every shared cluster $c\in\mathcal{T}$,
\begin{equation}
\begin{aligned}
\alpha_c^{o}\overline E_c^{o}
-
\widetilde g_c^{m}M_c
&=
(\alpha_c^{o}-\widetilde g_c^{m})M_c
\\
&\quad+
\alpha_c^{o}(\overline E_c^{o}-M_c).
\end{aligned}
\label{eq:appendix-merging-shared-cluster-identity}
\end{equation}
This yields
\begin{align}
e_{\mathrm{route}}^{m}(x)
&:=
\sum_{c\in\mathcal{T}}
(\alpha_c^{o}-\widetilde g_c^{m})M_c,
\label{eq:appendix-merging-routing-error}
\\
e_{\mathrm{approx}}^{m}(x)
&:=
\sum_{c\in\mathcal{T}}
\alpha_c^{o}(\overline E_c^{o}-M_c),
\label{eq:appendix-merging-approximation-error}
\\
e_{\mathrm{path}}^{m}(x)
&:=
\sum_{c\in\mathcal{D}}
\alpha_c^{o}\overline E_c^{o}
-
\sum_{c\in\mathcal{R}}
\widetilde g_c^{m}M_c.
\label{eq:appendix-merging-path-error}
\end{align}
Consequently,
\begin{equation}
y_{\mathrm{orig}}-y_{\mathrm{merge}}
=
e_{\mathrm{route}}^{m}
+
e_{\mathrm{approx}}^{m}
+
e_{\mathrm{path}}^{m}.
\label{eq:appendix-merging-three-way-residual}
\end{equation}
To match the two-source interpretation in Section~\ref{sec:degradation},
define
\begin{equation}
e_{\mathrm{expert/path}}^{m}
:=
e_{\mathrm{approx}}^{m}+e_{\mathrm{path}}^{m}.
\label{eq:appendix-merging-expert-path-error}
\end{equation}
Then
\begin{equation}
\lVert y_{\mathrm{orig}}-y_{\mathrm{merge}}\rVert
=
\left\lVert
e_{\mathrm{route}}^{m}
+
e_{\mathrm{expert/path}}^{m}
\right\rVert.
\label{eq:appendix-merging-two-source-residual}
\end{equation}

\paragraph{Full projected-path coverage.}
If $\mathcal{D}=\emptyset$, every cluster implied by the original active set
is selected. If also $\mathcal{R}=\emptyset$, only router reweighting and
merged-function approximation remain. If $\mathcal{R}\neq\emptyset$, every
required cluster is covered but additional clusters are activated.

\paragraph{Partial projected-path coverage.}
If $\mathcal{T}\neq\emptyset$ and $\mathcal{D}\neq\emptyset$, all three terms
in Eq.~\eqref{eq:appendix-merging-three-way-residual} may be nonzero.

\paragraph{Disjoint projected paths.}
If $\mathcal{T}=\emptyset$, the shared-cluster routing and approximation terms
vanish under this decomposition, and the discrepancy is represented by
$e_{\mathrm{path}}^{m}$. Disjointness does not prove a universally maximal
residual norm.

\paragraph{Capacity-limited coverage.}
Let $k_{\mathrm{phys}}$ be the maximum number of distinct physical merged
experts that can be active for one token. If
$|\mathcal{C}_{\mathrm{proj}}|>k_{\mathrm{phys}}$, then
$\mathcal{D}\neq\emptyset$ for every feasible $\mathcal{S}_{m}$. Thus, part
of the projected original path is structurally unrepresentable by the merged
routing capacity.

Router-only adjustment can alter physical cluster gate masses and may change
$\mathcal{S}_{m}$, but it cannot directly make $M_c$ equal to the
input-dependent original cluster output $\overline E_c^{o}$. Expert-inclusive
adjustment can also change the merged functions, while full-parameter
adjustment can further compensate for their downstream effects.

\paragraph{Sign convention for the propagation analysis.}
Under $\delta_\ell=\widehat h_\ell-h_\ell$, the same-input merged-model source
is
\begin{equation}
\begin{aligned}
\epsilon_\ell^{m}
:=
y_{\mathrm{merge},\ell}-y_{\mathrm{orig},\ell}
\\
=-\left(
e_{\mathrm{route},\ell}^{m}
+
e_{\mathrm{expert/path},\ell}^{m}
\right).
\label{eq:appendix-merging-propagation-sign}
\end{aligned}
\end{equation}

\newpage

\section{Experiment Results by Compression Paradigm}
\label{app:exp_per_methods}

This appendix breaks down the main results by compression paradigm. 
While Section~\ref{sec:results} reports averages over all pruning and merging checkpoints, here we separately analyze Expert Pruning and Expert Merging to verify that the observed trends are not driven by a single compression family. 
Across both paradigms, the same qualitative pattern appears: causal LM fine-tuning is generally stronger than KD under matched trainable scopes, router-only adjustment provides only limited recovery, and full-parameter fine-tuning achieves the largest recovery gain.

\subsection{Expert Pruning}

\begin{figure}[h]
    \centering 
    \includegraphics[width=\linewidth]{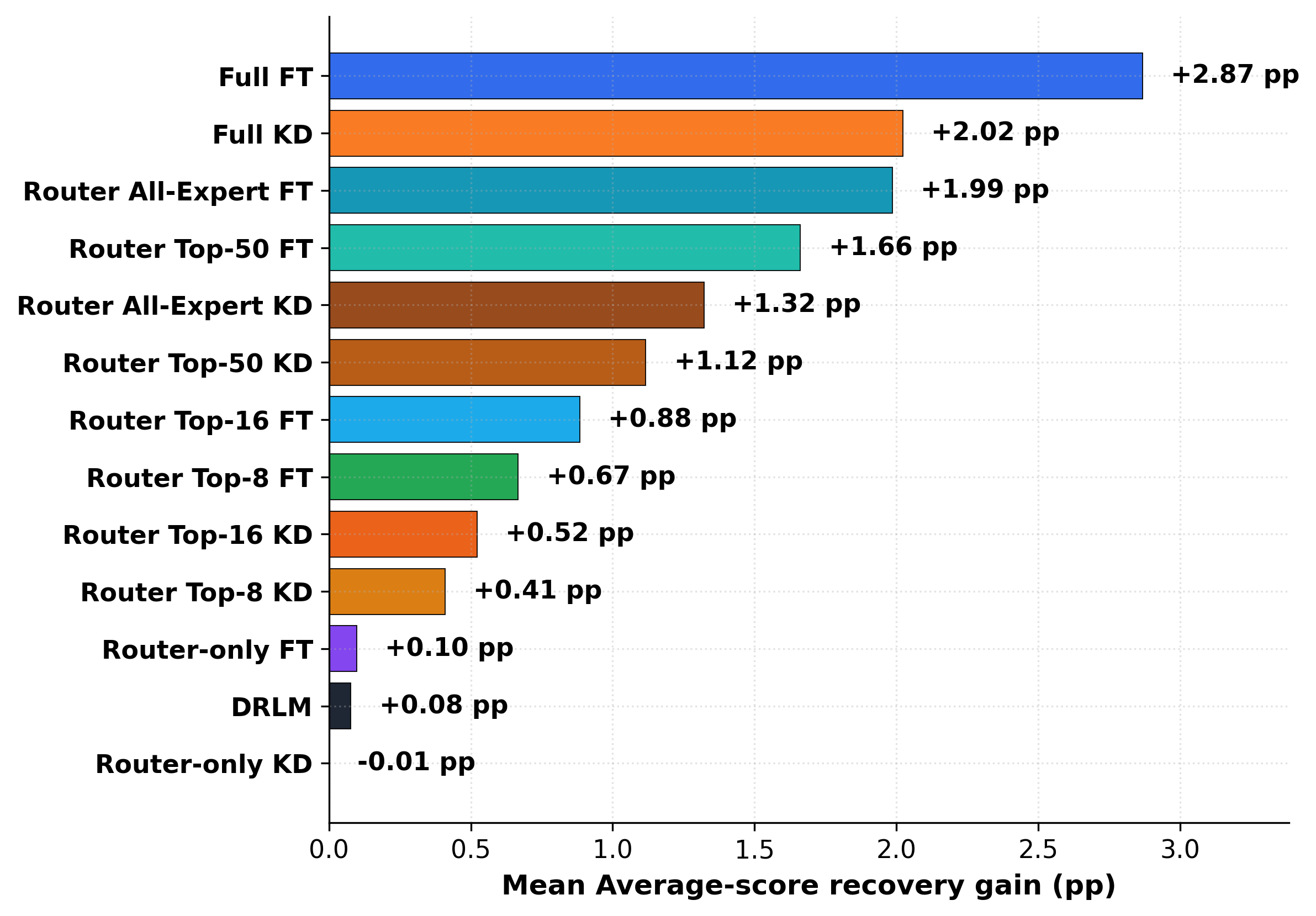}
    \caption{
Expert Pruning: Overall recovery gain of post-compression adjustment strategies.
Bars show mean average-score recovery gain, measured in percentage points (pp), over the unadjusted pruning baseline.
Results are averaged over pruning checkpoints, retention ratios, and benchmarks.
\textsc{Full FT} achieves the largest recovery, and causal LM fine-tuning outperforms KD under matched trainable scopes.
}
    \label{fig:overall_recovery_prune} 
\end{figure}

\begin{figure}[!t]
    \centering
    \includegraphics[width=0.95\linewidth]{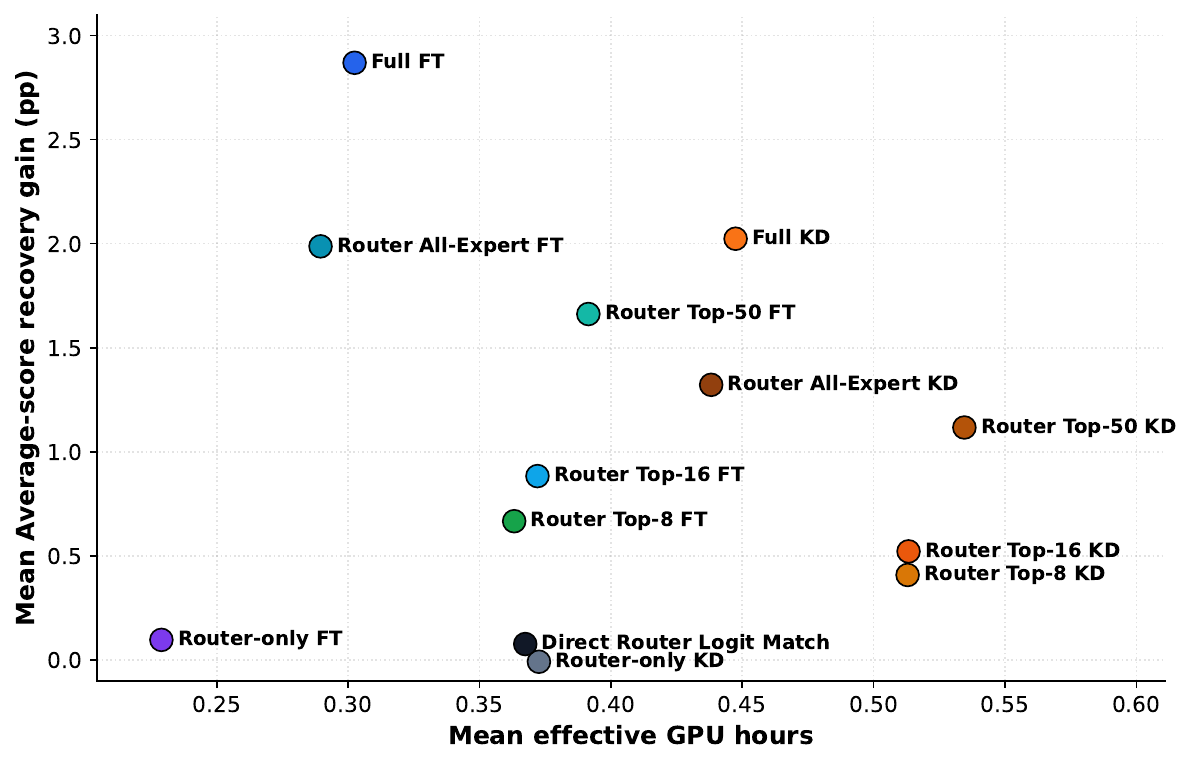}\\[1ex]
    \includegraphics[width=0.95\linewidth]{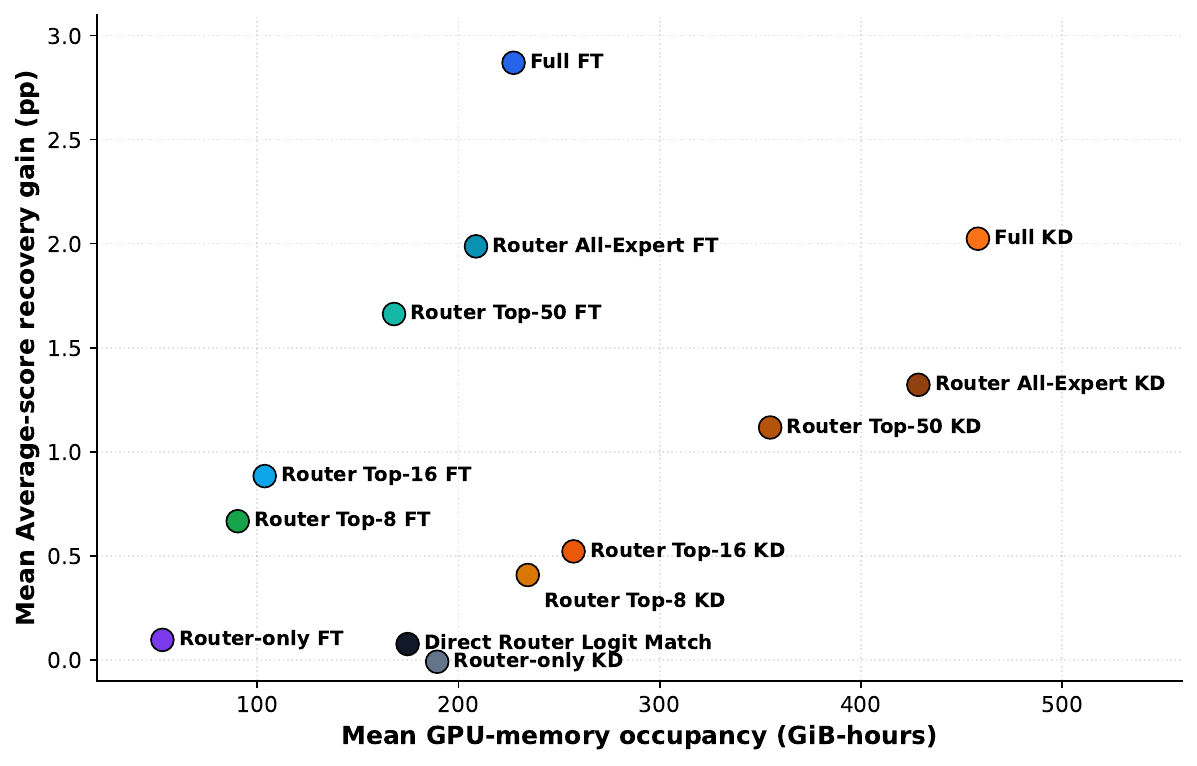}\\[1ex]
    \includegraphics[width=0.95\linewidth]{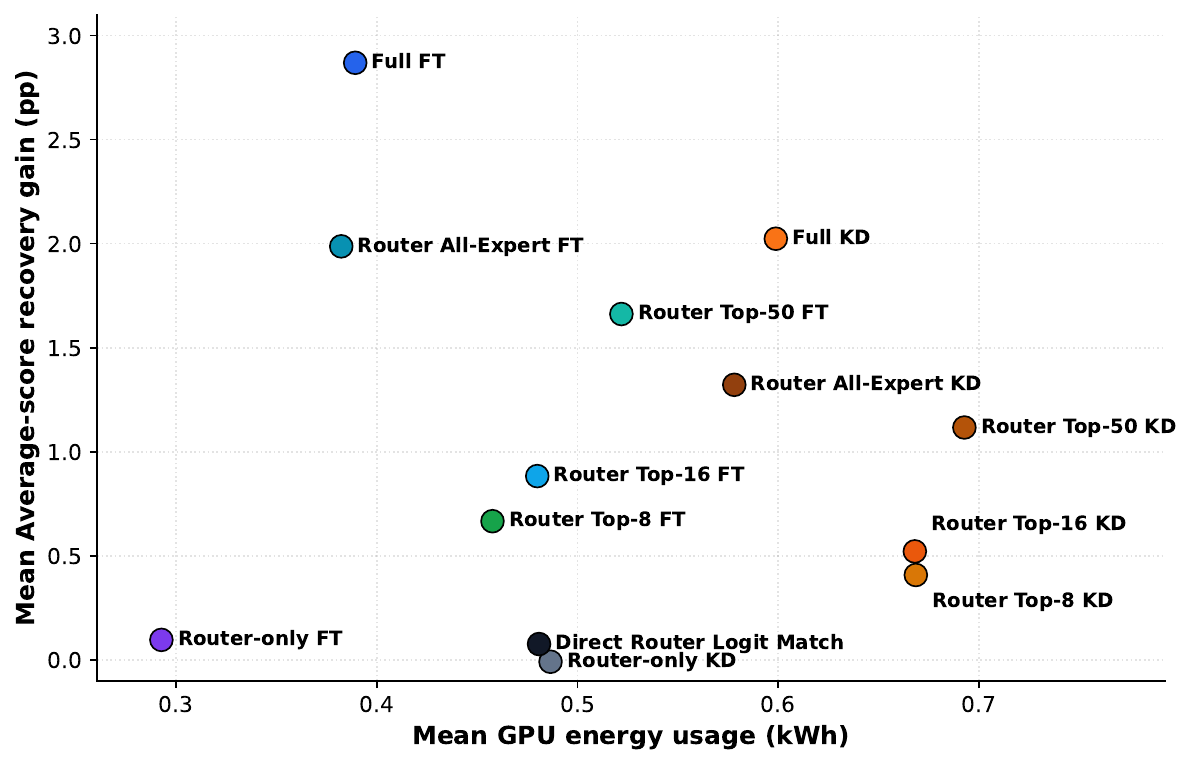}
    
   \caption{Expert Pruning: GPU cost--performance trade-off of
post-compression adjustment strategies. The y-axis shows mean
average-score recovery gain, while the x-axis shows effective
GPU-hours (top), GPU-memory occupancy in GiB-hours (middle), and GPU
energy in kWh (bottom). Each point is an unweighted arithmetic mean
over six pruning method--ratio settings (two pruning methods $\times$
three retention ratios). Both expert-selection cost and training-loop
cost are included when applicable. Upper-left is better.}
    \label{fig:cost_tradeoff_prune}
\end{figure}

\begin{figure}[t]
    \centering 
    \includegraphics[width=\linewidth]{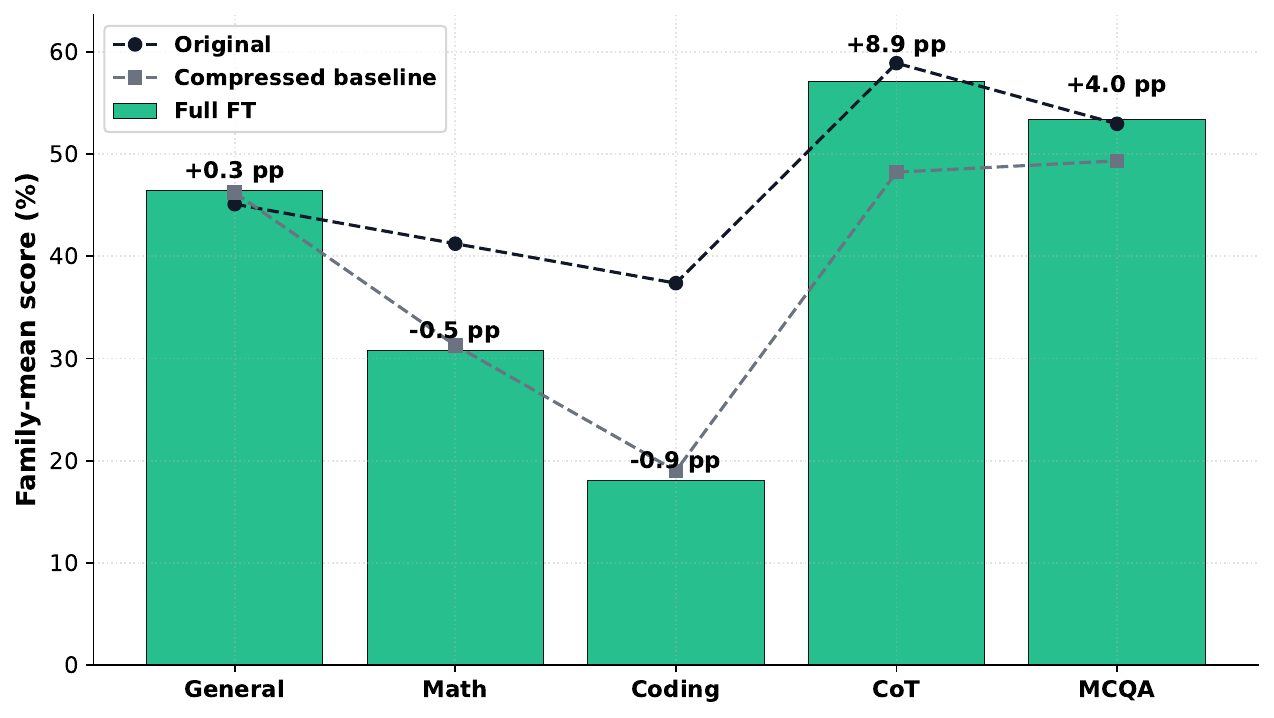}
    \caption{
Expert Pruning: Benchmark-family recovery of \textsc{Full FT}.
Bars show post-adjustment performance, and dashed lines denote the original model and the unadjusted pruning baseline.
\textsc{Full FT} yields gains on CoT, MCQA, and general benchmarks, while math and coding decrease slightly.
}
    \label{fig:family_recovery_prune} 
\end{figure}

\begin{figure}[h]
    \centering 
    \includegraphics[width=\linewidth]{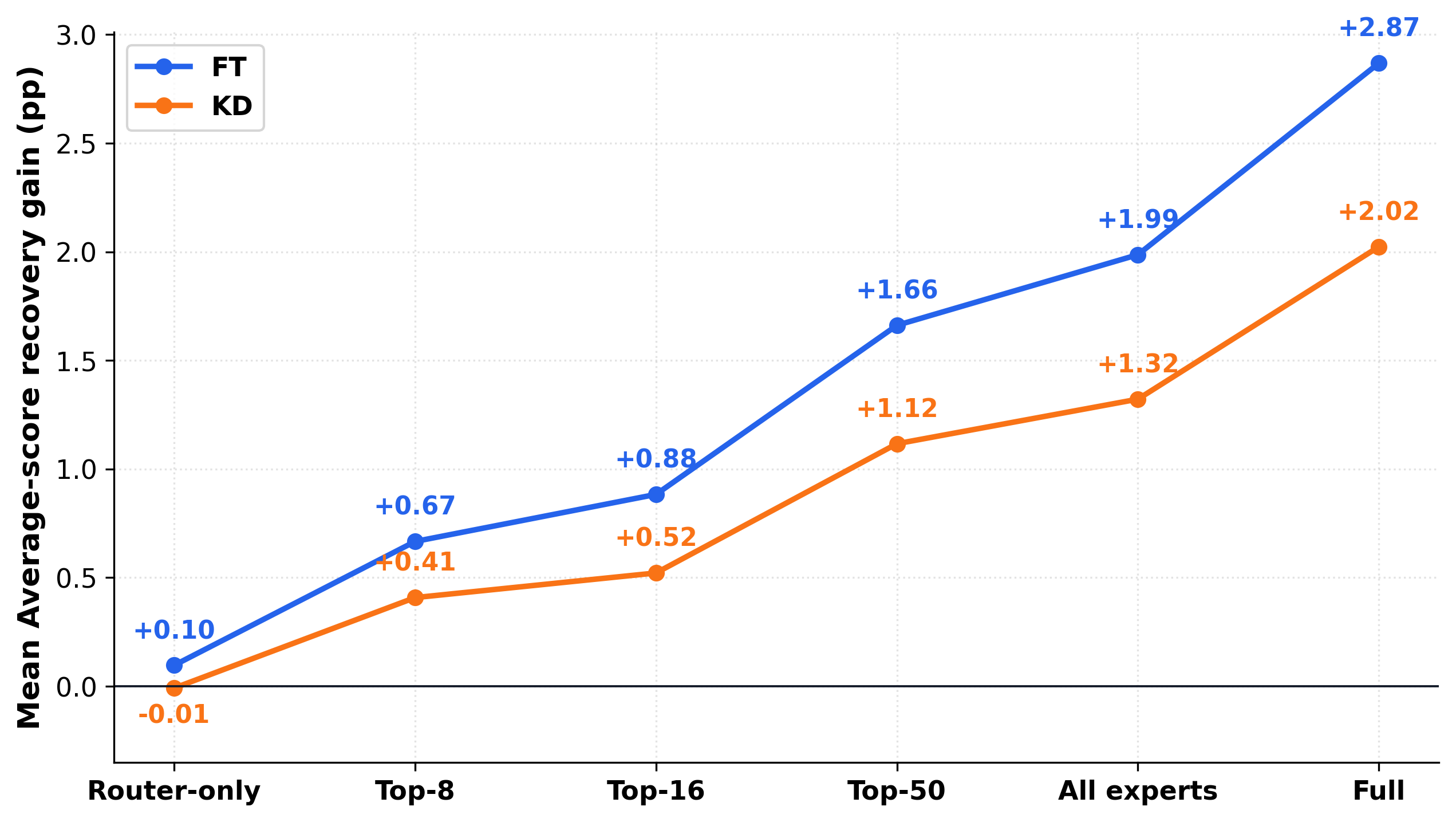}
    \caption{
Expert Pruning: Recovery gain as the trainable parameter scope expands.
For both LM fine-tuning and KD, recovery increases from router-only to router+top-$k$ experts, router+all-experts, and full-parameter adjustment.
LM fine-tuning remains stronger than KD at each matched scope.
}
    \label{fig:scope_progression_prune} 
\end{figure}

Figures~\ref{fig:overall_recovery_prune}, \ref{fig:cost_tradeoff_prune}, \ref{fig:family_recovery_prune},  and \ref{fig:scope_progression_prune} summarize the results for Expert Pruning checkpoints. 
The overall trend matches the main result: \textsc{Full FT} gives the largest recovery gain, followed by \textsc{Full KD} and expert-inclusive FT strategies. 
Router-only adjustment remains a low-cost but low-recovery option, indicating that pruning-induced degradation cannot be fully repaired by router reweighting alone. 
The benchmark-family view shows that \textsc{Full FT} is especially effective for CoT and MCQA. 
Finally, the scope-progression plot shows a monotonic improvement as the trainable scope expands, with causal LM fine-tuning outperforming KD at every matched scope.

    

\newpage

\newpage

\vspace*{\fill}
\null

\newpage

\vspace*{\fill}
\null

\newpage

\subsection{Expert Merging}

\begin{figure}[h]
    \centering 
    \includegraphics[width=\linewidth]{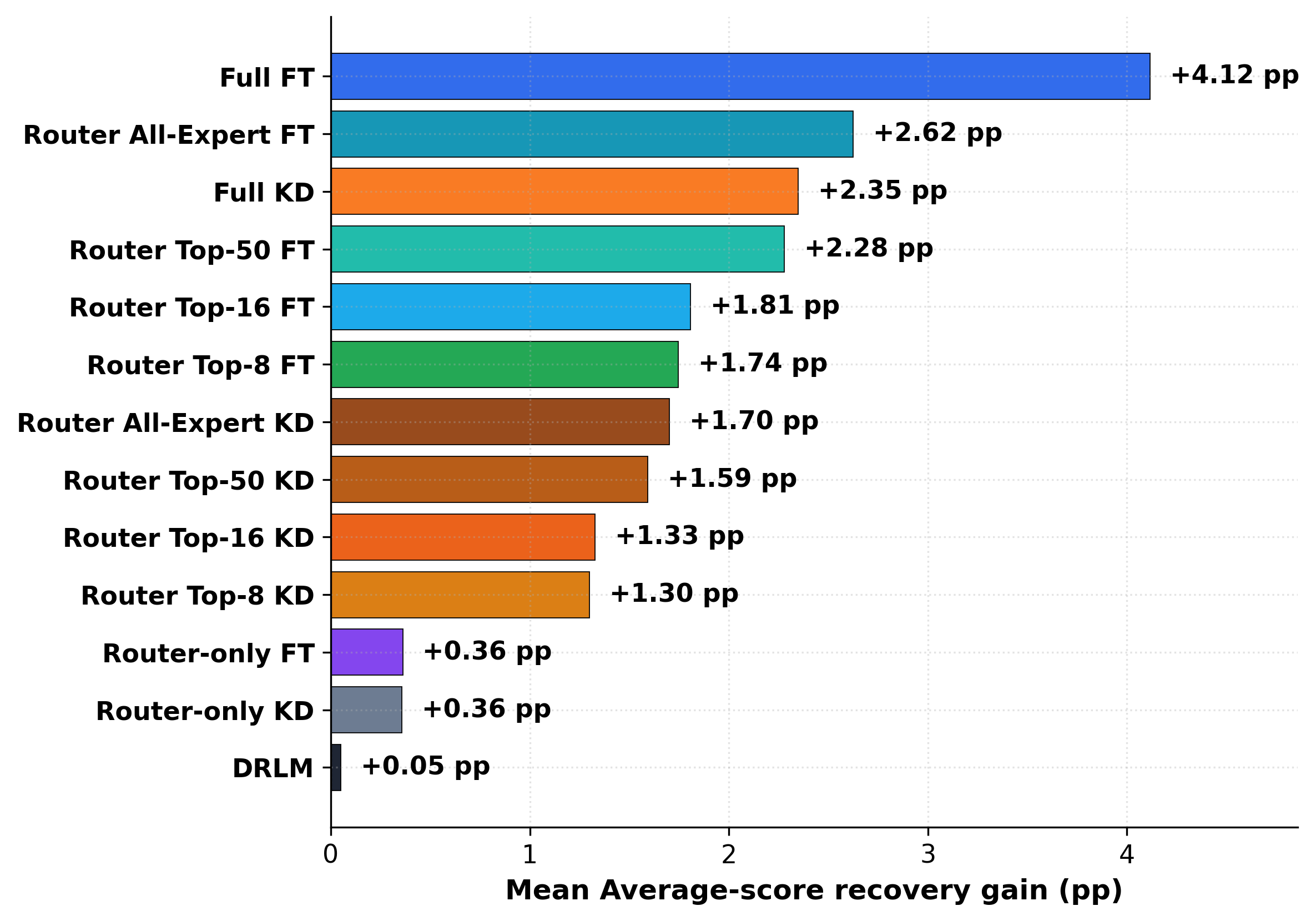}
    \caption{
Expert Merging: Overall recovery gain of post-compression adjustment strategies.
Bars show mean average-score recovery gain, measured in percentage points (pp), over the unadjusted merging baseline.
Results are averaged over merging checkpoints, retention ratios, and benchmarks.
\textsc{Full FT} achieves the largest recovery, and causal LM fine-tuning consistently outperforms KD under matched trainable scopes.
}
    \label{fig:overall_recovery_merge} 
\end{figure}

\begin{figure}[!t]
    \centering
    \includegraphics[width=0.95\linewidth]{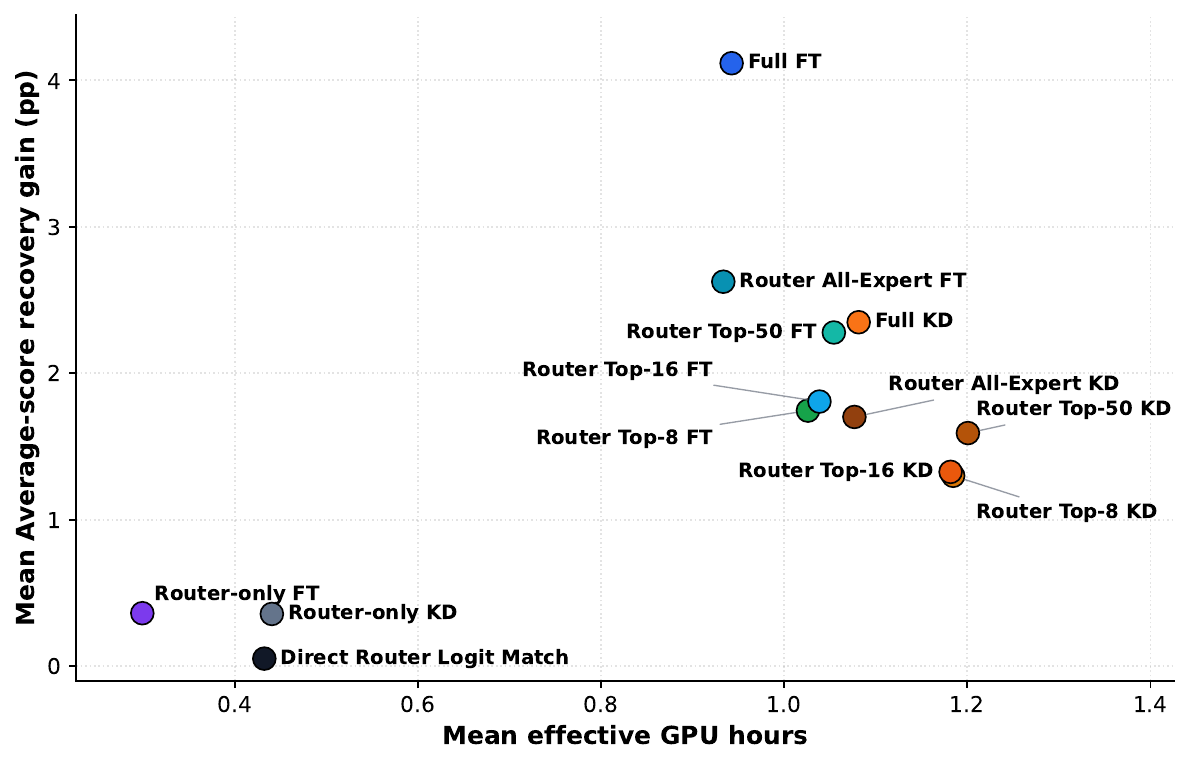}\\[1ex]
    \includegraphics[width=0.95\linewidth]{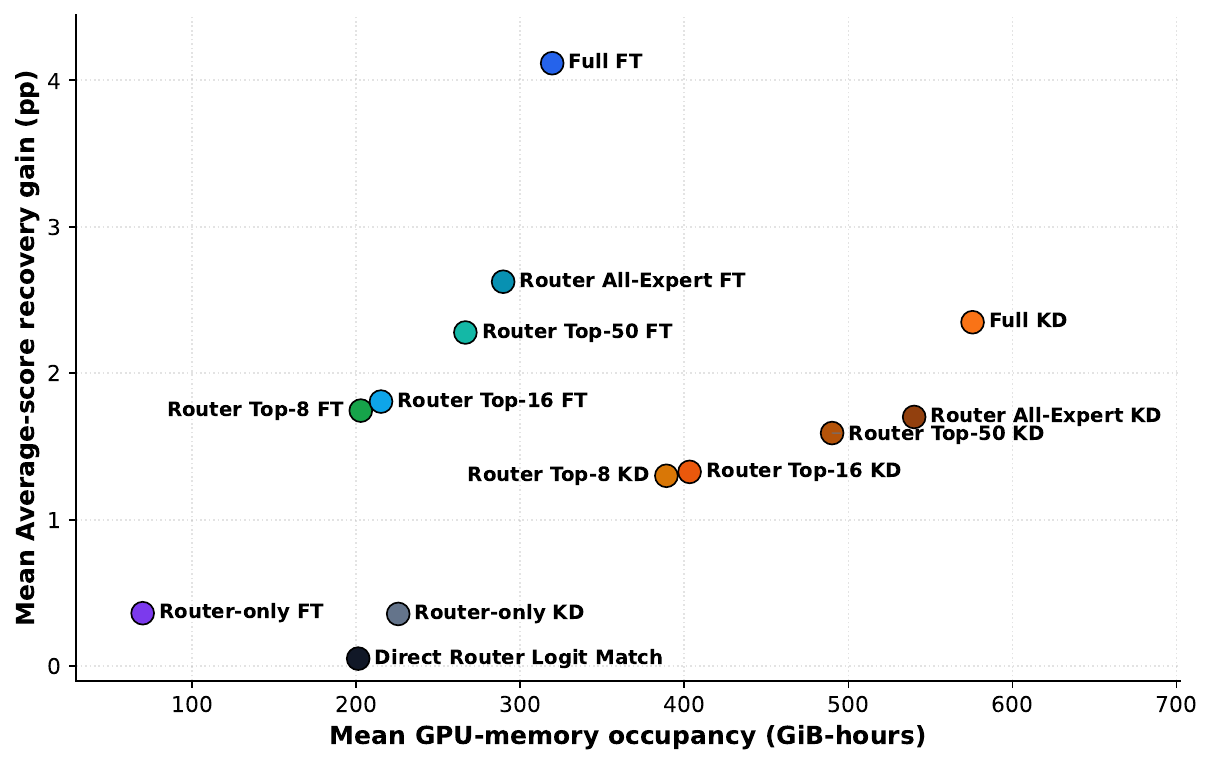}\\[1ex]
    \includegraphics[width=0.95\linewidth]{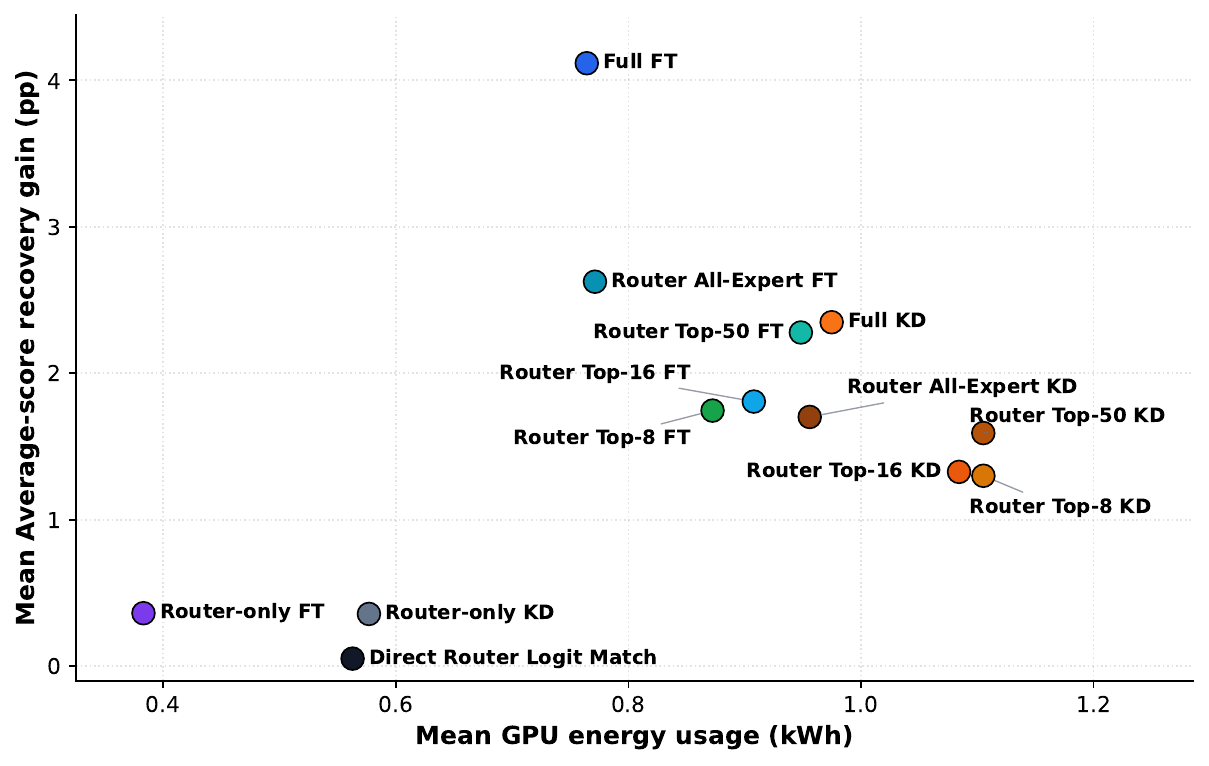}
    
    \caption{Expert Merging: GPU cost--performance trade-off of
post-compression adjustment strategies. The y-axis shows mean
average-score recovery gain, while the x-axis shows effective
GPU-hours (top), GPU-memory occupancy in GiB-hours (middle), and GPU
energy in kWh (bottom). Each point is an unweighted arithmetic mean
over six merging method--ratio settings (two merging methods $\times$
three retention ratios). Both expert-selection cost and training-loop
cost are included when applicable. Upper-left is better.}
    \label{fig:cost_tradeoff_merge}
\end{figure}

\begin{figure}[t]
    \centering 
    \includegraphics[width=\linewidth]{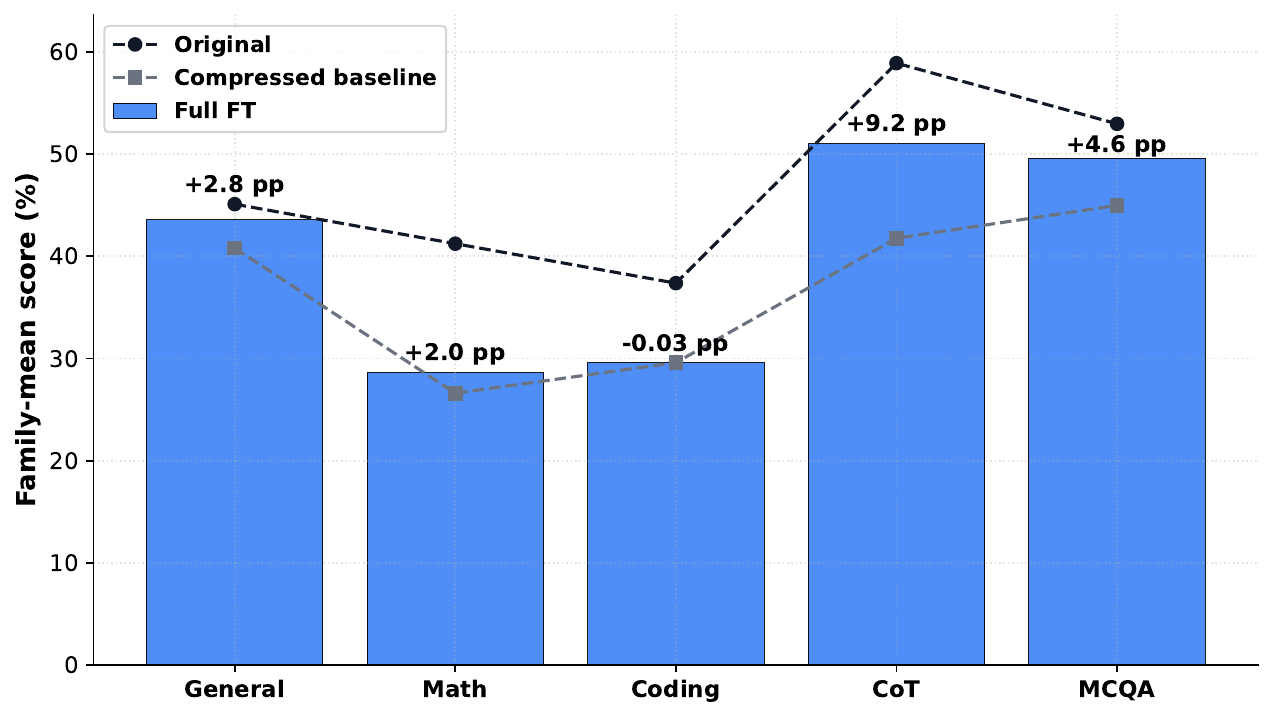}
    \caption{
Expert Merging: Benchmark-family recovery of \textsc{Full FT}.
Bars show post-adjustment performance, and dashed lines denote the original model and the unadjusted merging baseline.
\textsc{Full FT} recovers strongly on CoT, MCQA, general, and math benchmarks, while coding is
essentially unchanged. 
}
    \label{fig:family_recovery_merge} 
\end{figure}

\begin{figure}[t]
    \centering 
    \includegraphics[width=\linewidth]{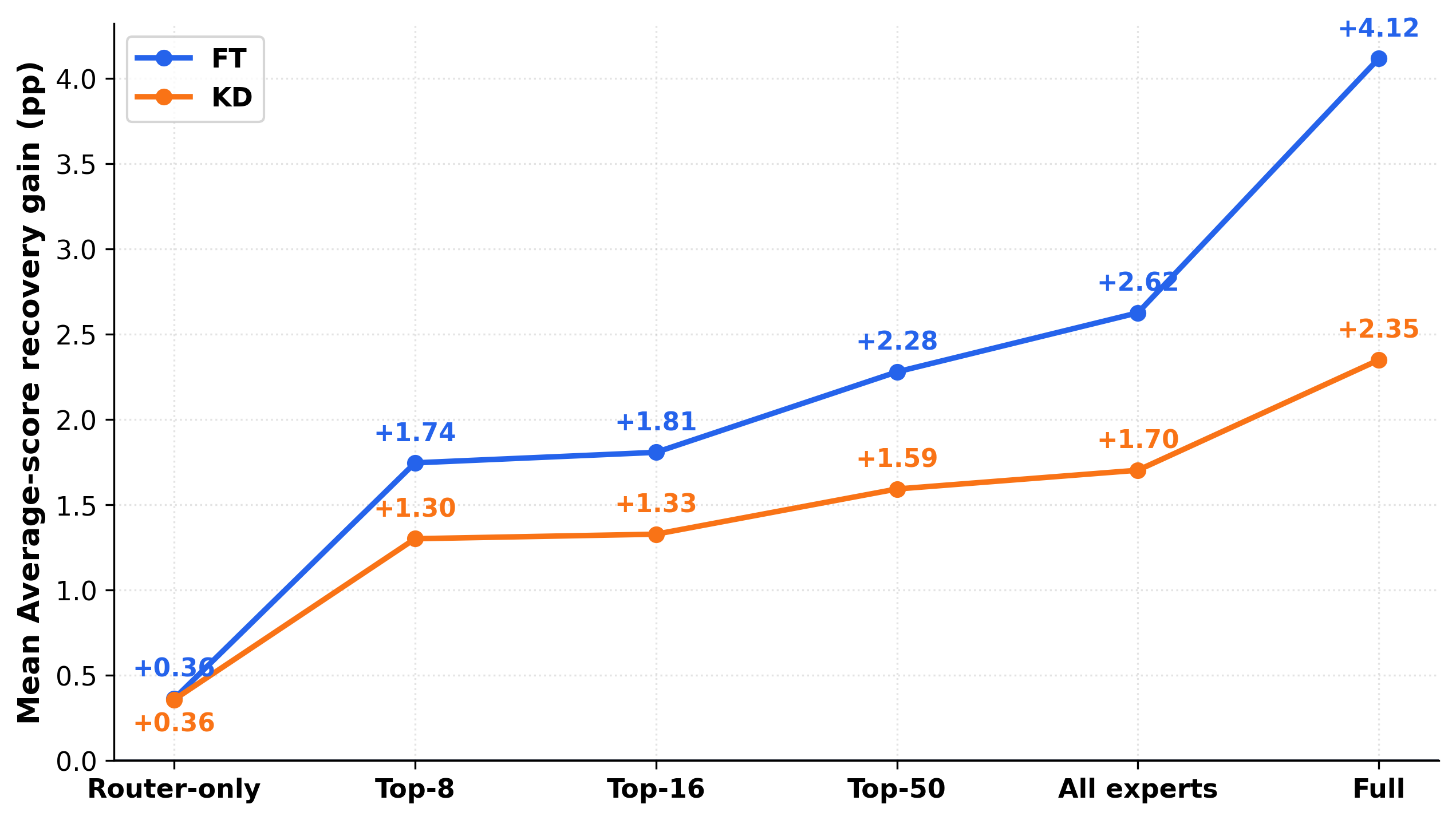}
    \caption{
Expert Merging: Recovery gain as the trainable parameter scope expands.
For both LM fine-tuning and KD, recovery increases as the trainable scope expands from router-only to router+top-$k$ experts, router+all-experts, and full-parameter adjustment.
The large final jump to full-parameter adjustment mirrors the main result and suggests that merging-induced residuals are not confined to MoE expert parameters alone.
}
    \label{fig:scope_progression_merge} 
\end{figure}

Figures~\ref{fig:overall_recovery_merge}, \ref{fig:cost_tradeoff_merge}, \ref{fig:family_recovery_merge} and \ref{fig:scope_progression_merge} report the same analysis for Expert Merging checkpoints. 
Again, \textsc{Full FT} achieves the strongest recovery, and LM fine-tuning consistently provides higher gains than KD under the same trainable scope. 
The recovery gain is larger than in the pruning-only breakdown, suggesting that merged checkpoints also leave substantial recoverable residual errors. 
Router+top-$k$ strategies improve over router-only adjustment, but they do not dominate router+all-experts or full-parameter adjustment once recovery gain and expert-selection overhead are considered together. 
The benchmark-family and scope-progression plots further confirm the same conclusion as the main text: small post-compression adjustment is broadly useful, but expanding the trainable scope and using LM loss are key to strong recovery.

    

\newpage

\vspace*{\fill}
\null

\newpage

\vspace*{\fill}
\null

\newpage
\section{Expert Editing}
\label{appendix:equations_editing}

\subsection{Compatibility-Mode Analysis for Expert Editing}
\label{appendix:expert_editing_compatibility}

\paragraph{Why Expert Editing is reported separately.}
Expert Editing methods reduce the number of expert parameters by
factorizing or tensor-decomposing expert weights while keeping the number of
expert indices unchanged.  Representative methods include
MoBE~\cite{mobe} and TD-MoE~\cite{td_moe}, which compress expert internals
through rank or tensor decomposition.  In principle, post-compression
adjustment should be performed directly on the native factorized
representation in order to preserve the compressed parameterization.
However, native Expert Editing implementations often lack the optimized
training and inference support available for standard MoE checkpoints.
Efficient end-to-end execution may require custom fused or mega-kernel
implementations for the factorized expert operators; without such kernels,
native checkpoints can be substantially slower to train and evaluate than
materialized standard-MoE checkpoints.

For this reason, we evaluate Expert Editing in a compatibility-mode setting:
we use the reconstructed standard-MoE realization supplied by the editing
method so that it can run in the same training, serving, and evaluation
stack as the pruning and merging checkpoints.  These experiments test
whether the materialized function produced by an editing compressor is
amenable to post-compression adjustment.  They should not be interpreted as
showing that expert-inclusive adjustment of the reconstructed checkpoint is
parameterization-equivalent to adjustment of the native factorization.

\paragraph{Forward equivalence before adjustment.}
Let $\psi$ denote the native factor parameters, and let $R$ denote the
reconstruction map that materializes those factors as standard expert
weights.  For expert $i$ and projection type $q$,
\begin{equation}
    \widehat W_i^q=R_i^q(\psi).
    \label{eq:editing-reconstruction-map}
\end{equation}
For example, in a basis-decomposition method, $R_i^q(\psi)$ may combine an
expert-specific factor with a linear combination of shared basis matrices.
The reconstructed checkpoint explicitly stores $\widehat W_i^q$.

Before adjustment, the native and reconstructed operators implement the
same materialized expert function up to numerical precision:
\begin{equation}
    E_i^{\mathrm{native}}(h;\psi)
    =
    E_i^{\mathrm{recon}}(h;R(\psi)).
    \label{eq:editing-forward-equivalence-expert}
\end{equation}
Hence, when the router and all non-expert parameters are identical,
\begin{equation}
    f_{\mathrm{native}}(x;\psi,\theta_R)
    =
    f_{\mathrm{recon}}(x;R(\psi),\theta_R)
    \label{eq:editing-forward-equivalence-model}
\end{equation}
up to numerical precision.  The reconstructed checkpoint is therefore a
valid functional realization of the native edited model at initialization.

\paragraph{Router-only adjustment is equivalent under frozen experts.}
If only router parameters are updated, both $\psi$ and $R(\psi)$ remain
fixed.  Provided the native and reconstructed expert operators implement
the same differentiable function, their loss values and router gradients
are equal up to numerical precision:
\begin{equation}
    \nabla_{\theta_R}
    \mathcal{L}_{\mathrm{native}}(\theta_R;\psi)
    =
    \nabla_{\theta_R}
    \mathcal{L}_{\mathrm{recon}}(\theta_R;R(\psi)).
    \label{eq:editing-router-gradient-equivalence}
\end{equation}
Thus, router-only adjustment of the reconstructed standard-MoE realization
is functionally equivalent to router-only adjustment of the native
factorized model, subject to ordinary numerical differences between the two
implementations.

\paragraph{Expert-inclusive and full adjustment are not generally equivalent.}
The equivalence does not extend to updates of expert parameters.  The native
model moves in factor space, whereas the reconstructed model moves directly
in materialized weight space.  To make this distinction explicit, consider
one Euclidean gradient step; adaptive optimizers or weight decay introduce
additional differences.

For the reconstructed checkpoint,
\begin{equation}
    \widehat W_{t+1}
    =
    \widehat W_t
    -
    \eta\nabla_{\widehat W}\mathcal{L}(\widehat W_t).
    \label{eq:editing-direct-materialized-update}
\end{equation}
For the native factorization,
\begin{equation}
    \psi_{t+1}
    =
    \psi_t
    -
    \eta\nabla_{\psi}\mathcal{L}(R(\psi_t)).
    \label{eq:editing-native-factor-update}
\end{equation}
Writing $J_R(\psi_t)$ for the Jacobian of the reconstruction map and
$\widehat W_t=R(\psi_t)$, the chain rule gives
\begin{equation}
    \nabla_{\psi}\mathcal{L}(R(\psi_t))
    =
    J_R(\psi_t)^{\top}
    \nabla_{\widehat W}\mathcal{L}(\widehat W_t).
    \label{eq:editing-factor-chain-rule}
\end{equation}
A first-order Taylor expansion of the induced materialized update yields
\begin{equation}
\begin{aligned}
R(\psi_{t+1})
&=
\widehat W_t
-
\eta J_R(\psi_t)J_R(\psi_t)^{\top}
\\[-0.2ex]
&\qquad\cdot
\nabla_{\widehat W}\mathcal{L}(\widehat W_t)
+O(\eta^2).
\end{aligned}
\label{eq:editing-induced-materialized-update}
\end{equation}
The first-order native update equals the direct materialized update only if
\begin{equation}
\begin{aligned}
J_R(\psi_t)J_R(\psi_t)^{\top}
\nabla_{\widehat W}\mathcal{L}(\widehat W_t)
&=
\nabla_{\widehat W}\mathcal{L}(\widehat W_t).
\end{aligned}
\label{eq:editing-update-equivalence-condition}
\end{equation}
This is a restrictive fixed-point condition.  Membership of the loss
gradient in the tangent space
$\operatorname{range}(J_R(\psi_t))$ is necessary, but is not sufficient in
general, because $J_RJ_R^{\top}$ need not be the orthogonal projector or the
identity on that tangent space.  Equality additionally requires
$J_RJ_R^{\top}$ to act as the identity on the particular gradient
direction.  Shared bases or tensor factors can also couple the updates of
multiple experts, whereas direct reconstructed-weight adjustment can move
materialized expert matrices more freely.

Accordingly, expert-inclusive and full-parameter adjustment of a
reconstructed Expert Editing checkpoint should be interpreted as an
unconstrained compatibility-mode update in materialized weight space, not as
an update equivalent to optimizing the native compressed parameterization.

\paragraph{Empirical trend and interpretation.}
Although compatibility-mode Expert Editing is excluded from the primary
compressed-artifact cost comparison, we report it for completeness.  The
observed trend is broadly consistent with the pruning and merging results:
small post-compression adjustment improves the reconstructed checkpoints,
causal LM fine-tuning tends to outperform pure teacher KD under the same
small budget, router-only adjustment gives limited but nonzero gains, and
recovery increases as the trainable scope expands.  Because
expert-inclusive and full updates do not preserve the native factorized
constraint, these results are supplementary evidence about the materialized
edited function rather than primary evidence about deployable native
Expert Editing artifacts.

\subsection{Active-Set Formulation for Expert Editing}
\label{appendix:expert_editing_formulation}

We adopt the Expert Editing notation and active-set analysis of
\citet{hyeon2026retraining}, but reorganize the residual into
routing, expert-function, and active-path components that match the
adjustment-scope analysis in Section~\ref{sec:degradation}. To connect the
formulation directly to
Appendix~\ref{appendix:expert_editing_compatibility}, we denote the
materialized edited expert by $\widehat E_i$.

Throughout Appendices~F.2--F.6, $x$ denotes a common local hidden-state input
supplied to both the original and edited layer maps. The end-to-end state
shift between the two networks is handled by the propagation analysis in
Section~\ref{sec:degradation}. Expert Editing preserves the routed-expert
index set $\mathcal{E}=\{0,\ldots,N-1\}$ but replaces $E_i$ by
$\widehat E_i$.

Let $\mathcal{S}$ and
$\mathcal{S}_{e}:=\mathcal{S}^{\mathrm{edit}}$ be the top-$k$ active sets in the original and edited layer maps at the reference input $x$, respectively.
Let $g_i^{\mathrm{edit}}(x)\ge 0$ denote the edited model's gate mass, not a raw router logit. For any nonempty active set $A\subseteq\mathcal E$ satisfying $\sum_{j\in A}g_j^{\mathrm{edit}}(x)>0$, define
\begin{equation}
\widetilde g_i^{e,\mathcal{A}}(x)
:=
\frac{g_i^{\mathrm{edit}}(x)}
     {\sum_{j\in\mathcal{A}}g_j^{\mathrm{edit}}(x)},
\qquad i\in\mathcal{A}.
\label{eq:appendix-edit-renormalized-gate}
\end{equation}
The two routed outputs are
\begin{align}
y_{\mathrm{orig}}(x)
&=
\sum_{i\in\mathcal{S}}
\widetilde g_i^{o,\mathcal{S}}(x)E_i(x),
\label{eq:appendix-edit-original-output}
\\
y_{\mathrm{edit}}(x)
&=
\sum_{i\in\mathcal{S}_{e}}
\widetilde g_i^{e,\mathcal{S}_{e}}(x)
\widehat E_i(x).
\label{eq:appendix-edit-edited-output}
\end{align}
For compact displays, write
\begin{equation}
\begin{aligned}
\widetilde g_i^{o}
&:=
\widetilde g_i^{o,\mathcal{S}}(x),
&
\widetilde g_i^{e}
&:=
\widetilde g_i^{e,\mathcal{S}_{e}}(x),\\
E_i&:=E_i(x),
&
\widehat E_i&:=\widehat E_i(x).
\end{aligned}
\label{eq:appendix-edit-display-shorthand}
\end{equation}
Define
\begin{equation}
\begin{aligned}
\mathcal{T} &= \mathcal{S}\cap\mathcal{S}_{e},
&\mathcal{D} &= \mathcal{S}\setminus\mathcal{S}_{e},\\
\mathcal{R} &= \mathcal{S}_{e}\setminus\mathcal{S}.
\end{aligned}
\label{eq:appendix-edit-set-decomposition}
\end{equation}
The exact residual is
\begin{equation}
\begin{aligned}
y_{\mathrm{orig}}-y_{\mathrm{edit}}
&=
\sum_{i\in\mathcal{T}}
\left(
\widetilde g_i^{o}E_i
-
\widetilde g_i^{e}\widehat E_i
\right)
\\
&\quad+
\sum_{i\in\mathcal{D}}
\widetilde g_i^{o}E_i
-
\sum_{i\in\mathcal{R}}
\widetilde g_i^{e}\widehat E_i.
\end{aligned}
\label{eq:appendix-edit-exact-residual}
\end{equation}
For every shared index $i\in\mathcal{T}$,
\begin{equation}
\begin{aligned}
\widetilde g_i^{o}E_i
-
\widetilde g_i^{e}\widehat E_i
&=
(\widetilde g_i^{o}-\widetilde g_i^{e})
\widehat E_i
\\
&\quad+
\widetilde g_i^{o}
(E_i-\widehat E_i).
\end{aligned}
\label{eq:appendix-edit-shared-index-identity}
\end{equation}
We therefore define
\begin{align}
e_{\mathrm{route}}^{e}(x)
&:=
\sum_{i\in\mathcal{T}}
(\widetilde g_i^{o}-\widetilde g_i^{e})
\widehat E_i,
\label{eq:appendix-edit-routing-error}
\\
e_{\mathrm{func}}^{e}(x)
&:=
\sum_{i\in\mathcal{T}}
\widetilde g_i^{o}(E_i-\widehat E_i),
\label{eq:appendix-edit-function-error}
\\
e_{\mathrm{path}}^{e}(x)
&:=
\sum_{i\in\mathcal{D}}
\widetilde g_i^{o}E_i
-
\sum_{i\in\mathcal{R}}
\widetilde g_i^{e}\widehat E_i.
\label{eq:appendix-edit-path-error}
\end{align}
Thus,
\begin{equation}
y_{\mathrm{orig}}-y_{\mathrm{edit}}
=
e_{\mathrm{route}}^{e}
+
e_{\mathrm{func}}^{e}
+
e_{\mathrm{path}}^{e}.
\label{eq:appendix-edit-three-way-residual}
\end{equation}
If the editing compressor leaves the router and its gate implementation
unchanged, then at a common input $x$,
$\widetilde g_i^{e}=\widetilde g_i^{o}$ and
$\mathcal{S}_{e}=\mathcal{S}$ before adjustment. In that special case,
$e_{\mathrm{route}}^{e}=e_{\mathrm{path}}^{e}=0$ locally, and the initial
same-input residual is purely an expert-function approximation term. Realized
routing shifts in deeper layers can still arise through the upstream
hidden-state error described in Section~\ref{sec:degradation}. We retain the
general three-way identity because it also covers altered gate
implementations and post-compression router adjustment.

\subsection{Best Scenario: Preserved Active Path}
\label{appendix:expert_editing_best}

If $\mathcal{S}_{e}=\mathcal{S}$, then
$\mathcal{T}=\mathcal{S}$ and
$\mathcal{D}=\mathcal{R}=\emptyset$. Therefore,
\begin{equation}
y_{\mathrm{orig}}-y_{\mathrm{edit}}
=
e_{\mathrm{route}}^{e}
+
e_{\mathrm{func}}^{e},
\label{eq:appendix-edit-best-residual}
\end{equation}
with
\begin{equation}
\begin{aligned}
y_{\mathrm{orig}}-y_{\mathrm{edit}}
&=
\sum_{i\in\mathcal{S}}
(\widetilde g_i^{o}-\widetilde g_i^{e})
\widehat E_i
\\
&\quad+
\sum_{i\in\mathcal{S}}
\widetilde g_i^{o}(E_i-\widehat E_i).
\end{aligned}
\label{eq:appendix-edit-best-expanded}
\end{equation}
Preserving the selected indices does not imply a zero residual. In the
general identity, shared-index router reweighting and expert-function
approximation may both remain. If the router is unchanged and both layer maps
receive the same input, the first term vanishes and only the function
approximation term remains. No ordering between the two terms is implied
without additional assumptions.

\subsection{Most Common Scenario: Partial Active-Set Overlap}
\label{appendix:expert_editing_partial}

If $0<|\mathcal{T}|<k$, then
$\mathcal{D}\neq\emptyset$ and $\mathcal{R}\neq\emptyset$. All three terms in
Eq.~\eqref{eq:appendix-edit-three-way-residual} may be nonzero:
$e_{\mathrm{route}}^{e}$ measures gate-weight disagreement on shared indices,
$e_{\mathrm{func}}^{e}$ measures function mismatch for those shared indices,
and $e_{\mathrm{path}}^{e}$ compares original contributions that leave the
active path with newly activated edited-expert contributions. This separates
within-index editing error from cross-index path replacement.

\subsection{Worst Scenario: Disjoint Active Paths}
\label{appendix:expert_editing_disjoint}

If $\mathcal{S}\cap\mathcal{S}_{e}=\emptyset$, then
$\mathcal{T}=\emptyset$, $\mathcal{D}=\mathcal{S}$, and
$\mathcal{R}=\mathcal{S}_{e}$. Under the shared-index decomposition,
\begin{equation}
e_{\mathrm{route}}^{e}(x)
=
e_{\mathrm{func}}^{e}(x)
=0,
\label{eq:appendix-edit-disjoint-zero}
\end{equation}
and
\begin{equation}
\begin{aligned}
y_{\mathrm{orig}}-y_{\mathrm{edit}}
&=
e_{\mathrm{path}}^{e}(x)
\\
&=
\sum_{i\in\mathcal{S}}
\widetilde g_i^{o}E_i
-
\sum_{i\in\mathcal{S}_{e}}
\widetilde g_i^{e}\widehat E_i.
\end{aligned}
\label{eq:appendix-edit-disjoint-residual}
\end{equation}
This is complete active-path replacement. The term ``worst'' names the
disjoint-set scenario; it does not establish that the residual norm is
maximal for every input.

\subsection{Connection to Adjustment Scope}
\label{appendix:expert_editing_scope}

To match the two-source interpretation in Section~\ref{sec:degradation},
define
\begin{equation}
e_{\mathrm{expert/path}}^{e}
:=
e_{\mathrm{func}}^{e}+e_{\mathrm{path}}^{e}.
\label{eq:appendix-edit-expert-path-error}
\end{equation}
Then
\begin{equation}
y_{\mathrm{orig}}-y_{\mathrm{edit}}
=
e_{\mathrm{route}}^{e}
+
e_{\mathrm{expert/path}}^{e}.
\label{eq:appendix-edit-two-source-residual}
\end{equation}
Because Section~\ref{sec:degradation} uses
$\delta_\ell=\widehat h_\ell-h_\ell$, the same-input edited-model source is
\begin{equation}
\epsilon_\ell^{e}
:=
y_{\mathrm{edit},\ell}-y_{\mathrm{orig},\ell}
=
-\left(
e_{\mathrm{route},\ell}^{e}
+
e_{\mathrm{expert/path},\ell}^{e}
\right).
\label{eq:appendix-edit-propagation-sign}
\end{equation}

Under router-only adjustment, the materialized expert functions
$\widehat E_i$ remain fixed. The router can change
$e_{\mathrm{route}}^{e}$ and, by moving top-$k$ boundaries, may also change
$\mathcal{T}$, $\mathcal{D}$, and $\mathcal{R}$. It cannot, however, change
the expert functions themselves or directly eliminate
$E_i-\widehat E_i$. As established in
Appendix~\ref{appendix:expert_editing_compatibility}, router-only adjustment
is functionally equivalent between the native factorized model and its
reconstructed standard-MoE realization, up to numerical precision.

Expert-inclusive or full adjustment of the reconstructed checkpoint can also
change $\widehat E_i$, but these updates occur in unconstrained materialized
weight space and are not generally equivalent to native factor updates.
Full-parameter adjustment additionally exposes dense/shared correction
directions for downstream effects of $\epsilon_\ell^{e}$, consistent with
Section~\ref{sec:degradation}.

\subsection{Expert Editing Experiment Results}

\begin{figure}[h]
    \centering 
    \includegraphics[width=\linewidth]{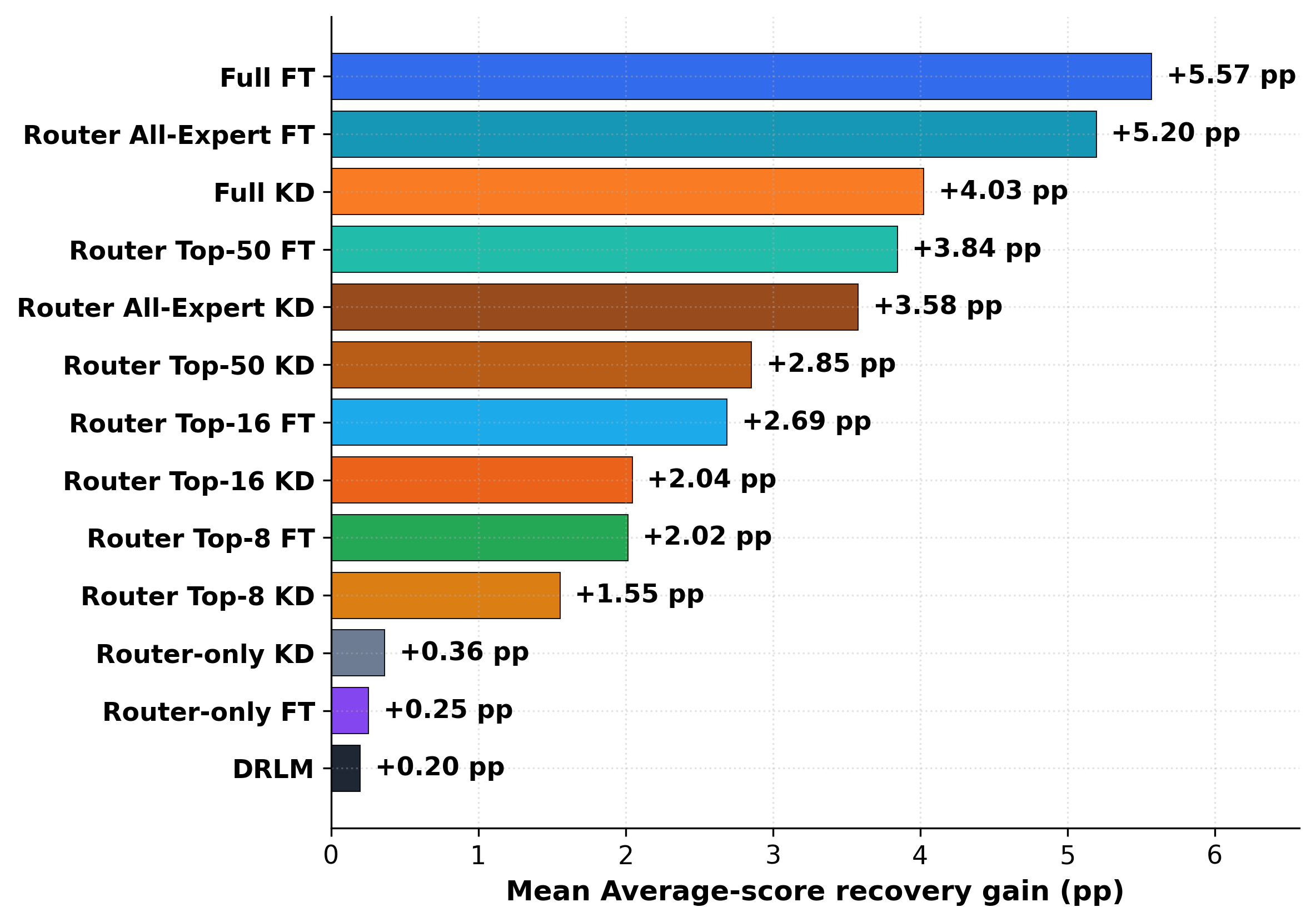}
    \caption{
Compatibility-mode Expert Editing: Overall recovery gain of post-compression adjustment strategies.
Bars show the mean average-score recovery gain, measured in percentage points (pp), over the unadjusted reconstructed Expert Editing baseline. \textsc{Full FT} achieves the largest recovery. LM fine-tuning outperforms KD at every expert-inclusive and full-parameter matched scope; the router-only pair is the sole exception, where \textsc{Router-only KD} yields $+0.36$ pp versus $+0.25$ pp for \textsc{Router-only FT}. Router-only adjustment provides only limited gains. These results are reported as supplementary compatibility-mode evidence because adjustment is performed on reconstructed standard-MoE weights rather than native factorized Expert Editing parameters.
}
    \label{fig:overall_recovery_edit} 
\end{figure}

\begin{figure}[t]
    \centering 
    \includegraphics[width=\linewidth]{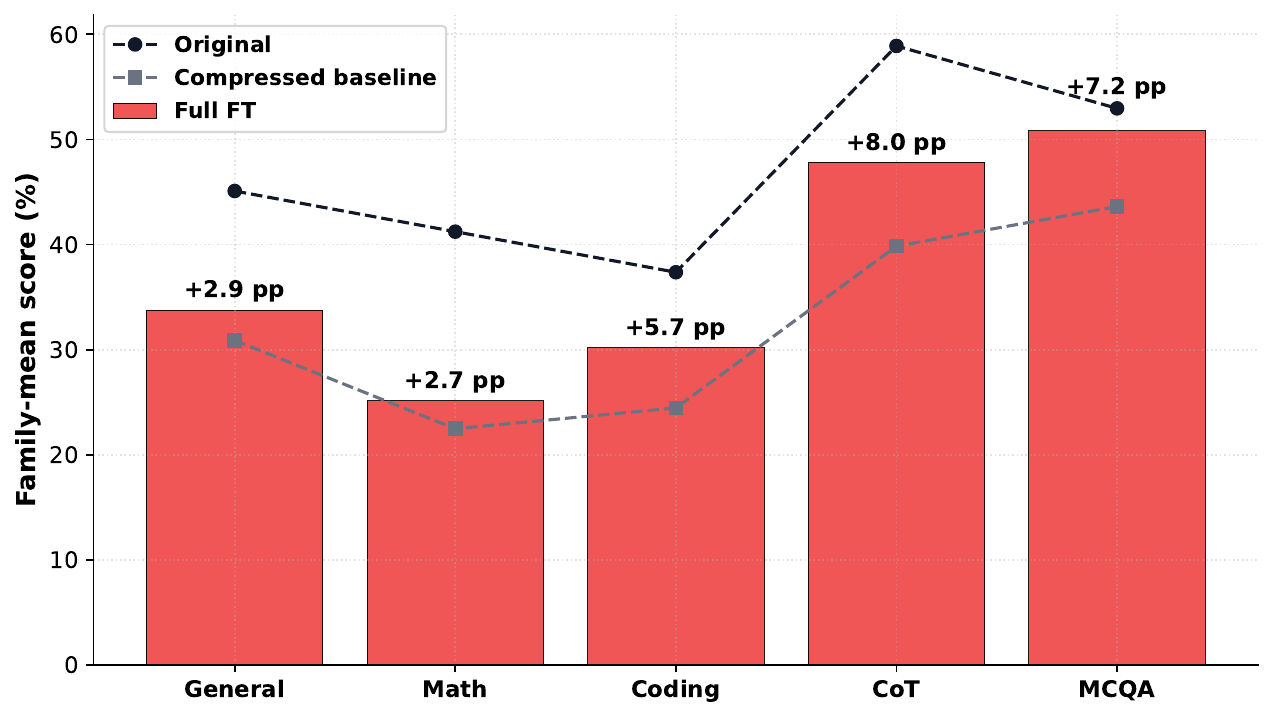}
    \caption{
Compatibility-mode Expert Editing: Benchmark-family recovery of \textsc{Full FT}.
Bars show post-adjustment performance, and dashed lines denote the original model and the unadjusted reconstructed Expert Editing baseline.
\textsc{Full FT} yields large gains on CoT and MCQA benchmarks, while math shows the weakest recovery in this setting.
}
    \label{fig:family_recovery_edit} 
\end{figure}

\begin{figure}[h!]
    \centering 
    \includegraphics[width=\linewidth]{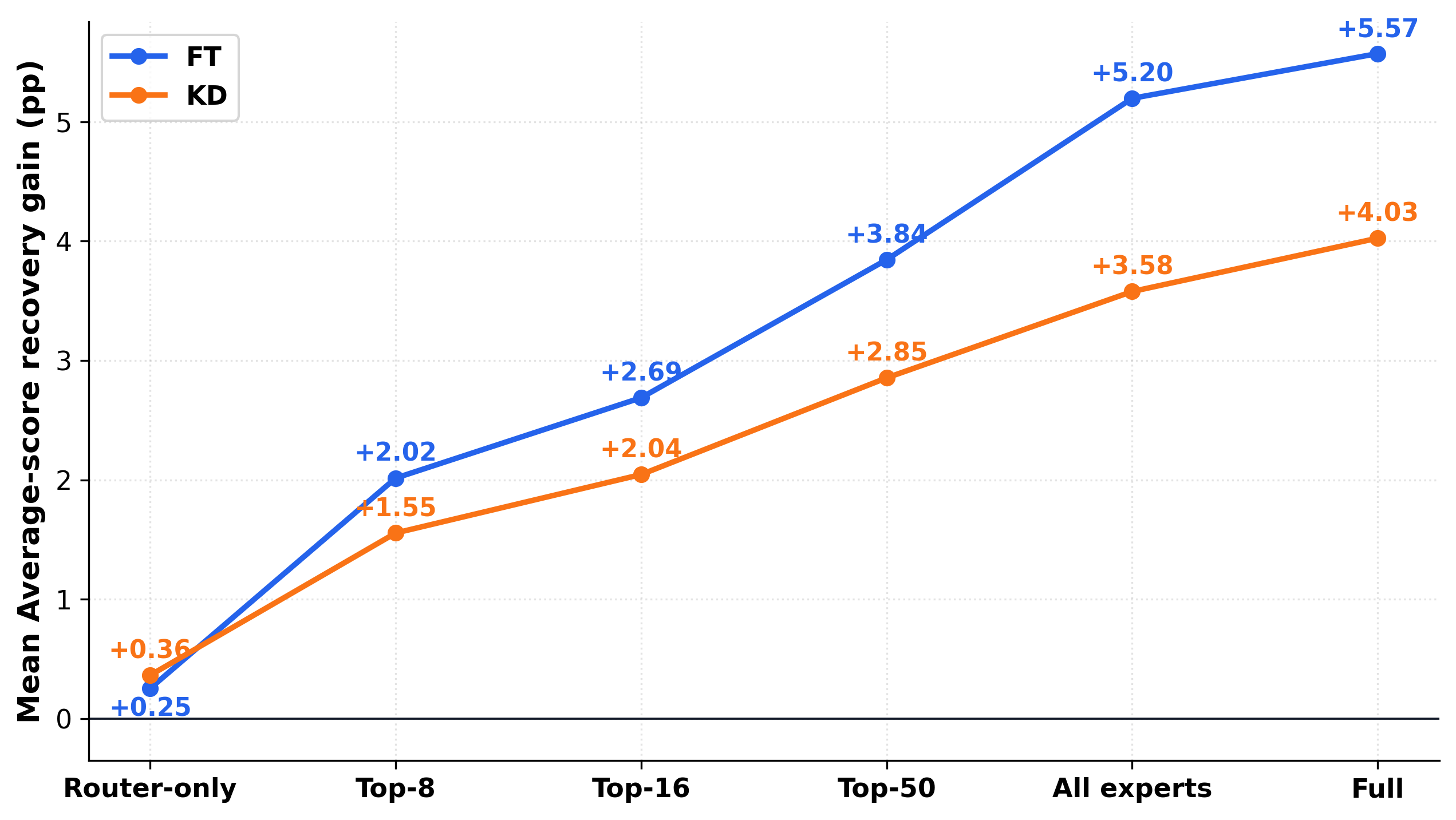}
    \caption{
Compatibility-mode Expert Editing: Recovery gain as the trainable parameter scope expands.
For both LM fine-tuning and KD, recovery increases from router-only adjustment to router+top-$k$ experts, router+all-experts, and full-parameter adjustment.
The same monotonic scope-expansion trend observed in the main pruning/merging experiments also appears for reconstructed Expert Editing checkpoints.
}
    \label{fig:scope_progression_edit} 
\end{figure}

Figures~\ref{fig:overall_recovery_edit}, \ref{fig:family_recovery_edit} and  \ref{fig:scope_progression_edit} summarize the compatibility-mode Expert Editing results.
Although these results are excluded from the main deployable compressed-artifact comparison, they show trends broadly consistent with the main Expert Pruning and Expert Merging experiments.
First, small post-compression adjustment improves the reconstructed Expert Editing checkpoints, with \textsc{Full FT} achieving the largest average recovery gain.
Second, causal LM fine-tuning is stronger than token-level KD at every expert-inclusive and full-parameter matched scope; the router-only pair is the sole exception, where \textsc{Router-only KD} yields $+0.36$ pp versus $+0.25$ pp for \textsc{Router-only FT}.
Third, router-only adjustment gives limited but nonzero gains, while expanding the trainable scope from router-only to router+experts and full-parameter adjustment monotonically improves recovery.
These observations suggest that Expert Editing checkpoints are also adjustment-friendly.
However, because these experiments update reconstructed standard-MoE weights rather than native factorized Expert Editing parameters, they should be interpreted as supplementary compatibility-mode evidence rather than primary evidence for compressed-artifact efficiency.

\section{Qualitative Examples of Generation-Level Recovery}
\label{app:qual_rep_example}

In Section~\ref{sec:results}, we observe that \textsc{Full FT} recovers most strongly on general, CoT, and MCQA-style benchmarks, where fluent continuation and stable natural-language generation are important. 
Here we provide representative qualitative examples illustrating the type of generation-level degradation that appears after compression and how causal LM fine-tuning repairs it.

Figure~\ref{fig:resolved-full-ft-examples} shows two cases: an Expert Pruning checkpoint on Gemma4 with AIMER at 50\% retention, and an Expert Merging checkpoint on Qwen3 with HC-SMoE at 62.5\% retention. 
Before adjustment, both compressed models exhibit degenerate continuation patterns, such as repeated tokens or broken word fragments. 
After \textsc{Full FT}, the same prompts produce substantially more fluent and coherent explanations without any task-specific supervision. 
These examples are not intended as a comprehensive evaluation of generation quality, but they support the interpretation that small causal LM adjustment can restore the model's ability to express and compose its remaining knowledge in natural language.

\begin{figure*}[t]
\small
\setlength{\fboxsep}{5pt}
\newcommand{\examplegap}{\vspace{0.75em}}

\noindent\textbf{Example 1: Expert pruning, Gemma4 AIMER 50\%.} \\
\textbf{Prompt:} what are forces (gravity, magnetism) made of? \\

\vspace{0.35em}
\noindent\begin{minipage}[t]{0.48\textwidth}
\textbf{Compressed model before \textsc{Full FT}}
\begin{quote}\footnotesize
Imagine you are playing in a playground. You see a ball rolling down down down down down down down down down down down down down down down down down down down down down down down down down down down down down down down down down down down down down down down down ......
\end{quote}
\end{minipage}\hfill
\begin{minipage}[t]{0.48\textwidth}
\textbf{After \textsc{Full FT}}
\begin{quote}\footnotesize
Imagine you are playing in a playground with a big, heavy ball. If you throw that ball down, it does not just go away and disappear. It stays in the ground, and if you throw it up, it stays in the sky. It does not just float on its own; it stays there because it is ``stuck'' to the ground or the sky. That ``stickiness'' is what we call gravity. ...... When we talk about forces like gravity or magnetism, we are talking about the rules of the universe. These rules are made of the way things are connected.
\end{quote}
\end{minipage}

\examplegap
\noindent\textbf{Example 2: Expert merging, Qwen3 HC-SMoE 62.5\%.} \\
\textbf{Prompt:} How do overlays in NFL look so real? Like paint? How are they so accurate? \\

\vspace{0.35em}
\noindent\begin{minipage}[t]{0.48\textwidth}
\textbf{Compressed model before \textsc{Full FT}}
\begin{quote}\footnotesize
Okay, let us talk about this like you are a kid who is just learning how to talk. So imagine you are watching a video of a football game, and you see this thing called an ``overlay.'' It looks like a flat layer of color that just sits on the ground, like paint. But here is the cool part: even though it looks like paint, it is not really paint at all. It is not like paint that you can touch or that you can walk on and it will fl fl fl fl fl fl fl fl fl fl fl fl fl fl fl ......
\end{quote}
\end{minipage}\hfill
\begin{minipage}[t]{0.48\textwidth}
\textbf{After \textsc{Full FT}}
\begin{quote}\footnotesize
Okay, imagine you are playing football and you are on the field. You are running, jumping, dodging, and suddenly you see these bright lines and colors that look like paint on the grass. They are not just lines; they are little signs that show where the ball is going and where the players are. ...... Instead, they use a digital overlay technology. The computer knows exactly where every player is, where the ball is, and where the field lines are, so it draws the graphics in real time and makes them look as if they belong to the field.
\end{quote}
\end{minipage}

\caption{
Qualitative examples of generation-level recovery after \textsc{Full FT}.
We compare outputs before and after causal LM full-parameter fine-tuning for two compressed checkpoints: Gemma4 AIMER at 50\% retention and Qwen3 HC-SMoE at 62.5\% retention.
Before adjustment, the compressed models exhibit degenerate generation patterns such as token repetition and broken fragments.
After \textsc{Full FT}, the same prompts yield more fluent, coherent, and complete natural-language explanations without task-specific supervision.
}
\label{fig:resolved-full-ft-examples}
\end{figure*}

\newpage

\newpage

\vspace*{\fill}
\null

\newpage

\vspace*{\fill}
\null

\newpage

\section{Related Works}
\label{appendix:related_works}

Efficient LLM execution is increasingly important not only for serving and deploying individual models~\cite{vLLM,sheng2023flexgen,dyninomni,yu2022orca}, but also for multi-agent systems and frameworks, where a single task may require multiple model invocations~\cite{li2023camel,wu2023autogen,hyeon2026mata,hong2024metagpt}. Against this broader efficiency backdrop, we focus on reducing the deployment cost of Mixture-of-Experts (MoE) LLMs through parameter-level compression and small post-compression adjustment.

We consider methods that reduce the number of expert parameters rather than methods that change the bit representation of individual parameters. Following the taxonomy introduced in the previous study~\cite{hyeon2026retraining}, we use \textbf{Expert Pruning}, \textbf{Expert Editing}, and \textbf{Expert Merging} as descriptive categories for parameter-level MoE compression. These categories differ in whether they remove experts, re-parameterize expert internals, or aggregate multiple experts into a smaller expert set. 

\subsection{Compression Operators and Their Consequences}

\paragraph{Removing experts.}
Expert Pruning reduces the feasible expert set by deleting experts that are estimated to be redundant or weakly contributive. Existing approaches differ mainly in how they estimate expert importance. NAEE~\cite{naee} searches for the retained expert combination that minimizes layer-wise reconstruction loss on calibration data, whereas DiEP~\cite{bai2025diep} relaxes discrete expert selection into a differentiable optimization problem. REAP~\cite{reap} argues that one-shot merging can collapse functional subspaces and instead scores experts using both router gate values and expert-output magnitudes. CFES~\cite{anonymous2025compressing} derives a layer-wise output-discrepancy bound and uses it to replace an expensive global search with a coarse-to-fine layer-wise procedure. Other criteria exploit domain-conditioned expert-output statistics and token variation~\cite{easy_ep}, activation trajectories accumulated across layers~\cite{moepathfinder}, or unified evaluations of alternative expert-dropping signals~\cite{mc_suite}. Beyond removing individual experts, coarser approaches also consider deleting complete MoE layers or transformer blocks~\cite{RS_TMLR}.

\paragraph{Aggregating experts.}
Expert Merging maps multiple original experts to a smaller set of synthesized experts. Its motivation is related to model-merging results showing that related parameter sets can often be combined while retaining useful behavior~\cite{modelsoup,mergingmodelsfisher}. HC-SMoE~\cite{hcsmoe} clusters experts using output similarity measured on calibration data. PuzzleMoE~\cite{puzzlemoe} uses an entry-wise similarity mask together with an activation-weighted saliency mask to merge selected parameters while retaining expert-specific information. MergeMoE~\cite{mergemoe} formulates merging in output space and estimates a compression matrix by least squares. In each case, aggregation changes the correspondence between router decisions and the expert functions that are ultimately executed.

\paragraph{Re-parameterizing experts.}
Expert Editing keeps the logical expert count fixed while replacing each expert with a lower-dimensional, shared, or tensorized representation. MoE-SVD~\cite{moe-svd} applies singular value decomposition to expert weight matrices. MoLAE~\cite{molae} separates a shared latent mapping from expert-specific transformations, while MoBE~\cite{mobe} represents each expert through an expert-specific component and a learned combination of shared basis matrices. TD-MoE~\cite{td_moe} models experts jointly as correlated tensors and applies whitening followed by Tucker-style decomposition. Although these methods preserve routing slots, the functions associated with those slots are modified, so the pre-compression router may no longer be optimally matched to the edited experts.

\paragraph{Combining compression operators.}
Hybrid methods combine removal, aggregation, and low-rank re-parameterization in a single pipeline. DM-MoE~\cite{drop_or_merge} adaptively drops experts before merging the remaining ones, while DERN~\cite{dern} prunes experts, reallocates their neurons, and recombines the resulting components. EEP~\cite{eep} couples gradient-free expert pruning with knowledge-preserving merging. MoNE~\cite{mone} replaces redundant experts with lightweight, input-agnostic novice vectors, and MoE-I$^2$~\cite{moei2} combines inter-expert pruning, intra-expert low-rank decomposition, and LoRA-based recovery. $D^{2}$-MoE~\cite{d2moe} represents experts as low-rank edits around a shared base, whereas MC-SMoE~\cite{mcsmoe} first integrates experts using routing-policy statistics and then applies low-rank compression. These methods illustrate that the residual error after compression can depend on several coupled transformations rather than on a single operator.

\subsection{Post-Compression Adjustment and Recovery Training}
\label{appendix:post_compression_adjustment_related}

Several MoE compression methods attach a recovery procedure to a particular compressor. MoE-Pruner~\cite{moepruner} applies expert-wise knowledge distillation after pruning. CD-MoE~\cite{cdmoe} condenses sparse MoE layers and fine-tunes only the condensed components. MoE-I$^2$~\cite{moei2} uses LoRA after its pruning-and-decomposition pipeline, while Sub-MoE~\cite{submoe} reports additional gains from fine-tuning after subspace expert merging. At substantially larger training scales, SlimMoE~\cite{slimmoe} combines structured expert slimming with distillation, and SlimQwen~\cite{slimqwen} studies pruning, distillation, and continued training as part of large-scale MoE model construction. Beyond these compressor-specific recovery procedures, prior work analyzes router--expert mismatch across pruning, editing, and merging and introduces Router Knowledge Distillation (Router KD), which distills the original model's next-token distribution while updating only the compressed model's router~\cite{hyeon2026retraining}. The router-only KD setting evaluated here corresponds to Router KD in its objective and trainable parameter scope; we include it as a baseline within a broader comparison of training objectives and parameter-update scopes.

These studies show that compressed MoE checkpoints can remain recoverable, but their recovery procedures are generally coupled to one compressor, one objective, or one predefined trainable scope. To our knowledge, the works reviewed above do not provide a matched comparison on the same compressed checkpoints and under the same small data budget of (i) causal LM fine-tuning versus teacher-based KD, (ii) router-only, selected-expert, all-expert, and full-parameter updates, and (iii) recovery gain versus end-to-end GPU cost. This distinction matters because a smaller trainable parameter set does not automatically imply lower total cost: expert-subset strategies can require a separate selection pass, and KD additionally requires teacher forward passes.

The present paper therefore treats \textbf{Post-Compression Adjustment} itself as the object of study rather than as a method-specific add-on. Under a matched small-data protocol, we compare causal LM fine-tuning and teacher-based KD across multiple trainable scopes, and we measure both performance recovery and GPU cost. The main experiments apply this protocol to pruning and merging checkpoints across multiple retention ratios and MoE backbones, while the Expert Editing appendix reports supplementary compatibility-mode evidence. This design asks a practical question left open by compressor-specific recovery studies: \textit{given an already-compressed MoE LLM, which adjustment objective and parameter scope provide the strongest cost--recovery trade-off?}

\section{Additional Calibration Robustness}
\label{sec:additional-robustness}

We conduct two complementary checks of whether the aggregate Post-Compression Adjustment (PCA) conclusions depend on the size or domain of the calibration corpus. Both checks retain the one-epoch adjustment protocol, adjustment objectives and scopes, evaluation procedure, and the same 28 downstream benchmarks; only the calibration budget or corpus is changed. Figures~\ref{fig:small-budget-recovery} and~\ref{fig:calibration-domain-consistency} summarize the main patterns, while Tables~\ref{tab:small-budget-aggregate} and~\ref{tab:calibration-domain-aggregate} report the strategy-level performance and measured costs.

\paragraph{Smaller calibration budget.}
We reduce the C4 adjustment budget from 3,000 to 1,024 examples for four backbone--compressor pairs---Qwen3/REAP, Qwen3/HC-SMoE, Gemma4/AIMER, and Gemma4/M-SMoE---at 50\%, 62.5\%, and 75\% expert-retention ratios. Table~\ref{tab:small-budget-aggregate} averages the resulting 12 settings over all 28 evaluation benchmarks. Although the smaller budget reduces the amount of available adjustment data, the leading pattern is preserved: Full FT gives the largest mean recovery ($+2.78$ pp), followed by Router All-Expert FT ($+1.86$ pp) and Full KD ($+1.77$ pp), and FT remains ahead of KD at every matched trainable scope. The table additionally reports end-to-end GPU energy, effective GPU-hours, and GPU-memory occupancy in GiB-hours, including expert-selection and online-teacher stages when applicable.

\begin{table*}[t]
\centering
\small
\setlength{\tabcolsep}{7pt}
\begin{tabular}{lrrrrr}
\toprule
Strategy & Score & $\Delta$ (pp) & Energy (kWh) & Eff. GPU-h & Mem. (GiB-h) \\
\midrule
Original & 0.484 & +9.31 & -- & -- & -- \\
Compressed Only & 0.390 & +0.00 & -- & -- & -- \\
\addlinespace[2pt]
Full FT & 0.418 & +2.78 & 0.200 & 0.213 & 93.7 \\
Router All-Expert FT & 0.409 & +1.86 & 0.201 & 0.209 & 85.7 \\
Router Top-50 FT & 0.406 & +1.58 & 0.255 & 0.248 & 75.0 \\
Router Top-16 FT & 0.401 & +1.03 & 0.241 & 0.242 & 55.1 \\
Router Top-8 FT & 0.399 & +0.82 & 0.231 & 0.239 & 50.7 \\
Router-only FT & 0.392 & +0.13 & 0.116 & 0.090 & 20.8 \\
\addlinespace[2pt]
Full KD & 0.408 & +1.77 & 0.272 & 0.262 & 177.2 \\
Router All-Expert KD & 0.401 & +1.09 & 0.266 & 0.259 & 166.8 \\
Router Top-50 KD & 0.401 & +1.05 & 0.311 & 0.297 & 145.6 \\
Router Top-16 KD & 0.398 & +0.74 & 0.302 & 0.289 & 113.2 \\
Router Top-8 KD & 0.396 & +0.60 & 0.306 & 0.290 & 107.0 \\
Router-only KD & 0.391 & +0.08 & 0.182 & 0.139 & 68.6 \\
\addlinespace[2pt]
Direct Router Logit Match & 0.390 & -0.08 & 0.181 & 0.137 & 65.1 \\
\bottomrule
\end{tabular}
\caption{Adjustment with a 1,024-example C4 calibration budget. Values are macro-averaged over 28 evaluation benchmarks and 12 backbone--compressor--retention settings (four backbone--compressor pairs at three retention ratios). $\Delta$ is the score change from the unadjusted compressed model in percentage points. Costs are end-to-end means; Mem. is GPU-memory occupancy in GiB-hours. Original and Compressed Only require no adjustment and therefore have no adjustment cost. The Original and Compressed Only scores for this check were re-evaluated; their aggregate gap differs from Table~\ref{tab:pca-overall-performance-cost} by evaluation-level noise of about 0.05 pp.}
\label{tab:small-budget-aggregate}
\end{table*}

\paragraph{Calibration-domain shift.}
We next compare C4 with the math-domain OpenR1-Math-220k dataset~\cite{openr1} under the same 1,024-example calibration budget. The OpenR1-Math runs use the first 1,024 examples from the dataset's \texttt{all/train} split. We evaluate six matched settings: Qwen3/HC-SMoE and Gemma4/AIMER, each at 50\%, 62.5\%, and 75\% expert retention. The unadjusted checkpoints and the 28-task evaluation suite are held fixed, so the comparison isolates the calibration-corpus change within these settings.

Table~\ref{tab:calibration-domain-aggregate} shows that the aggregate strategy-level ordering is strongly associated across the two domains (Spearman $\rho=0.973$; Kendall $\tau_b=0.897$). Full FT has the largest two-domain mean recovery (4.04 pp), and the same three strategies---Full FT, Full KD, and Router All-Expert FT---form the top-three set in both domain aggregates. This result supports robustness of the paper's overall strategy trend to a calibration-domain change. It does not imply that the ranking is invariant for every compressed checkpoint: the six setting-level Spearman correlations average 0.672 and range from 0.071 to 0.962. Accordingly, we treat the domain-shift result as aggregate evidence and retain the setting-level variation as an explicit limitation.

\begin{table*}[t]
\centering
\scriptsize
\setlength{\tabcolsep}{7pt}
\begin{tabular}{lrrrrrrr}
\toprule
& \multicolumn{2}{c}{Mean score} & \multicolumn{2}{c}{Recovery (pp)} & \multicolumn{3}{c}{Math-calibration cost} \\
\cmidrule(lr){2-3}\cmidrule(lr){4-5}\cmidrule(lr){6-8}
Strategy & C4 & Math & C4 & Math & kWh & Eff. GPU-h & GiB-h \\
\midrule
Original & 0.484 & 0.484 & +11.66 & +11.66 & -- & -- & -- \\
Compressed Only & 0.367 & 0.367 & +0.00 & +0.00 & -- & -- & -- \\
\addlinespace[2pt]
Full FT & 0.402 & 0.413 & +3.50 & +4.58 & 0.126 & 0.098 & 70.1 \\
Router All-Expert FT & 0.386 & 0.407 & +1.89 & +4.01 & 0.124 & 0.093 & 64.3 \\
Router Top-50 FT & 0.385 & 0.404 & +1.81 & +3.70 & 0.167 & 0.124 & 52.7 \\
Router Top-16 FT & 0.381 & 0.395 & +1.37 & +2.78 & 0.152 & 0.117 & 32.4 \\
Router Top-8 FT & 0.380 & 0.384 & +1.28 & +1.71 & 0.146 & 0.115 & 28.4 \\
Router-only FT & 0.370 & 0.369 & +0.29 & +0.19 & 0.094 & 0.074 & 16.2 \\
Full KD & 0.388 & 0.414 & +2.14 & +4.66 & 0.202 & 0.152 & 149.6 \\
Router All-Expert KD & 0.381 & 0.391 & +1.44 & +2.44 & 0.196 & 0.148 & 141.0 \\
Router Top-50 KD & 0.383 & 0.388 & +1.61 & +2.14 & 0.232 & 0.177 & 117.7 \\
Router Top-16 KD & 0.380 & 0.380 & +1.25 & +1.27 & 0.222 & 0.171 & 84.3 \\
Router Top-8 KD & 0.378 & 0.375 & +1.12 & +0.84 & 0.225 & 0.170 & 78.2 \\
Router-only KD & 0.369 & 0.368 & +0.21 & +0.11 & 0.165 & 0.127 & 61.6 \\
\addlinespace[2pt]
Direct Router Logit Match & 0.367 & 0.367 & +0.02 & -0.01 & 0.163 & 0.125 & 58.0 \\
\bottomrule
\end{tabular}
\caption{Calibration-domain comparison on six matched settings (Qwen3/HC-SMoE and Gemma4/AIMER, each at three retention ratios). Both C4 and OpenR1-Math-220k~\cite{openr1} use 1,024 calibration examples, and scores average the same 28 evaluation benchmarks. Recovery is measured from the unadjusted compressed model. The last three columns report end-to-end adjustment costs for OpenR1-Math calibration; GiB-h is GPU-memory occupancy in GiB-hours.}
\label{tab:calibration-domain-aggregate}
\end{table*}

\section{Behavioral Proxies}
\label{sec:behavioral-proxies}

We complement the performance analysis with six heterogeneous behavioral proxies, evaluated for the Original model, the unadjusted Compressed Only checkpoint, all LM fine-tuning and KD scopes, and Direct Router Logit Match over the same 12 backbone--compressor--retention settings, using the checkpoints adjusted with the 1,024-example C4 budget of Appendix~\ref{sec:additional-robustness} (Table~\ref{tab:behavioral-proxies}). These measurements are diagnostic proxies rather than a comprehensive safety evaluation. Their meanings and preferred directions differ, so we report them separately and neither average them into a scalar safety score nor identify a single ``safest'' strategy. Higher is better for IFEval \cite{IFEval}, TruthfulQA MC1 \cite{truthfulqa}, and WinoGender \cite{WinoGender} accuracy; ToxiGen \cite{ToxiGen} here is classification accuracy rather than a direct measure of generated toxicity; higher WMDP \cite{WMDP} indicates greater hazardous-knowledge accuracy; and a CrowS-Pairs \cite{CrowS-Pairs} score closer to 0.5 is more neutral.

\paragraph{Compression and subsequent adjustment.}
On IFEval, TruthfulQA, and ToxiGen, compression changes the scores from 0.853 to 0.761, from 0.421 to 0.365, and from 0.812 to 0.708, respectively. The means across the 13 adjusted strategies are 0.756, 0.362, and 0.705, corresponding to additional changes of only $-0.47$, $-0.24$, and $-0.23$ pp relative to Compressed Only when computed from the unrounded scores. Thus, at the strategy-aggregate level, compression is the larger source of change on these three proxies. This is not uniform across strategies or metrics: Full FT further changes IFEval by $-4.13$ pp, while its TruthfulQA and ToxiGen changes are $-0.38$ and $-1.49$ pp. We therefore do not interpret the aggregate result as evidence that every PCA strategy is behaviorally neutral.

\paragraph{Restoration has a dual character.}
For WMDP, CrowS-Pairs, and WinoGender, adjustment tends to move the strategy-mean score back toward, or slightly beyond, the Original model's level. From Compressed Only, the 13-strategy means change from 0.437 to 0.450 on WMDP, from 0.556 to 0.565 on CrowS-Pairs, and from 0.567 to 0.585 on WinoGender; the corresponding Original scores are 0.508, 0.576, and 0.581. Full FT produces larger shifts to 0.477, 0.577, and 0.610. This pattern is consistent with PCA restoring parts of the behavioral and knowledge profile attenuated by compression, but the restoration is not selectively beneficial: it can recover useful behavior such as coreference accuracy while also recovering hazardous knowledge measured by WMDP or stereotype-related associations reflected by CrowS-Pairs. These proxies do not establish the underlying causal mechanism, and they should not be read as showing that PCA improves safety.

\paragraph{Implications.}
PCA is designed here as an efficiency and capability-recovery stage, not as a safety-alignment intervention. A compressed and adjusted model should therefore undergo separate, application-specific safety and alignment evaluation after PCA, with additional mitigation or alignment when needed. Designing safety-aware PCA objectives is an important direction for future work.

\begin{table*}[t]
\centering
\small
\setlength{\tabcolsep}{5.1pt}
\begin{tabular}{lrrrrrr}
\toprule
Strategy & IFEval & TruthfulQA & ToxiGen & WMDP & CrowS & WinoGender \\
\midrule
Original & 0.853 & 0.421 & 0.812 & 0.508 & 0.576 & 0.581 \\
Compressed Only & 0.761 & 0.365 & 0.708 & 0.437 & 0.556 & 0.567 \\
\addlinespace[2pt]
Full FT & 0.720 & 0.361 & 0.693 & 0.477 & 0.577 & 0.610 \\
Router All-Expert FT & 0.752 & 0.360 & 0.702 & 0.461 & 0.572 & 0.591 \\
Router Top-50 FT & 0.755 & 0.360 & 0.711 & 0.460 & 0.570 & 0.588 \\
Router Top-16 FT & 0.760 & 0.360 & 0.712 & 0.455 & 0.569 & 0.582 \\
Router Top-8 FT & 0.754 & 0.361 & 0.709 & 0.452 & 0.567 & 0.581 \\
Router-only FT & 0.765 & 0.362 & 0.708 & 0.437 & 0.560 & 0.572 \\
\addlinespace[2pt]
Full KD & 0.753 & 0.368 & 0.690 & 0.451 & 0.568 & 0.594 \\
Router All-Expert KD & 0.767 & 0.362 & 0.709 & 0.449 & 0.564 & 0.590 \\
Router Top-50 KD & 0.763 & 0.363 & 0.706 & 0.448 & 0.562 & 0.588 \\
Router Top-16 KD & 0.766 & 0.363 & 0.708 & 0.446 & 0.561 & 0.585 \\
Router Top-8 KD & 0.760 & 0.364 & 0.708 & 0.444 & 0.561 & 0.582 \\
Router-only KD & 0.757 & 0.362 & 0.707 & 0.438 & 0.561 & 0.567 \\
\addlinespace[2pt]
Direct Router Logit Match & 0.757 & 0.363 & 0.708 & 0.436 & 0.555 & 0.572 \\
\bottomrule
\end{tabular}
\caption{Behavioral-proxy results averaged over 12 backbone--compressor--retention settings. Adjusted strategies are the 1,024-example C4 runs of Table~\ref{tab:small-budget-aggregate}. The table jointly reports the two baselines, all LM fine-tuning and KD scopes, and Direct Router Logit Match. Metric directions differ: higher is better for IFEval, TruthfulQA MC1, and WinoGender accuracy; ToxiGen is classification accuracy rather than a monotone safety score; higher WMDP means greater hazardous-knowledge accuracy; and CrowS is more neutral near 0.5. We therefore neither average across columns nor designate a single ``safest'' condition.}
\label{tab:behavioral-proxies}
\end{table*}

\begin{table*}[t]
\centering
\small
\setlength{\tabcolsep}{4.0pt}
\resizebox{\textwidth}{!}{%
\begin{tabular}{lrrrrrr}
\toprule
Strategy & Score & $\Delta$ (pp) & Gap (\%) & Energy (kWh) & Eff. GPU-h & Memory (GiB-h) \\
\midrule
Original & 0.483650 & -- & -- & -- & -- & -- \\
Compressed Only & 0.390050 & +0.00 & -- & -- & -- & -- \\
\midrule
Full FT & 0.424975 & +3.49 & 37.3 & 0.576679 & 0.622545 & 273.4334 \\
Full KD & 0.411908 & +2.19 & 23.4 & 0.786771 & 0.764422 & 516.9228 \\
Router-only FT & 0.392342 & +0.23 & 2.4 & 0.337770 & 0.263998 & 61.3633 \\
Router-only KD & 0.391792 & +0.17 & 1.9 & 0.531668 & 0.406578 & 207.5186 \\
Router Top-8 FT & 0.402108 & +1.21 & 12.9 & 0.664952 & 0.694597 & 146.5624 \\
Router Top-8 KD & 0.398592 & +0.85 & 9.1 & 0.886884 & 0.848731 & 311.7546 \\
Router Top-16 FT & 0.403500 & +1.35 & 14.4 & 0.693858 & 0.705302 & 159.4321 \\
Router Top-16 KD & 0.399292 & +0.92 & 9.9 & 0.876166 & 0.847518 & 330.2068 \\
Router Top-50 FT & 0.409750 & +1.97 & 21.0 & 0.735050 & 0.722846 & 217.2813 \\
Router Top-50 KD & 0.403592 & +1.35 & 14.5 & 0.898942 & 0.867598 & 422.4902 \\
Router All-Expert FT & 0.413108 & +2.31 & 24.6 & 0.576720 & 0.611530 & 249.1521 \\
Router All-Expert KD & 0.405167 & +1.51 & 16.2 & 0.766984 & 0.757444 & 484.3358 \\
Direct Router Logit Match & 0.390692 & +0.06 & 0.7 & 0.521777 & 0.399852 & 187.9796 \\
\bottomrule
\end{tabular}}
\caption{Overall performance and measured cost of the 13 Post-Compression Adjustment strategies. Score is the unweighted mean over 28 benchmarks and 12 compressor--retention settings (four compressors $\times$ three retention ratios). $\Delta$ is the score change from Compressed Only in percentage points, and Gap is the recovered fraction of the aggregate Original-to-Compressed gap. Cost columns are unweighted means over the same 12 settings; GPU-memory values are time-integrated occupancy. Expert-selection and online-teacher passes are included when applicable.}
\label{tab:pca-overall-performance-cost}
\end{table*}

\begin{table*}[t]
\centering
\scriptsize
\setlength{\tabcolsep}{5.0pt}
\renewcommand{\arraystretch}{1.18}
\begin{tabular*}{\textwidth}{@{\extracolsep{\fill}}lccc@{}}
\toprule
Strategy & 50\% retention & 62.5\% retention & 75\% retention \\
\midrule
\multicolumn{4}{c}{Each cell: Score; then GPU energy / effective GPU-hours / GPU-memory occupancy} \\
\midrule
Original & \shortstack{0.483650\\[1pt]{\scriptsize -- / -- / --}} & \shortstack{0.483650\\[1pt]{\scriptsize -- / -- / --}} & \shortstack{0.483650\\[1pt]{\scriptsize -- / -- / --}} \\
Compressed Only & \shortstack{0.346475\\[1pt]{\scriptsize -- / -- / --}} & \shortstack{0.388450\\[1pt]{\scriptsize -- / -- / --}} & \shortstack{0.435225\\[1pt]{\scriptsize -- / -- / --}} \\
\midrule
Full FT & \shortstack{0.359000\\[1pt]{\scriptsize 0.449945 / 0.471098 / 178.8026}} & \shortstack{0.423075\\[1pt]{\scriptsize 0.575316 / 0.620035 / 267.0727}} & \shortstack{0.492850\\[1pt]{\scriptsize 0.704776 / 0.776501 / 374.4249}} \\
Full KD & \shortstack{0.358125\\[1pt]{\scriptsize 0.660507 / 0.614883 / 386.0558}} & \shortstack{0.410475\\[1pt]{\scriptsize 0.786565 / 0.760321 / 511.9236}} & \shortstack{0.467125\\[1pt]{\scriptsize 0.913239 / 0.918064 / 652.7890}} \\
Router-only FT & \shortstack{0.332550\\[1pt]{\scriptsize 0.289110 / 0.226473 / 44.4989}} & \shortstack{0.389950\\[1pt]{\scriptsize 0.335920 / 0.264699 / 60.0503}} & \shortstack{0.454525\\[1pt]{\scriptsize 0.388279 / 0.300822 / 79.5408}} \\
Router-only KD & \shortstack{0.333550\\[1pt]{\scriptsize 0.481915 / 0.369150 / 174.5792}} & \shortstack{0.389550\\[1pt]{\scriptsize 0.531264 / 0.407367 / 206.1724}} & \shortstack{0.452275\\[1pt]{\scriptsize 0.581826 / 0.443215 / 241.8043}} \\
Router Top-8 FT & \shortstack{0.346050\\[1pt]{\scriptsize 0.527959 / 0.533428 / 96.8893}} & \shortstack{0.399875\\[1pt]{\scriptsize 0.668756 / 0.692454 / 143.8744}} & \shortstack{0.460400\\[1pt]{\scriptsize 0.798140 / 0.857908 / 198.9235}} \\
Router Top-8 KD & \shortstack{0.342775\\[1pt]{\scriptsize 0.746301 / 0.685103 / 239.8124}} & \shortstack{0.396525\\[1pt]{\scriptsize 0.885385 / 0.844322 / 306.9395}} & \shortstack{0.456475\\[1pt]{\scriptsize 1.028967 / 1.016766 / 388.5119}} \\
Router Top-16 FT & \shortstack{0.346950\\[1pt]{\scriptsize 0.555016 / 0.542765 / 107.7231}} & \shortstack{0.401625\\[1pt]{\scriptsize 0.690790 / 0.702967 / 155.4038}} & \shortstack{0.461925\\[1pt]{\scriptsize 0.835770 / 0.870173 / 215.1695}} \\
Router Top-16 KD & \shortstack{0.343575\\[1pt]{\scriptsize 0.743883 / 0.681675 / 259.9779}} & \shortstack{0.396925\\[1pt]{\scriptsize 0.871934 / 0.843130 / 323.9177}} & \shortstack{0.457375\\[1pt]{\scriptsize 1.012682 / 1.017748 / 406.7249}} \\
Router Top-50 FT & \shortstack{0.350250\\[1pt]{\scriptsize 0.587784 / 0.559277 / 155.0790}} & \shortstack{0.410375\\[1pt]{\scriptsize 0.734636 / 0.720651 / 213.6255}} & \shortstack{0.468625\\[1pt]{\scriptsize 0.882731 / 0.888610 / 283.1395}} \\
Router Top-50 KD & \shortstack{0.349025\\[1pt]{\scriptsize 0.759724 / 0.703031 / 339.5399}} & \shortstack{0.402775\\[1pt]{\scriptsize 0.898835 / 0.863186 / 418.2020}} & \shortstack{0.458975\\[1pt]{\scriptsize 1.038266 / 1.036576 / 509.7286}} \\
Router All-Expert FT & \shortstack{0.350925\\[1pt]{\scriptsize 0.447933 / 0.459157 / 160.0400}} & \shortstack{0.410700\\[1pt]{\scriptsize 0.573588 / 0.608386 / 241.7992}} & \shortstack{0.477700\\[1pt]{\scriptsize 0.708640 / 0.767046 / 345.6170}} \\
Router All-Expert KD & \shortstack{0.350175\\[1pt]{\scriptsize 0.639433 / 0.605543 / 355.7063}} & \shortstack{0.404450\\[1pt]{\scriptsize 0.763486 / 0.753722 / 475.4654}} & \shortstack{0.460875\\[1pt]{\scriptsize 0.898032 / 0.913069 / 621.8356}} \\
Direct Router Logit Match & \shortstack{0.328625\\[1pt]{\scriptsize 0.470576 / 0.361338 / 156.2195}} & \shortstack{0.390650\\[1pt]{\scriptsize 0.522851 / 0.399363 / 187.2454}} & \shortstack{0.452800\\[1pt]{\scriptsize 0.571905 / 0.438855 / 220.4739}} \\
\bottomrule
\end{tabular*}
\caption{Retention-ratio sensitivity with all three measured GPU-cost metrics. Each cell reports Score on the first line and $E/H/M$ on the second, where $E$ is GPU energy (kWh), $H$ is effective GPU-hours, and $M$ is GPU-memory occupancy (GiB-hours). Score is an unweighted mean over four compressor--backbone settings and 28 benchmarks at the indicated retention ratio; each cost is the unweighted mean of the corresponding four strategy runs. Expert-selection and online-teacher passes are included when applicable.}
\label{tab:pca-by-retention}
\end{table*}

\begin{table*}[t]
\centering
\scriptsize
\setlength{\tabcolsep}{4.0pt}
\renewcommand{\arraystretch}{1.18}
\begin{tabular*}{\textwidth}{@{\extracolsep{\fill}}lll r cc@{}}
\toprule
Compressor & Backbone & Paradigm & Compressed & Full FT & Full KD \\
\midrule
\multicolumn{6}{c}{Full FT/KD cells: Score ($\Delta$ pp); then E/H/M} \\
\midrule
AIMER & Gemma4 & Pruning & 0.280333 & \shortstack{0.314933 (+3.46)\\[1pt]{\scriptsize 0.088906 / 0.119960 / 24.6553}} & \shortstack{0.297067 (+1.67)\\[1pt]{\scriptsize 0.118888 / 0.150347 / 44.9570}} \\
HC-SMoE & Qwen3 & Merging & 0.454000 & \shortstack{0.510667 (+5.67)\\[1pt]{\scriptsize 0.557599 / 0.394955 / 339.4779}} & \shortstack{0.493333 (+3.93)\\[1pt]{\scriptsize 0.939317 / 0.652387 / 740.1165}} \\
M-SMoE & Gemma4 & Merging & 0.297167 & \shortstack{0.322833 (+2.57)\\[1pt]{\scriptsize 0.970750 / 1.490283 / 299.4836}} & \shortstack{0.304800 (+0.76)\\[1pt]{\scriptsize 1.010224 / 1.510365 / 411.3137}} \\
REAP & Qwen3 & Pruning & 0.528700 & \shortstack{0.551467 (+2.28)\\[1pt]{\scriptsize 0.689461 / 0.484980 / 430.1168}} & \shortstack{0.552433 (+2.37)\\[1pt]{\scriptsize 1.078653 / 0.744590 / 871.3040}} \\
\bottomrule
\end{tabular*}
\caption{Integrated compressor-level performance and cost, averaged over the three retention ratios. Full FT and Full KD cells report Score and the change from that compressor's unadjusted checkpoint on the first line, followed by GPU energy (kWh), effective GPU-hours, and GPU-memory occupancy (GiB-hours) as $E/H/M$. Performance additionally averages 28 benchmarks; each cost averages three runs. Each compressor is paired with one backbone, so compressor and backbone effects are not separately identifiable.}
\label{tab:pca-by-compressor}
\end{table*}

\begin{table*}[t]
\centering
\scriptsize
\setlength{\tabcolsep}{5.0pt}
\renewcommand{\arraystretch}{1.18}
\begin{tabular*}{\textwidth}{@{\extracolsep{\fill}}lccc@{}}
\toprule
Strategy & Expert Pruning & Expert Merging & Expert Editing \\
\midrule
\multicolumn{4}{c}{Each cell: Score ($\Delta$ pp); then E/H/M} \\
\midrule
Original & \shortstack{0.483650\\[1pt]{\scriptsize -- / -- / --}} & \shortstack{0.483650\\[1pt]{\scriptsize -- / -- / --}} & \shortstack{0.483650\\[1pt]{\scriptsize -- / -- / --}} \\
Compressed Only & \shortstack{0.404517 (+0.00)\\[1pt]{\scriptsize -- / -- / --}} & \shortstack{0.375583 (+0.00)\\[1pt]{\scriptsize -- / -- / --}} & \shortstack{0.337283 (+0.00)\\[1pt]{\scriptsize -- / -- / --}} \\
\midrule
Full FT & \shortstack{0.433200 (+2.87)\\[1pt]{\scriptsize 0.389184 / 0.302470 / 227.3860}} & \shortstack{0.416750 (+4.12)\\[1pt]{\scriptsize 0.764175 / 0.942619 / 319.4808}} & \shortstack{0.393000 (+5.57)\\[1pt]{\scriptsize -- / -- / --}} \\
Full KD & \shortstack{0.424750 (+2.02)\\[1pt]{\scriptsize 0.598771 / 0.447469 / 458.1305}} & \shortstack{0.399067 (+2.35)\\[1pt]{\scriptsize 0.974771 / 1.081376 / 575.7151}} & \shortstack{0.377533 (+4.03)\\[1pt]{\scriptsize -- / -- / --}} \\
Router-only FT & \shortstack{0.405483 (+0.10)\\[1pt]{\scriptsize 0.292612 / 0.228956 / 52.8989}} & \shortstack{0.379200 (+0.36)\\[1pt]{\scriptsize 0.382928 / 0.299040 / 69.8278}} & \shortstack{0.339817 (+0.25)\\[1pt]{\scriptsize -- / -- / --}} \\
Router-only KD & \shortstack{0.404433 (-0.01)\\[1pt]{\scriptsize 0.486499 / 0.372627 / 189.4268}} & \shortstack{0.379150 (+0.36)\\[1pt]{\scriptsize 0.576838 / 0.440528 / 225.6104}} & \shortstack{0.340917 (+0.36)\\[1pt]{\scriptsize -- / -- / --}} \\
Router Top-8 FT & \shortstack{0.411183 (+0.67)\\[1pt]{\scriptsize 0.457579 / 0.363210 / 90.3311}} & \shortstack{0.393033 (+1.74)\\[1pt]{\scriptsize 0.872324 / 1.025983 / 202.7938}} & \shortstack{0.357433 (+2.02)\\[1pt]{\scriptsize -- / -- / --}} \\
Router Top-8 KD & \shortstack{0.408600 (+0.41)\\[1pt]{\scriptsize 0.668524 / 0.512920 / 234.4281}} & \shortstack{0.388583 (+1.30)\\[1pt]{\scriptsize 1.105244 / 1.184541 / 389.0810}} & \shortstack{0.352833 (+1.55)\\[1pt]{\scriptsize -- / -- / --}} \\
Router Top-16 FT & \shortstack{0.413350 (+0.88)\\[1pt]{\scriptsize 0.479891 / 0.372069 / 103.7273}} & \shortstack{0.393650 (+1.81)\\[1pt]{\scriptsize 0.907826 / 1.038535 / 215.1369}} & \shortstack{0.364167 (+2.69)\\[1pt]{\scriptsize -- / -- / --}} \\
Router Top-16 KD & \shortstack{0.409733 (+0.52)\\[1pt]{\scriptsize 0.668031 / 0.513285 / 257.1824}} & \shortstack{0.388850 (+1.33)\\[1pt]{\scriptsize 1.084301 / 1.181751 / 403.2312}} & \shortstack{0.357733 (+2.04)\\[1pt]{\scriptsize -- / -- / --}} \\
Router Top-50 FT & \shortstack{0.421133 (+1.66)\\[1pt]{\scriptsize 0.521821 / 0.391438 / 168.0111}} & \shortstack{0.398367 (+2.28)\\[1pt]{\scriptsize 0.948279 / 1.054254 / 266.5515}} & \shortstack{0.375733 (+3.84)\\[1pt]{\scriptsize -- / -- / --}} \\
Router Top-50 KD & \shortstack{0.415683 (+1.12)\\[1pt]{\scriptsize 0.692720 / 0.534534 / 354.8760}} & \shortstack{0.391500 (+1.59)\\[1pt]{\scriptsize 1.105163 / 1.200661 / 490.1043}} & \shortstack{0.365833 (+2.85)\\[1pt]{\scriptsize -- / -- / --}} \\
Router All-Expert FT & \shortstack{0.424383 (+1.99)\\[1pt]{\scriptsize 0.382234 / 0.289516 / 208.7277}} & \shortstack{0.401833 (+2.62)\\[1pt]{\scriptsize 0.771207 / 0.933543 / 289.5764}} & \shortstack{0.389250 (+5.20)\\[1pt]{\scriptsize -- / -- / --}} \\
Router All-Expert KD & \shortstack{0.417733 (+1.32)\\[1pt]{\scriptsize 0.578071 / 0.438173 / 428.5423}} & \shortstack{0.392600 (+1.70)\\[1pt]{\scriptsize 0.955896 / 1.076716 / 540.1293}} & \shortstack{0.373067 (+3.58)\\[1pt]{\scriptsize -- / -- / --}} \\
Direct Router Logit Match & \shortstack{0.405283 (+0.08)\\[1pt]{\scriptsize 0.480781 / 0.367415 / 174.7370}} & \shortstack{0.376100 (+0.05)\\[1pt]{\scriptsize 0.562774 / 0.432288 / 201.2222}} & \shortstack{0.339250 (+0.20)\\[1pt]{\scriptsize -- / -- / --}} \\
\bottomrule
\end{tabular*}
\caption{Performance--cost summary by compression paradigm with the complete measured cost tuple. For Expert Pruning and Expert Merging, $\Delta$ is the 28-benchmark score change from the unadjusted compressed checkpoint, averaged over two compressors and three retention ratios; $E/H/M$ denotes GPU energy (kWh), effective GPU-hours, and GPU-memory occupancy (GiB-hours), each averaged over the corresponding six runs. Expert Editing reports the same performance change over six compatibility-mode MoBE/TD-MoE settings; its cost tuple is shown as unavailable because those reconstructed checkpoints use a different parameterization and measurement boundary.}
\label{tab:pca-by-paradigm}
\end{table*}

\begin{table*}[t]
\centering
\scriptsize
\setlength{\tabcolsep}{2.5pt}
\renewcommand{\arraystretch}{1.15}
\begin{tabular}{lrrrrrrrr}
\toprule
\multicolumn{9}{c}{(a) Causal-LM fine-tuning scopes} \\
\midrule
Family & Orig. & Comp. & Full FT & \shortstack{Router-only\\FT} & \shortstack{Router Top-8\\FT} & \shortstack{Router Top-16\\FT} & \shortstack{Router Top-50\\FT} & \shortstack{Router All-Expert\\FT} \\
\midrule
General (4) & 0.450962 & 0.435135 & 0.450323 & 0.440379 & 0.447379 & 0.448342 & 0.452844 & 0.454240 \\
Math (6) & 0.412292 & 0.289432 & 0.297187 & 0.289442 & 0.296483 & 0.297693 & 0.299303 & 0.303750 \\
Coding (4) & 0.373688 & 0.242994 & 0.238104 & 0.240187 & 0.247377 & 0.245998 & 0.243981 & 0.241181 \\
CoT (6) & 0.588908 & 0.449946 & 0.540596 & 0.454926 & 0.465842 & 0.468947 & 0.486062 & 0.494618 \\
MCQA (8) & 0.529556 & 0.471557 & 0.514856 & 0.474628 & 0.488238 & 0.490059 & 0.496747 & 0.499443 \\
\bottomrule
\end{tabular}
\par\vspace{0.6em}
\begin{tabular}{lrrrrrrrrr}
\toprule
\multicolumn{10}{c}{(b) Knowledge-distillation scopes and router-logit matching} \\
\midrule
Family & Orig. & Comp. & Full KD & \shortstack{Router-only\\KD} & \shortstack{Router Top-8\\KD} & \shortstack{Router Top-16\\KD} & \shortstack{Router Top-50\\KD} & \shortstack{Router All-Expert\\KD} & \shortstack{Direct Router\\Logit Match} \\
\midrule
General (4) & 0.450962 & 0.435135 & 0.447823 & 0.435627 & 0.439844 & 0.440773 & 0.444046 & 0.446033 & 0.434225 \\
Math (6) & 0.412292 & 0.289432 & 0.301996 & 0.288156 & 0.297132 & 0.296033 & 0.300996 & 0.301814 & 0.290226 \\
Coding (4) & 0.373688 & 0.242994 & 0.250819 & 0.242962 & 0.246190 & 0.246263 & 0.248881 & 0.250410 & 0.243956 \\
CoT (6) & 0.588908 & 0.449946 & 0.492664 & 0.455543 & 0.462228 & 0.464406 & 0.469710 & 0.472000 & 0.450700 \\
MCQA (8) & 0.529556 & 0.471557 & 0.496322 & 0.474254 & 0.482503 & 0.483649 & 0.488119 & 0.489567 & 0.472597 \\
\bottomrule
\end{tabular}
\par\vspace{0.6em}
\begin{tabular}{lrrrrr}
\toprule
Family & General (4) & Math (6) & Coding (4) & CoT (6) & MCQA (8) \\
\midrule
Full FT $\Delta$ (pp) & +1.52 & +0.78 & -0.49 & +9.07 & +4.33 \\
\bottomrule
\end{tabular}
\caption{Benchmark-family performance averaged over all 12 compressor--retention settings. General contains BBH-Fewshot, BBH-Zeroshot, CoQA, and HellaSwag; the remaining families contain 6 Math, 4 Coding, 6 chain-of-thought (CoT), and 8 multiple-choice QA (MCQA) benchmarks. Cost is not allocated by benchmark because every strategy run evaluates all benchmark families after a single shared adjustment procedure; aggregate run-level costs are reported in Table~\ref{tab:pca-overall-performance-cost}. Family scores are aggregated from individually rounded benchmark scores.}
\label{tab:pca-by-benchmark-family}
\end{table*}

\end{document}